\documentclass[journal]{IEEEtran}
\usepackage{amsmath,amsfonts}
\usepackage[ruled,linesnumbered]{algorithm2e}
\usepackage{array}
\usepackage[caption=false,font=normalsize,labelfont=sf,textfont=sf]{subfig}
\usepackage{textcomp}
\usepackage{stfloats}
\usepackage{url}
\usepackage{verbatim}
\usepackage{graphicx}
\usepackage{cite}
\usepackage{booktabs}
\usepackage{tabularx}
\usepackage{multirow}
\usepackage{threeparttable}
\usepackage[normalem]{ulem}
\usepackage{bm}
\usepackage{amsthm,amssymb}
\usepackage{mathrsfs}
\usepackage{wasysym}
\usepackage{pifont}
\usepackage{makecell}
\usepackage{xcolor}
\newcolumntype{P}[1]{>{\Centering\hspace{0pt}}p{#1}}
\newcolumntype{Z}{>{\centering\let\newline\\\arraybackslash\hspace{0pt}}X}
\allowdisplaybreaks[4]
\begin{document}
\newtheorem{theorem}{Theorem}
\newtheorem{lemma}{Lemma} 
\newtheorem{assumption}{Assumption}
\newtheorem{definition}{Definition}
\newtheorem{corollary}{Corollary}
\newcommand\subscriptir{_{i,r}}
\newcommand\subscriptit{_{i,t}}
\newcommand\subscripttr[1][t]{_{r,#1}}
\newcommand\subscriptitr{_{i,r,t}}
\newcommand\subscriptitk[1][k]{_{i,r,t}^{(#1)}}
\newcommand\vol{{\rm vol}}

\newcommand\Acc{\Gamma}
\newcommand\lr{\zeta}
\newcommand\cyc{c}
\newcommand\rate{v}
\newcommand\quant{\Psi}
\newcommand\loss{L}

\newcommand\otherspi{{\pi}_{\!-\!r}}
\newcommand\selfpi{\pi_{r}}
\newcommand\jointaction[1][t]{\boldsymbol{a}_{#1}}
\newcommand\jointpi{\boldsymbol{\pi}}
\newcommand\bestjointpi{\jointpi^\dagger}
\newcommand{\bestotherspi}{\pi^{\dagger}_{\!-\!r}}
\newcommand{\bestgenotherspi}{\tilde{\pi}^{\dagger}_{\!-\!r}}
\newcommand{\selfaction}{a\subscripttr}
\newcommand{\otheraction}{a_{\!-\!r,t}}

\newcommand\subscriptrpi{_r^{\jointpi}}
\newcommand\subscriprbestpi{_r^{\bestjointpi}}
\newcommand\subscriptotheragent{_{\!-\!r,t}}
\newcommand\subscriptrstar[1][r]{_{#1}^\ast}
\newcommand\subscriptproofrt[1][r]{_{#1,t}}

\renewcommand\tabularxcolumn[1]{m{#1}}

\title{Pareto Actor-Critic for Communication and Computation Co-Optimization in Non-Cooperative Federated Learning Services}

\author{Renxuan Tan, Rongpeng Li, Xiaoxue Yu, Xianfu Chen, Xing Xu, and Zhifeng Zhao
\thanks{This work was primarily supported in part by the Zhejiang Provincial Natural Science Foundation of China under Grant LR23F010005, in part by the National Key Research and Development Program of China under Grant 2024YFE0200600, in part by Huawei Cooperation Project under Grant TC20240829036, and in part by the Big Data and Intelligent Computing Key Lab of CQUPT under Grant BDIC-2023-B-001. Xu‘s work was partially supported by the Hebei Natural Science Foundation of China under Grant F2025521001. Chen’s work was partially supported by the National Natural Science Foundation of China under Grant U24A20209.}
\thanks{ R. Tan, X. Yu and R. Li are with the College of Information Science and Electronic Engineering, Zhejiang University (email:\{ttrx, sdwhyxx, lirongpeng\}@zju.edu.cn).}
\thanks{X. Chen is with the Shenzhen CyberAray Network Technology Co., Ltd, China (e-mail: xianfu.chen@ieee.org).}
\thanks{X. Xu is with the Information and Communication Branch of State Grid Hebei Electric Power Co., Ltd, China (e-mail:hsuxing@zju.edu.cn).}
\thanks{Z. Zhao is with Zhejiang Lab as well as Zhejiang University (email: zhaozf@zhejianglab.com).}

}



\maketitle

\begin{abstract}
Federated learning (FL) in multi-service provider (SP) ecosystems is fundamentally hampered by non-cooperative dynamics, where privacy constraints and competing interests preclude the centralized optimization of multi-SP communication and computation resources. In this paper, we introduce \texttt{PAC-MCoFL}, a game-theoretic multi-agent reinforcement learning (MARL) framework where SPs act as agents to jointly optimize client assignment, adaptive quantization, and resource allocation. Within the framework, we integrate Pareto Actor-Critic (PAC) principles with expectile regression, enabling agents to conjecture optimal joint policies to achieve Pareto-optimal equilibria while modeling heterogeneous risk profiles. To manage the high-dimensional action space, we devise a ternary Cartesian decomposition (TCAD) mechanism that facilitates fine-grained control. Further, we develop \texttt{PAC-MCoFL-p}, a scalable variant featuring a parameterized conjecture generator that substantially reduces computational complexity with a provably bounded error. Alongside theoretical convergence guarantees, our framework's superiority is validated through extensive simulations -- \texttt{PAC-MCoFL} achieves approximately $5.8\%$ and $4.2\%$ improvements in total reward and hypervolume indicator (HVI), respectively, over the latest MARL solutions. The results also demonstrate that our method can more effectively balance individual SP and system performance in scaled deployments and under diverse data heterogeneity.
\end{abstract}


\section{Introduction}
\IEEEPARstart{W}{ith} the rapid proliferation of smart devices, federated learning (FL) has emerged due to its privacy-friendly nature to facilitate distributed learning. Beyond the well-studied scenario of a single service provider (SP) orchestrating FL \cite{intro7}, the burgeoning complexity of client-side applications necessitates a multi-provider ecosystem. Real-world scenarios include Apple's private cloud compute (PCC) \cite{apple2024pcc}, where multiple third-party services (via PCC encrypted proxies) interact with a shared pool of end-user devices. Compared with a single SP, this multi-SP setting, where SPs share finite communication and computational resources \cite{intro16}, introduces significant new challenges. Particularly, due to stringent privacy and competitiveness imperatives, there naturally emerges an inherent reluctance among SPs to share model details and co-optimization strategies. This reluctance becomes more pronounced for sharing the quality of service (QoS) metrics and operational trade-offs, precluding fully cooperative optimization. Under such partially observable environment, where only limited metadata on resource coordination is shared via third-party, regulatory-compliant channels \cite{3gpp_ts23501_2020}, there emerges a strong incentive to develop non-cooperative solutions.

Typically, the communication and computation co-optimization in FL shall simultaneously take into account client selection, resource allocation, and the dynamically changing network environment \cite{kim2023green}. The high dimensionality of the decision space, coupled with complex inter-variable dependencies, renders this multi-objective problem intractable for conventional optimization solvers \cite{JCARA}. In this regard, deep reinforcement learning (DRL), which learns optimal policies without prior statistical knowledge, offers a promising avenue to devise adaptive strategies. Recent advances in DRL-based optimization have demonstrated its potential in various facets of FL, including client selection \cite{FedMarl}, hyperparameter adaptive tuning \cite{rw11}, bandwidth partitioning \cite{CSBWA}, power management \cite{intro14}, and energy consumption scheduling \cite{rw13}. Nevertheless, these studies predominantly assume a single SP and cannot be straightforwardly extended to a multi-SP scenario, due to the non-cooperative nature \cite{xuBandwidthAllocationMultiple2022}. 

Multi-agent reinforcement learning (MARL) offers a promising framework to address the intricate co-optimization problem therein \cite{MAPPO,mean-filed1,mean-field2}. However, classical MARL algorithms overlook the impact of joint actions and often promote individual actions lacking synergy in early training stages \cite{le2025toward}. Furthermore, value function estimation under the mean squared error criterion implicitly assumes symmetric risk preferences \cite{Kostrikov2021OfflineRL}, and fails to capture the asymmetric risk attitudes commonly exhibited by SPs (e.g. conservative versus aggressive policy).
Consequently, games in MARL may converge to a risk-averse equilibrium rather than a Pareto-optimal one \cite{intro18}, yielding suboptimal resource allocation, degraded service quality, and inefficient global resource utilization. The Pareto actor-critic (PAC) framework \cite{paretoac} lays the groundwork for addressing equilibrium selection in MARL systems, yet its practical application to multi-SP FL is severely limited by its reliance on computationally prohibitive brute-force policy conjectures and restrictive single-dimensional action control.

In this paper, we propose a PAC-based communication and computation co-optimization scheme \texttt{PAC-MCoFL} for non-cooperative multi-SP FL services. \texttt{PAC-MCoFL} treats each FL SP as a PAC agent, which autonomously determines its operational policy and resource allocation (encompassing quantization levels, aggregation strategies, computational resource assignments, and bandwidth allocation) based on environmental observations and inferred actions of other agents. Unlike classical PAC implementations \cite{paretoac}, we first develop a critic network grounded in expectile regression to capture heterogeneous risk preferences among SPs, enabling a more granular and realistic characterization of SPs. Meanwhile, a ternary Cartesian action decomposition (TCAD) mechanism is introduced to facilitate fine-grained, multi-dimensional control over SP operations. Furthermore, we introduce \texttt{PAC-MCoFL-p}, a variant of \texttt{PAC-MCoFL}, by replacing the computationally intensive brute-force conjecture of joint actions with a neural generator. This generator can be seamlessly integrated into \texttt{PAC-MCoFL} through end-to-end training, achieving comparable performance with provably bounded approximate error.
In brief, the main contributions of this research can be summarized as follows.

\begin{itemize}
    	\item We propose \texttt{PAC-MCoFL}, a framework that enables decentralized communication and computation co-optimization in partially observable, non-cooperative multi-SP FL systems. \texttt{PAC-MCoFL} adaptively optimizes operational policies and resource allocation, leveraging game-theoretic principles to conjecture actions from other agents, thereby mitigating convergence to suboptimal equilibria. It incorporates two key mechanisms: (i) an expectile regression-based critic for capturing asymmetric risk profiles among SPs, and (ii) a TCAD mechanism for fine-grained multi-dimensional action space navigation.
    	\item We develop \texttt{PAC-MCoFL-p} to overcome the scalability bottleneck of exhaustive action conjecture in \texttt{PAC-MCoFL}. This practical variant features a parameterized conjecture generator that substantially reduces computational complexity, making the solution feasible for large-scale multi-SP scenarios. We formally establish that this approximation has a provably bounded error, ensuring its reliability.
    	\item We provide a theoretical analysis of \texttt{PAC-MCoFL}, focusing on the convergence property of $Q$-function. Extensive simulations have validated the superiority and robustness of our proposed algorithm.	 
    \end{itemize}

The remainder of this paper is organized as follows. Section \ref{section_relatedworks} summarizes relevant literature. Section \ref{section2} introduces the system model and formulates the problem. Section \ref{section3} presents the \texttt{PAC-MCoFL} algorithm, while Section \ref{section4} provides simulation results and analyses to highlight the significance of our method. Finally, Section \ref{section_conclusion} concludes the paper. 
 
 \section{Related Works}
 \label{section_relatedworks}
 \begin{table*}[tp]
    \centering
    \caption{The Summary of Differences with Related Literature.}
    \label{tab:related work}
    \def\arraystretch{1.25} 
    \hspace*{-0.2cm}
    \begin{tabular}{>{\centering\arraybackslash}p{1.5cm}
    |>{\centering\arraybackslash}m{1.5cm}
    >{\centering\arraybackslash}m{1.5cm}
    >{\centering\arraybackslash}m{1.5cm}
    >{\centering\arraybackslash}m{1.5cm}
    >{\centering\arraybackslash}m{1.7cm}|m{4.8cm}}
\toprule
    References & multi-SP& Client Assignment& Adaptive Quantization& Computation Efficiency & Communication Efficiency& Brief Description \\
    \midrule
    \cite{FedMarl},\cite{CSBWA}& $\Circle$ & $\CIRCLE$& $\Circle$& $\Circle$ & $\CIRCLE$ & Over-ideal assumption that heterogeneous services are scheduled by the same server. \\
    \hline
    \cite{rw11} & $\Circle $ & $\CIRCLE$ & $\Circle$ & $\CIRCLE$ & $\CIRCLE$ & A simple single-agent model. \\
    \hline
    \cite{liu2024fedeco} & $\Circle$ & $\Circle$ & $\CIRCLE$ & $\CIRCLE$ & $\Circle$ & Only energy efficiency analysis with traditional convex optimization methods. \\
    \hline
    \cite{xuBandwidthAllocationMultiple2022,xueBandwidthAllocationFederated2024} & $\CIRCLE$ & $\Circle$ & $\Circle$ & $\Circle$ & $\CIRCLE$ & Absence of analysis of energy consumption. \\
    \hline
    \cite{nguyenMultipleFederatedLearning2023,baiMulticoreFederatedLearning2023} & $\CIRCLE$ & $\Circle$ & $\Circle$ & $\CIRCLE$ & $\CIRCLE$ & Neglection of fine-grained optimization within a service. \\
    \hline
    
    \textbf{This work} & $\CIRCLE$ & $\CIRCLE$ & $\CIRCLE$ & $\CIRCLE$ & $\CIRCLE$ & For a FL with multiple SPs, a partial observable, non-cooperative game, involving \textit{expectile regression}, \textit{TCAD} and \textit{parameterized conjecture generator}, is introduced to achieve joint optimization of communication and computation. \\
    \bottomrule
    \multicolumn{7}{>{\footnotesize\itshape}r}{Notations: \rm{$\Circle$} \textit{indicates not included}; \rm{$\CIRCLE$} \textit{indicates fully included.}} \\
    \end{tabular}
\end{table*}
\noindent\textbf{Efficient FL with Single SP.}\quad Improving communication and computation efficiency in FL has received significant attention. Approaches like AdaQuantFL \cite{Jhunjhunwala2021AdaptiveQO} and FedDQ \cite{FedDQ} dynamically adjust quantization levels to balance communication cost and convergence. FedDrop \cite{feddropout} applies heterogeneous dropout to prune global models into personalized subnets, reducing both communication and computation loads. Moreover, FedEco \cite{liu2024fedeco} employs adaptive hyperparameter tuning to better exploit client-side computing resources, significantly lowering energy consumption across participating devices. However, these static or heuristic-based methods often fail to adapt to the dynamic and heterogeneous nature of real-world FL environments. To overcome this, recent works have applied DRL to resource optimization in wireless systems like mobile edge computing (MEC) and internet of vehicles (IoV) \cite{addedpaper1, addedpaper4}. For example, \cite{addedpaper2} uses a soft actor-critic (SAC) framework for context-aware power control in IoV-MEC scenarios, while \cite{addedpaper3} adopts a double deep $Q$-network (DDQN)-based approach to jointly optimize energy, latency, and communication overheads. Within the FL domain, Ref. \cite{CSBWA} employs REINFORCE to automate client selection and bandwidth allocation strategies based on historical feedback, reducing latency and energy while enhancing long-term FL performance. Similarly, a proximal policy optimization (PPO)-based scheme in \cite{rw11} dynamically adjusts the quantization interval during global model distribution, accelerating convergence and improving accuracy.

\noindent\textbf{Efficient FL with multi-SP.}\quad FL involving multiple SPs presents additional complexity due to the heterogeneous configurations, competing interests, and the inherent asymmetry risks of SPs \cite{baiMulticoreFederatedLearning2023,nguyenMultipleFederatedLearning2023,xuBandwidthAllocationMultiple2022}. These challenges necessitate adaptive and decentralized coordination schemes for efficient FL across SPs. To address client heterogeneity, Ref. \cite{baiMulticoreFederatedLearning2023} proposes a multi-core FL architecture where multiple global models — ranging from shallow to deep — are concurrently trained, with task inter-dependencies explicitly modeled. Resource allocation strategies among SPs have also been explored. Centralized \cite{nguyenMultipleFederatedLearning2023} and distributed \cite{xuBandwidthAllocationMultiple2022} optimization methods are proposed to allocate bandwidth and CPU power. However, most of these approaches overlook practical power constraints. Under limited power budgets, Ref. \cite{xueBandwidthAllocationFederated2024} frames the joint optimization of client selection and bandwidth allocation as a novel variant of the knapsack problem \cite{Martello1981ABA} to accelerate multiple FL models' training in wireless networks. Game-theoretic models have also been applied to capture the strategic behavior of SPs. Ref. \cite{yuIncentiveFrameworkCrossDevice2022} designs a multi-leader-follower game for multi-task FL, establishing equilibrium between SPs and clients. A two-stage Stackelberg game in \cite{fuJointOptimizationDevice2024} co-optimizes client selection and resource allocation, improving cluster efficiency while reducing FL costs. Despite these advancements, the majority of existing studies have largely downplayed the significance of non-cooperative game dynamics among multiple SPs.

\noindent\textbf{Equilibrium Games.}\quad Nash equilibrium (NE) represents a milestone in multi-agent stochastic games \cite{nashq}. Over decades, MARL has emerged as a powerful approach for dynamically identifying NE equilibria in complex, multi-agent settings, effectively bridging classical game theory with modern computational techniques \cite{friendQL,COMA,SHAQ}. In friend-and-foe $Q$-learning \cite{friendQL}, an agent $r$ assumes that its ``friend agent", $-r$, voluntarily selects actions that maximize $r$'s joint $Q$-value. Conversely, foe $Q$-learning is used in adversarial settings\cite{friendQL}, where each agent seeks to minimize the opponent's expected return. Ref. \cite{COMA} subtracts a counterfactual baseline from $Q$ function to highlight the individual's effect, facilitating rational independent reward allocation in multi-agent games. The Shapley value method is applied in Ref. \cite{SHAQ}, within the context of global reward games, to address the inaccurate allocation of contributions. On the basis of NE, many studies seek to find the Pareto optimal equilibrium for every agent involved \cite{le2025toward,paretoac}. Ref. \cite{le2025toward} connects the multiple gradient descent algorithm to the MARL, leading to strong Pareto optimal solutions. Recently, the PAC algorithm \cite{paretoac}, which operates within a non-conflict game framework, assumes multi-agent interactions in an ``optimistic" manner and ensures that all agents converge toward Pareto optimal equilibria. 

We summarize the key differences between our algorithm and highly related literature in Table \ref{tab:related work}.

\section{System Model and Problem Formulation}
\label{section2}
In Section \ref{section:sys_model}, we describe the FL process and sequentially present the models of quantization, delay, and energy consumption. Subsequently, in Section \ref{section: prob_formulation}, we introduce the optimization problem, which aims to minimize network overhead while ensuring the convergence of FL training.

Beforehand, major notations in the paper are summarized in Table \ref{tab:notations}.
\begin{figure}[tbp]
	\centerline{\includegraphics[width=\linewidth]{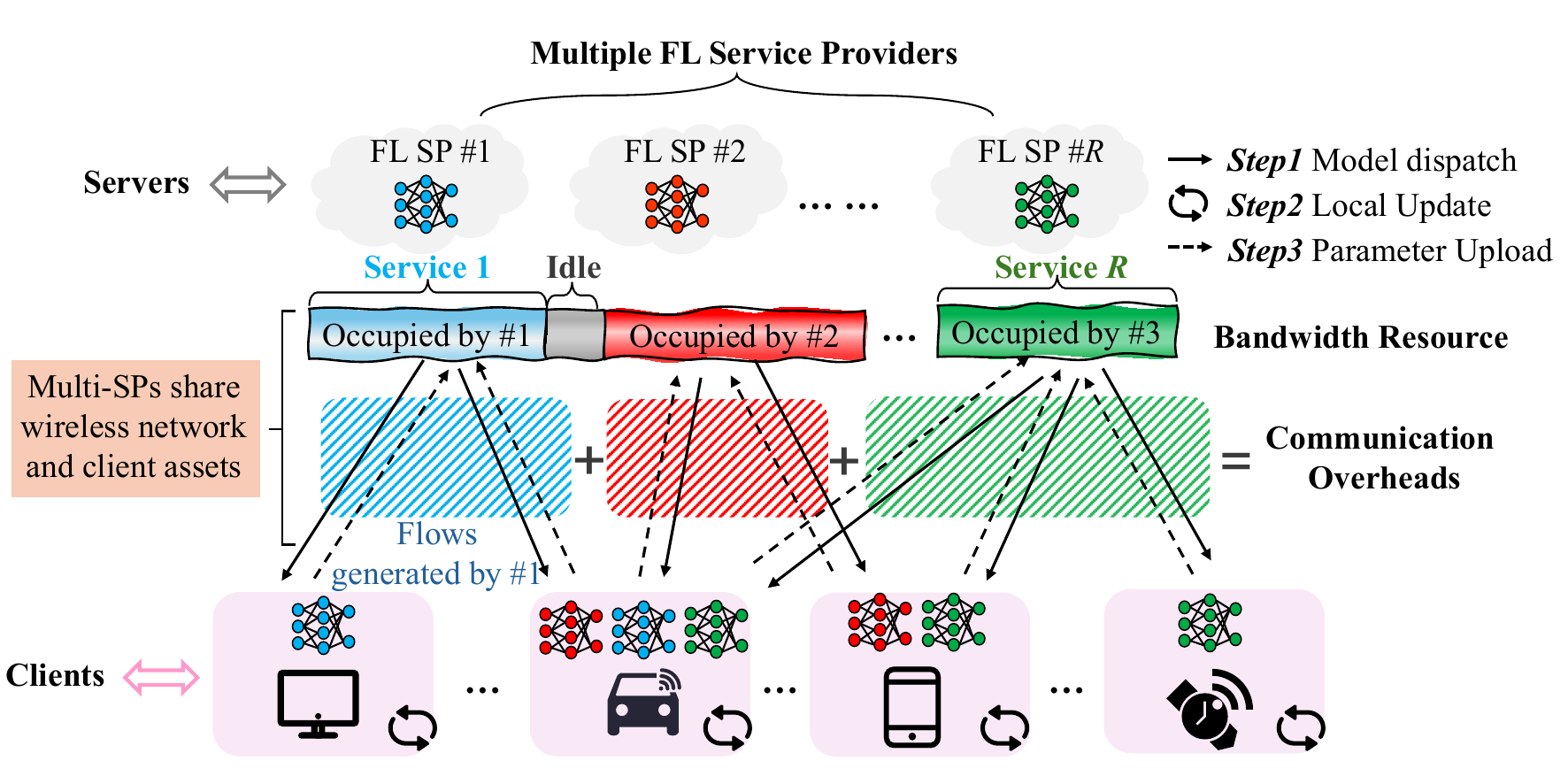}}
	\caption{A system overview of non-cooperative FL services with multiple SPs.}
	\label{fig1}
\end{figure}
\subsection{System Model}
\label{section:sys_model}
\subsubsection{FL Process}
We primarily consider an FL scenario on shared network resources, as illustrated in Fig. \ref{fig1}, which comprises a group of SPs $\mathcal{R}=\left\{1,\cdots,r,\cdots, R \right\}$ with an associated set of clients $\mathcal{N}=\left\{ 1,\cdots,i,\cdots,N \right\}$. Specifically, the server at each SP $r$ is responsible for training a specific service model, while clients leverage their private data to provide local training, benefiting the learning of these models corresponding to different SPs.
Without loss of generality, the local dataset at client $i$ associated with service $r$ is denoted as $\mathcal{D} \subscriptir$. Each sample $\xi \in \mathcal{D} \subscriptir$ 
consists of the feature set and its corresponding label. Thus, the local loss function for each client $i$ can be defined as
\begin{equation}
    \label{eq:local_loss_function}
	\loss \subscriptir\left( \boldsymbol{\omega } \right) =\frac{1}{|\mathcal{D} \subscriptir|}\sum\nolimits_{\xi\in \mathcal{D} \subscriptir}{l\left( \boldsymbol{\omega};\xi  \right)},
\end{equation}
where the function $l\left( \boldsymbol{\omega};\xi \right) $ measures the difference between predicted and true values based on the parameters $\boldsymbol{\omega}= [\omega_1,\cdots,\omega_d,\cdots,\omega_{\vert\boldsymbol{\omega} \vert}]$ and the sampled data $\xi$. 
Correspondingly, the global minimization problem for service $r$ can be defined as minimizing the aggregation of multiple local loss functions, namely,
\begin{equation}
\label{eq:global_loss_function}
	\boldsymbol{\omega}^\ast=\arg\min_{\boldsymbol{\omega }} \loss_r\left( \boldsymbol{\omega } \right) =\arg\min_{\boldsymbol{\omega }}  \sum\nolimits_{i=1}^N{\kappa\subscriptir\loss \subscriptir\left( \boldsymbol{\omega }\right)},
\end{equation}
Typically, the aggregation weight $\kappa \subscriptir={{|\mathcal{D} \subscriptir|}\big/{\sum\nolimits_{i=1}^N{|\mathcal{D} \subscriptir|}}}$. 

Considering the $t$-th global training round for service $r$, where the global model in Eq. \eqref{eq:global_loss_function} is represented as $\boldsymbol{\omega}\subscripttr$, the server distributes the model $\boldsymbol{\omega }\subscripttr$ to selected clients, who then perform $\iota$ steps' stochastic gradient descent (SGD)-based local updates to complete one full round of training. Mathematically, the local update method can be expressed as
\begin{equation}
\label{eq:local update}
	\boldsymbol{\omega }\subscriptitk=\boldsymbol{\omega }\subscriptitk[k-1]-\eta_k\tilde{\nabla} \loss\subscriptir\left( \boldsymbol{\omega }\subscriptitk[k-1] \right),
\end{equation}
where $k \in \{1, \cdots, \iota\}$, $\eta_k$ denotes a $k$-relevant learning rate, $\tilde{\nabla} \loss\subscriptir\left( \cdot \right) $ represents the stochastic gradient of the local loss function and $\boldsymbol{\omega }\subscriptitk[0]\!=\!\boldsymbol{\omega }\subscripttr$. After performing $\iota$ local updates, the client acquires a new model and uploads the quantization $\quant (\boldsymbol{\omega }\subscriptitk[\iota])$ of corresponding parameters $\boldsymbol{\omega }\subscriptitk[\iota]$ to designated servers. On the server side, consistent with Ref. \cite{fedavg}, these parameters uploaded by selected clients will be aggregated as
\begin{equation}
\label{eq:model aggretion}
	\boldsymbol{\omega }\subscripttr[t+1]=\sum\nolimits_{i=1}^N{\kappa \subscriptir\quant \left( \boldsymbol{\omega }\subscriptitk[\iota] \right)}.
\end{equation}
The client uploads the model before compressing it according to the quantization policy $\quant (\cdot )$ specified by the server. Notably, servers at different SPs can also calibrate suitable aggregation strategies, such as client selection, alongside effective resource allocation techniques, which will be discussed as follows.

\begin{table}[tb]
	\caption{Notations mainly
  used in this paper.}
	\label{tab:notations}
	\begin{tabularx}{\columnwidth}{cX}
		\toprule
		Notation & Description  \\
		\midrule
		$i \in [1,\cdots,N]$ & Index of $N$ FL clients  \\ 
		$r \in [1,\cdots,R]$ & Number of SPs; also total number of agents \\
		$T = [1,\cdots,t,\cdots]$ & FL global training round \\
		$\iota$ &  Total steps of FL local update\\
		$\mathcal{D}\subscriptir$ & Client $i$'s data set about service ${r}$ \\
		$\boldsymbol{\omega},\boldsymbol{\omega}\subscripttr,\boldsymbol{\omega }\subscriptitk[\iota]$ & FL model parameters\\
		$\loss_{{r}},\loss\subscriptir$     & Global  and local loss function of client $i$ with respect to service $r$\\
		$\vol\subscriptitr,\vol\subscripttr$   & Client $i$'s communication overheads for service $r$ in the $t$-th round; the combination of overhead $\vol\subscriptitr$ from clients within service $r$ \\
		$T^{\rm cmp}\subscriptitr$, $T^{\rm com}\subscriptitr$   &  Client $i$'s computational and communication delay at the $t$-th round with respect to service $r$\\
        $E^{\rm cmp}\subscriptitr$, $E^{\rm com}\subscriptitr$   &  Client $i$'s energy consumption at the $t$-th round with respect to service $r$\\
        $\rate\subscriptitr$ & Client $i$'s transmission rate of service $r$ in the $t$-th round\\
        $n\subscripttr$ &SP $r$'s client selection in the $t$-th round \\
		$B\subscriptitr,B\subscripttr$ & Client $i$'s bandwidth in the $t$-th round, the combination of selected clients' bandwidth\\
		$\Acc\subscripttr$ & Model accuracy of service $r$ in the $t$-th round\\
		$q\subscripttr, q\subscriptitr$   & SP $r$'s quantization decision in the $t$-th round, and the corresponding quantization level for client $i$\\
        $f\subscripttr, f\subscriptitr$   & SP $r$'s quantization decision in the $t$-th round, and the corresponding quantization level for client $i$\\
		$o\subscripttr,\selfaction$  &  Agent $r$'s observation and action in the $t$-th round\\
        $Z\subscripttr,\Theta\subscripttr$ & SP $r$'s model training status in the $t$-th round\\
		${\rm rwd}\subscripttr$   &  Reward of agent $r$ in the $t$-th round\\
		$J_r$& Cumulative expected reward of agent $r$ \\
            $M_r$ & Loss functions of agent $r$'s value network \\
		$\selfpi$,$\otherspi$,$\bestjointpi$ & Agent $r$'s policy, joint policy apart from agent $r$ and conjectured joint policy\\
		$V_r^{\bm \pi}$,$Q_r^{\bm \pi}$ & State value function and $Q$-function of agent $r$\\
		\bottomrule
	\end{tabularx}
\end{table}

\subsubsection{Quantization Process}
Following Ref. \cite{Alistarh2016QSGDRQ}, we reduce the size of data flow by statistically quantizing elements of model parameters $\boldsymbol{\omega}$ into some discrete levels, with its $d$-th element's maximum-$q$-level quantization expressed as
\begin{equation}
	\quant_q\left( \omega_d \right) =\left\| \boldsymbol{\omega} \right\|_p\cdot \mathrm{sgn}(\omega_d) \cdot \Xi_q \left(\omega_d,q \right),
\end{equation}
where $\left\| \cdot \right\|_p$ represents the $p$-norm, $\mathrm{sgn}(\cdot)$ denotes the sign function, and $\Xi_q \left(\omega_d,q \right)$ is a random mapping defined as
\begin{equation}
		\Xi_q \left(\omega_d,q \right) =
        \begin{cases}
			u/q, &  \varepsilon \leq 1-P\left( \frac{|\omega _d|}{\left\| \boldsymbol{\omega } \right\| _p},q \right)\\
			\left( u+1 \right) /q, & \rm{otherwise}\\
		\end{cases}
\end{equation}
Here, $\varepsilon$ denotes the probabilistic output of a uniformly distributed random variable. $P\left(e,q \right) =eq-u$, $0\le e\le 1$, where $u$ is any integer satisfying $u \in [0,q)$ and $\frac{u}{q}\!\leq\! \frac{|\omega _d|}{\left\| \boldsymbol{\omega } \right\| _p} \!\leq\!\frac{u+1}{q} $ \cite{Alistarh2016QSGDRQ}.
Attributed to the value quantized by $\lceil \log_2 q\rceil+1$ bits per element (including a one-bit sign), as well as a $32$-bit gradient norm, the communication overheads for a $\vert \boldsymbol{\omega} \vert$-length quantized parameter vector that client $i$ needs to upload for service $r$ during the $t$-th round is reduced from the original $32\vert \boldsymbol{\omega} \vert$\footnote{Each pre-quantization parameter $\omega_d$ is represented as $32$-bit floating-point numbers, consistent with standard deep learning frameworks.} to
\begin{equation}
\label{vol}
	\vol\subscriptitr=\vert \boldsymbol{\omega} \vert \left(\lceil \log_2  q\subscriptitr \rceil+1\right)+32,
\end{equation}
where $q\subscriptitr$ is the quantization level chosen by client $i$ of the service $r$ in the $t$-th round. 
Therefore, the total communication overheads for a single service is $\vol\subscripttr=\sum_{i\in \mathcal{N}}\vol\subscriptitr$.

\subsubsection{Communication Latency and Energy Consumption}
In wireless network scenarios, FL training latency and total energy consumption primarily arise during computation and transmission of local model updates.
The computational energy consumption for client $i$ \cite{liu2024fedeco} during the $t$-th round training for service $r$ can be expressed as
\begin{equation}
    \label{eq:client Ecmp}
	E\subscriptitr^{\rm cmp}=\mu_i \cyc\subscriptir \vert \mathcal{D}\subscriptir\vert f\subscriptitr^{2},
\end{equation}
where $\mu_i$ is the effective capacitance constant determined by the chip architecture, $f\subscriptitr$ denotes the CPU cycle frequency of client $i$ in the $t$-th round, and $\cyc\subscriptir$ represents the number of CPU cycles required to execute one sample for service $r$ on client $i$. Concurrently, the incurred local computation latency can be written as
\begin{equation}
    \label{eq:client Tcmp}
	 T\subscriptitr^{\rm cmp}=\cyc\subscriptir \vert \mathcal{D}\subscriptir\vert/{f\subscriptitr}.
\end{equation}

On the other hand, in the model parameter transmission phase, we consider frequency-division multiple access (FDMA) technology for communication between clients and SPs. Based on the Shannon formula, the transmission rate of client $i$ can be denoted as
\begin{equation}
	\rate\subscriptitr=B\subscriptitr\log_2\left( 1+\frac{g\subscriptit p\subscriptit}{B\subscriptitr N_0} \right) ,
\end{equation}	
where $B\subscriptitr$ represent the service $r$-related bandwidth allocated to client $i$ in the $t$-th global training round, $g\subscriptit$ and $p\subscriptit$ denote the channel gain and transmission power corresponding to client $i$, respectively, and $N_{0}$ is the single-sided white noise power spectral density. Therefore, the communication latency and transmission energy consumption of client $i$ can be calculated as
\begin{equation}
    \label{eq:client Tcom}
	T\subscriptitr^{\rm com}=\frac{\vol\subscriptitr}{\rate\subscriptitr},
\end{equation}
\begin{equation}
\label{eq:client Ecom}
    E\subscriptitr^{\rm com}=T\subscriptitr^{\rm com}p\subscriptit=\frac{\vol\subscriptitr p\subscriptit}{\rate\subscriptitr}.
\end{equation}

In summary, the total energy consumption and total latency of service $r$ in the $t$-th round are expressed as follows
\begin{subequations}
\label{eq:total totalTE}
    \begin{equation}
 	E\subscripttr^{\rm total}=\sum\nolimits_{i\in \mathcal{N}}{\left( E\subscriptitr^{\rm com}+E\subscriptitr^{\rm cmp} \right)},
 	\end{equation}
    \begin{equation}
 	T\subscripttr^{\rm total}=\max _{i\in \mathcal{N}}\left( T\subscriptitr^{\rm cmp}+T\subscriptitr^{\rm com} \right).
    \end{equation}
\end{subequations}

\subsection{Problem Formulation}
\label{section: prob_formulation}
Based on the communication and computation models above, each SP $r$ independently aims to maximize its FL performance by optimizing client selection $n\subscripttr$, quantization $q\subscripttr$, and resource allocation ($B\subscripttr = \sum_{i\in\mathcal{N}} B\subscriptitr$, $f\subscripttr$
 ), balancing model accuracy, communication efficiency, and energy consumption over a finite horizon. This objective is formalized for each SP as
{
\begin{align}
\label{eq:MDP}
	\min _{\bm f, \bm n, \bm q, \bm B} & \sum\nolimits_{t=0}^{T-1} \gamma^t\Upsilon \left( \loss_r(\boldsymbol{\omega}\subscripttr), \vol\subscripttr, E\subscripttr^{\rm {total }},T\subscripttr^{\rm {total }}\right) \\
	\text { s.t. } \quad & \boldsymbol{C 1} \quad E\subscripttr^{\rm {total }} \leq E^{\max }_r, T\subscripttr^{\rm {total }} \leq T^{\max }_r, \forall t, r \notag\\
	& \boldsymbol{C 2} \quad n\subscripttr \in [1, N], \forall t, r \notag\\
	& \boldsymbol{C 3} \quad f^{\rm {min }} \leq f\subscripttr \leq f^{\max },   \forall t, r \notag\\
	& \boldsymbol{C 4} \quad B^{\rm {min}} \leq \sum\nolimits_r B\subscripttr \leq B^{\rm {max }}, \forall t \notag\\
	& \boldsymbol{C 5} \quad q\subscripttr \in[q^{\rm min},q^{\rm max}], \forall t, r\notag
\end{align}
\noindent Here, $\gamma$ denotes a discount factor while the monotonically decreasing function $\Upsilon \left( \loss_r(\boldsymbol{\omega}\subscripttr), \vol\subscripttr, E\subscripttr^{\rm {total }}, T\subscripttr^{\rm {total }}\right)$ weights the components therein, whose exact format will be discussed subsequently. 
Constraints $\boldsymbol{C 1}$, $\boldsymbol{C 3}$, and $\boldsymbol{C 4}$ impose limits on per-service energy consumption $E^{\max }_r$, total latency $T^{\max }_r$, computing capabilities $[f^{\min},f^{\max}]$, and the overall bandwidth budget $[B^{\min},B^{\max}]$ respectively, while $\boldsymbol{C 2}$ and $\boldsymbol{C 5}$ define the permissible ranges for client selection and quantization level. 
Implicitly, the shared constraint like $\boldsymbol{C 4}$ underscores the inherent non-cooperative competition among SPs, where each seeks to maximize its individual resource acquisition for superior performance. Consequently, the systemic objective is to judiciously balance individual service quality with the achievement of a global equilibrium.

We reformulate Eq. \eqref{eq:MDP} as a Markov decision process (MDP) for iterative optimization. To resolve the non-cooperative dynamics and achieve a robust system-wide equilibrium, we propose a game-theoretic Pareto actor-critic framework with expectile regression. This approach enables each SP to independently learn its optimal policy by conjecturing the joint policies of others, and accounting for heterogeneous risk preferences via expectile regression, ultimately guiding the system towards a global Pareto-optimal equilibrium.

\section{\texttt{PAC-MCoFL} for Communication and Computation Co-optimization}
\label{section3}
\subsection{Markov Decision Process}
\label{sec:Markov Decision Process}
We formulate the communication and computation co-optimization problem for FL with multi-SP as a sequential decision-making problem, modeled as an MDP. This MDP is represented by a quintuple $\left( \mathcal{O} ,\mathcal{A},{\rm rwd},\Lambda,\gamma \right) $, where $\mathcal{O}$ and $\mathcal{A}$ denotes the observation and action space, while ${\rm rwd} $ is the reward function, obtained by mapping the current time state and action to a reward value. $\Lambda$ indicates the state transition probability function, which captures the dynamics of the environment, and $\gamma$ is a discount factor. The specific definitions of the observation, action, and reward are as follows.
\begin{itemize}
\item  \textbf{Observation\ Space} 
As shown in Fig. \ref{workflow}, we use the terminology \emph{agent} to denote the intelligent decision-maker deployed at each SP. Consistent with the black solid line in Fig. \ref{workflow}, the observation of agent $r$ after the $t$-th FL round can be represented as 
\begin{equation}
\label{observation}
	o\subscripttr\!=\!\left\{ Z\subscripttr(\boldsymbol{\omega }),\Theta\subscripttr,\boldsymbol{B}_t \right\},
\end{equation}
where $Z\subscripttr(\boldsymbol{\omega }) \!=\! \left\{ t,\loss_r(\boldsymbol{\omega}\subscripttr), \Acc\subscripttr, q\subscripttr \right\}$ records the round time, loss, model accuracy and quantization level of FL training, 
$\Theta\subscripttr \!=\! \left\{ T\subscripttr^{\rm total}, E\subscripttr^{\rm total},\vol\subscripttr \right\}$ 
encompasses the current FL model training status, and 
$\boldsymbol{B}_t \!=\! \left\{B\subscripttr \right\}_{r \in \mathcal{R}}$ 
represents the bandwidth allocation for all SPs. This setting has practical significance. On one hand, since different SPs often resist publicly disclosing the specific operational details, the first two elements, $\Theta\subscripttr$ and $Z\subscripttr$, contain internally available information only. On the other hand, to establish a game-theoretic relationship among agents in MARL, we deem the bandwidth allocation information as public information that can be acquired from the network operator \cite{3gpp_ts23501_2020}.
\item  \textbf{Action\ Space} As shown by the brown dashed line in Fig. \ref{workflow}, the action space, comprising all decision variables in Eq. \eqref{eq:MDP}, can be written as 
\begin{equation}
	\selfaction \! = \!\left\{ n\subscripttr,f\subscripttr,B\subscripttr,q\subscripttr \right\} \in \mathcal{A},
\end{equation}
on which the underlying clients further determine their behavior. Specifically, the bandwidth $B \subscripttr$ is assumed to be equally allocated among the selected clients within each service, and the CPU frequency $f\subscriptitr$ and quantization level $q\subscriptitr$ for each client will fluctuate slightly under a given action\cite{rw10,FedDQ}, i.e., $f\subscriptitr\sim G \left( f\subscripttr,\Sigma _f \right), q\subscriptitr\sim G \left( q\subscripttr,\Sigma _s \right)$, $G$ is the Gaussian distribution. In addition, we denote $\jointaction = [a_{1,t},\cdots,a_{R,t}]$, and we assume the actions from other agents can be acquired from network operator and participated clients.

\item \textbf{Reward\ Function}
To match the objective function $\Upsilon$ in Eq. \eqref{eq:MDP}, in line with the purple dotted line in Fig. \ref{workflow}, we define the reward function as
\begin{equation}
\label{reward}
{\rm rwd}\subscripttr  \!=\! 
\sigma_1 \Acc\subscripttr
\!+\!\sigma_2 \varPhi\subscripttr(\boldsymbol{q})
\!-\!\sigma_3 E\subscripttr^{ \rm total}
\!-\! \sigma_4 T\subscripttr^{\rm total},
\end{equation}
where $\Gamma\subscripttr$ denotes test accuracy, the weights $\sigma_j$, $j\in \{1,\cdots, 4\}$, are positive constants, and the negative sign transforms the minimization problem of latency and energy consumption into a reward maximization problem. The reward function captures immediate performance metrics and actions. Through a single composite utility, the QoS of an SP, one type of highly sensitive information, has not been disclosed explicitly, reinforcing the non-cooperative nature of the problem. Besides, an adversarial factor $\varPhi$, which  characterizes the quantization policy game across SPs to compete for communication resources, is defined as 
\begin{equation}
	\varPhi\subscripttr(\boldsymbol{q}) \! = \!\frac{n\subscripttr \ q\subscripttr}{\epsilon \times\vol\subscripttr+\sum_{j\in \mathcal{R} /\left\{ r \right\}} {n_{j,t}\  q_{j,t}}},
\end{equation}
where $\epsilon$ is a constant. Although FL convergence is inherently dependent on sequential state transitions, the discounted cumulative reward can propagate long-term dependencies via value function updates, enabling a tractable approximation for real-time co-optimization.
Notably, when each agent makes the decision, it does not know the action concurrently taken by other agents. After each agent finally executes a determined action, the part $\sum_{j\in \mathcal{R} /\left\{ r \right\}} {n_{j,t}\  q_{j,t}}$ can be known through the network operator's control plane, which is consistent with 3GPP-defined network exposure functions \cite{3gpp2017study}. The communication overhead for action sharing is negligible compared to the transmission of FL model updates, as shown in Appendix D-A.
\end{itemize}
With this formulated MDP, each SP executes an action based on its perceived environment, guided by a game-theoretic policy, and subsequently receives a corresponding reward. We will next discuss how to resort to the PAC algorithm to obtain a learned policy $\pi_r$ parameterized by $\phi$, which determines the action $a\subscripttr$ based on certain observations and strives to maintain service quality while minimizing communication and computation costs.
\begin{figure}[tbp]
	\centerline{\includegraphics[width=\linewidth]{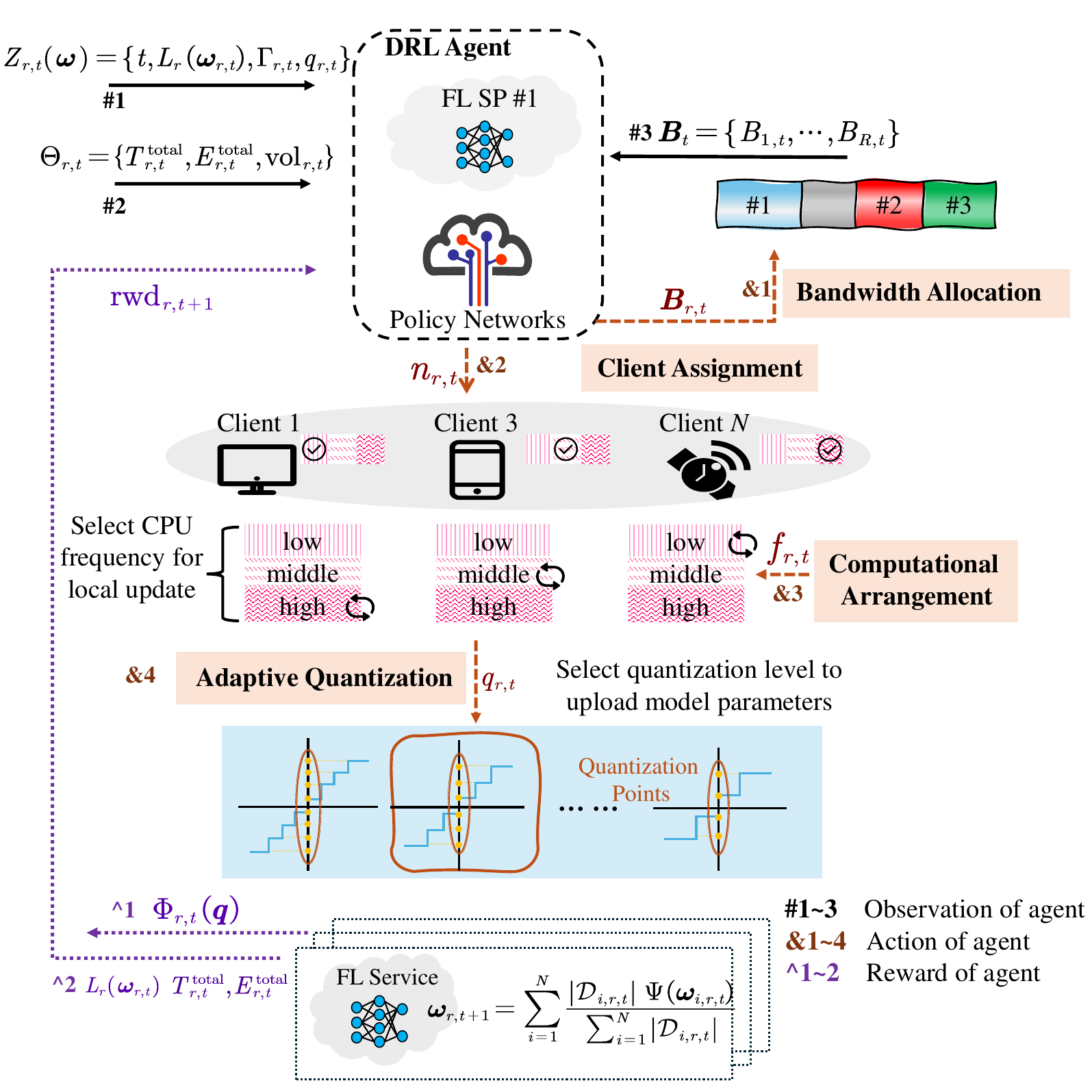}}
	\caption{The MDP diagram for a single FL service provider.}
	\label{workflow}
\end{figure}
\subsection{Expectile Regression-Based PAC for Multi-SP Game with Heterogeneous Risk Modeling}
\begin{algorithm}[bp]
	\caption{Training process of the \texttt{PAC-MCoFL} algorithm from the perspective of agent (i.e., SP) $r$.}
	\label{alg:algorithm1}
	\KwIn{Total episodes ${\rm num}_{\max}$; initialized parameters of the actor $\phi_r$ and critic $\theta_r$ of the agent $r$; FL global model $\bm \omega_r$, FL global training rounds $T$; total client set $\mathcal{N}$. }
	\KwOut{Trained parameters of the actor $\phi_r$ and critic $\theta_r$ of the agent $r$.}  
	\BlankLine
	Load the training dataset of FL service; ${\rm num}_{\rm eps}\leftarrow 0$;
 
	\While{each episode ${\rm num}_{\rm eps} \le {\rm num}_{\max}$}{
		Initialize global model parameters $\boldsymbol{\omega}\subscripttr[0]$ and client features, e.g., $|\mathcal{D}\subscriptir|,o\subscripttr[0],f\subscripttr[0]$;

		\For{each round $t=0,1,\cdots,T-1$ in}{
            Get observation $o\subscripttr$ and $o\subscriptotheragent$;
            
            Sample action delta $\{\psi_m\}_{m=1}^4$ from actor network $\selfpi$, where $\psi_m \in \{-1,0,1\}$;
            
            Construct full action $\selfaction\!=\!\{ n\subscripttr,f\subscripttr,B\subscripttr,q\subscripttr\}$ by TCAD update rule from Eq. \eqref{eq:TCAD2};

            \For{each selected client $i \in \mathcal{N}$ }{
                Local update by Eq. \eqref{eq:local update} and get $\boldsymbol{\omega}\subscriptitr$;

                Obtain $E^{\rm cmp}_{i,t,r}$, $T^{\rm cmp}_{i,t,r}$ by Eq. \eqref{eq:client Ecmp} and \eqref{eq:client Tcmp};
                
                Obtain $E^{\rm com}_{i,t,r}$, $T^{\rm com}_{i,t,r}$ by Eq. \eqref{eq:client Ecom} and \eqref{eq:client Tcom};
              }
            Global update ${ \boldsymbol{\omega}\subscripttr[t+1]=\!\sum_{i=1}^N \!\frac{\left|\mathcal{D}\subscriptitr\right| \quant\left(\boldsymbol{\omega}\subscriptitr\right)}{\sum_{i=1}^N\left|\mathcal{D}\subscriptitr\right|}}$;

            Distributes global model $\boldsymbol{\omega}\subscripttr[t+1]$ to all clients;
            
            Obtain $\Acc\subscripttr$ through model test;

            Obtain $E^{\rm total}_{t,r}$, $T^{\rm total}_{t,r}$ by Eq. \eqref{eq:total totalTE};

            Get $o\subscripttr[t+1],o_{-r,t+1}$ by Eq. \eqref{observation};
            
            Compute ${\rm rwd}\subscripttr$ by Eq. \eqref{reward};

            Put $(o\subscripttr,\selfaction,{\rm rwd}\subscripttr,o\subscripttr[t+1])$ into an episode batch.
		}

        Insert the episode batch into a replay buffer $H$;

        Clear the episode batch.
        
        \If{$H$ reaches the buffer size for training}{
        Sample a single batch from $H$;
        

        Obtain Conjectured joint policy for other agents $\bestotherspi = \arg \max_{\otheraction} Q\subscriprbestpi(o\subscripttr,\selfaction,\otheraction), \forall t$;

        
        //\textit{Note: For the scalable \texttt{PAC-MCOFL-p}, this brute-force search is replaced by the parameterized generator as Eq. \eqref{eq:PCG}};
        
        Update critic parameters $\theta_r$ by performing a gradient step to minimize the expectile loss $M_r^{(\tau)}$ from Eq \eqref{eq:critic};

        Update actor parameters $\phi_r$ by ascending the policy gradient $\nabla_{\phi_r} J_r$ from Eq. \eqref{eq:policy_gradient}, which is conditioned on the conjectured policy $\bestotherspi$;}

        ${\rm num}_{\rm eps} \longleftarrow {\rm num}_{\rm eps} + 1$.

	}
\end{algorithm}
In this subsection, we present the algorithmic framework of \texttt{PAC-MCoFL}. We first define the joint policy $\jointpi \!=\! (\selfpi,\otherspi)$ where $\otherspi$ represents the set of policies from all other agents (denoted as $-r$). In addition, there is a finite time horizon structure for RL in \texttt{PAC-MCoFL}. For each SP $r$, we denotes the cumulative expected reward under joint policy $\jointpi$ as
\begin{equation}
\label{eq:cumulative retrun}
J_r\left( \jointpi \right) \!=\!\mathbb{E}_{o\subscripttr[0]} \left[ V\subscriptrpi(o\subscripttr[0])\right] \!=\! \mathbb{E}_{o\subscripttr[0]} \left[\sum\limits_{t=0}^{T-1}\gamma^t \mathbb{E}_{\jointpi} \left[ { {\rm rwd}\subscripttr}|o\subscripttr[0],\jointpi\right]\right] ,
\end{equation}
where $V\subscriptrpi$ is state value function of agent $r$ under joint policy $\jointpi$, and $\gamma \in \lbrack 0,1)$ denotes the discount factor. Or equivalently, 
\begin{equation}
    J_r\left( \jointpi \right) =\sum\nolimits_{o\subscripttr}{d^{\jointpi}(o\subscripttr)\left[ V\subscriptrpi(o\subscripttr) \right]},
\end{equation}
where $d^{\jointpi}(o)$ is the transfer probability from initial observation $o\subscripttr[0]$ to observation $o$ in the smooth distribution of the MDP under $\jointpi$.
Besides, the $Q$-function can be expressed as $Q\subscriptrpi(o\subscripttr,\jointaction) \!=\! {\rm rwd}\subscripttr(o\subscripttr,\jointaction) \!+\! \gamma \mathbb{E}[V\subscriptrpi(o\subscripttr[t+1])]$.

A Pareto-optimal equilibrium \cite{paretoac} is a state where no agent can enhance its expected return without reducing the expected return of another agent. Drawing inspiration from the prisoner's dilemma \cite{intro19}, if agents can coordinate effectively and trust that neither will deviate from the equilibrium, the resulting joint policy not only maximizes the expected return for an individual agent but also for all agents involved.
In other words, for agent $r$, if the other agents adhere $\bestotherspi$ maximizing $J_r\left( \selfpi,\otherspi \right)$ as
\begin{equation}
    \bestotherspi = \mathop {\arg\max}\nolimits_{\otherspi}J_r\left( \selfpi,\otherspi \right),
\end{equation}
it is feasible for the PAC-based agent to learn a policy $\selfpi$ that maximises $J_r\left( \selfpi,\bestotherspi \right)$. This process culminates in a joint policy $\bestjointpi \!=\!(\selfpi,\!\bestotherspi)$ which maximizes the cumulative rewards of all agents and eventually replaces NE with Pareto optimal equilibrium \cite{paretoac} as 
\begin{equation}
    \selfpi = \mathop{\arg\max}\nolimits_{\selfpi}J_r\left( \selfpi,\bestotherspi \right).
\end{equation}
Therefore, the difficulty turns to be conjecturing a joint policy $\bestotherspi$, which in practice, can be approximated by evaluating all potential joint actions of the other agents (i.e. all $\otheraction \in \mathcal{A} _{-r}, \mathcal{A} _{-r}=\bigcup_{j\ne r}\mathcal{A}_j$) on the joint $Q$-value and taking the maximum operator, formally,
\begin{equation}
\label{eq:bestotherspi}
    \bestotherspi = \mathop {\arg\max}\nolimits_{\otheraction}Q\subscriprbestpi\left( o\subscripttr,\selfaction,\otheraction \right) .
\end{equation}
\begin{figure*}
\begin{equation}
        \label{eq:critic opt_obj original}
        M_r(\jointpi) \triangleq \mathbb{E}_{\selfaction, a\subscripttr[t+1] \sim \selfpi, \otheraction \sim \otherspi} 
        \Big[\Big( 
        \underbrace{{\rm rwd}\subscripttr+\gamma \max _{\otheraction} Q\subscriprbestpi\left(o\subscripttr[t+1], a\subscripttr[t+1], \otheraction\right)-Q\subscriprbestpi\left(o\subscripttr, \selfaction, \otheraction\right)}_{\text{TD error term}\ \delta}
        \Big)^2 \Big]
    \end{equation}
    \begin{equation}
        \label{eq:J}
        \begin{aligned}
            \nabla_{\phi_r} J_r\left(\jointpi\right) 
            &= \sum_{o\subscripttr} d^{\jointpi}(o\subscripttr)  \sum_{\selfaction,\otheraction} \nabla_{\phi_r} \pi^{\dagger}(\jointaction  \mid  o\subscripttr) Q\subscriprbestpi (o\subscripttr, \selfaction,\otheraction) \\
             &= \sum_{o\subscripttr} d^{\jointpi}(o\subscripttr)  \sum_{\selfaction,\otheraction} \bestotherspi\left(\otheraction  \mid  o\subscripttr, \selfaction\right) \nabla_{\phi_r}\selfpi\left(\selfaction  \mid  o\subscripttr\right)  Q\subscriprbestpi \left(o\subscripttr, \selfaction, \otheraction\right) \\
            &= \sum_{o\subscripttr} d^{\jointpi}(o\subscripttr)  \sum_{\selfaction,\otheraction} \selfpi\left(\selfaction  \mid  o\subscripttr\right) \bestotherspi\left(\otheraction  \mid  o\subscripttr, \selfaction\right) \cdot\nabla_{\phi_r} \log \selfpi\left(\selfaction  \mid  o\subscripttr\right) Q\subscriprbestpi \left(o\subscripttr, \selfaction, \otheraction\right) \\
        \end{aligned}
    \end{equation}
    \hrulefill
\end{figure*}

In classic RL, the loss function for $Q$-function updates is defined by the temporal difference (TD) error in Eq. \eqref{eq:critic opt_obj original}. Nevertheless, considering heterogeneous risk preferences among SPs, such as spanning conservative to aggressive policies, we employ expectile regression \cite{Kostrikov2021OfflineRL} for critic loss computation
 \begin{equation}
        \label{eq:critic opt_obj}
        M_r^{\rm (\tau)}(\jointpi) \triangleq \mathbb{E}_{\selfaction, a\subscripttr[t+1] \sim \selfpi, \otheraction \sim \otherspi} 
        \left[\tau \cdot\delta^2_{+} + \left(1-\tau\right) \cdot\delta^2_{-}\right],
    \end{equation}
where $\tau \in (0,1)$ is adjustable expectile factor controlling sensitivity to overestimation ($\tau < 0.5$, aggressive) or underestimation ($\tau > 0.5$, conservative); $\delta_+$ and $\delta_-$ denote the positive and negative components of $\delta$, respectively. By tuning $\tau$, the critic flexibly characterizes the differential risk preferences of SPs toward policy decisions.
Accordingly the critic is updated by
\begin{equation}
    \label{eq:critic}
     \theta_r = \theta_r - \alpha \nabla_{\theta_r} M_r^{\rm (\tau)}(\jointpi),
\end{equation}
where $\alpha$ denotes the learning rate for updates. 
On the other hand, to maximize the expected discounted cumulative reward, the policy network can be updated by
\begin{equation}
    \label{eq:policy_gradient}
     \phi_r = \phi_r + \lr \nabla_{\phi_r} J_r(\jointpi),
\end{equation}
where the policy gradient can be computed by taking the derivative of $\phi$ as shown in Eq. \eqref{eq:J} with a learning rate $\lr$. As a side note, a baseline is commonly subtracted from $J_r(\jointpi)$ without altering the gradient computation but reducing the bias of the gradient estimate \cite{MAPPO, MASAC}. We can compute this baseline through $\sum_{\selfaction}\selfpi(\selfaction|o\subscripttr;\phi_r)Q\subscriprbestpi\left(o\subscripttr, \selfaction, \otheraction\right)$ to avoid training an additional state value network.

Overall, compared to the existing architecture \cite{MAPPO,MASAC}, \texttt{PAC-MCoFL} benefits from PAC, which conjectures other agents' policies, to facilitate the determination of co-optimization strategies, including quantization levels, resource allocation, and client assignment. Therefore, \texttt{PAC-MCoFL} has the advantage of preventing the game from easily falling into suboptimal solutions.
In summary, the pseudocode for \texttt{PAC-MCoFL} is summarized in the Algorithm \ref{alg:algorithm1}.
\subsection{TCAD for Dimension Reduction}
\label{sec:TCAD}
The inherent complexity of multi-dimensional action spaces in non-cooperative multi-SP FL poses significant challenges for conventional MARL frameworks, due to the curse of dimensionality \cite{10487895}. Additionally, independent action selection often result in coordination blindness, failing to capture critical interdependencies \cite{CSBWA}. Therefore, we propose TCAD, a hierarchical action decomposition mechanism enabling efficient exploration in high-dimensional spaces while preserving cross-variable dependencies.

Let the hybrid action vector of SP $r$ be defined as $\selfaction =\left\{ n\subscripttr,f\subscripttr,B\subscripttr,q\subscripttr \right\} $. We establish a ternary projection operator $\mathcal{T}_{\rm TCAD}$ and the action space decomposition is formalized through
\begin{equation}
    \label{eq:TCAD}
    \mathcal{T}_{\rm TCAD}:a\subscripttr \rightarrow \{\varsigma_m \cdot \psi_m\}^4_{m=1},\quad \psi_m=\left\{ -1,0,1 \right\},
\end{equation}
where $m$ represents the dimensionality, $\varsigma_m$ denotes the step-size granularity, which is a pre-defined hyperparameter tailored to each variable's physical constraints (see Appendix C-C for related hyperparameter evaluation). 
The discrete-continuous hybrid update rules are unified as
\begin{equation}
    \label{eq:TCAD2}
    a\subscripttr[t+1]^{\prime(m)}={\rm Proj}_{\mathcal{C}_m}(a\subscripttr^{\prime(m)} + \psi_m \cdot \varsigma_m),
\end{equation}
where ${\rm Proj}_{\mathcal{C}_m}(\cdot)$ denotes the projection operator enforcing domain constraints 
\begin{equation}
\mathcal{C}_m= 
\begin{cases}\{1, \ldots, N\} & m=n \\
{\left[f^{\min }, f^{\max }\right]} & m=f \\
{\left[B^{\min }, B^{\max }\right]} & m=B \\
{\mathbb{Z}} \cap\left[q^{\min }, q^{\max }\right] & m=q
\end{cases}
\end{equation}
The Cartesian reconstruction $\mathcal{A}^{\prime}=\prod_{m=1}^{4}{\{-1,0,1\}}$ reduces the joint action space cardinality from $O\left(N\cdot\frac{f^{\max}-f^{\min}}{\varsigma_f}\cdot\frac{B^{\max}-B^{\min}}{\varsigma_B}\cdot q^{\max}\right)$ to constant complexity, facilitating tractable computation of Eq. \eqref{eq:bestotherspi} and enabling fine-grained multi-dimensional control. The computational efficacy analysis of TCAD can be found in Appendix C-B.
\subsection{Parameterized Conjecture Generator}
The original conjecture mechanism in Eq. \eqref{eq:bestotherspi} requires exhaustive search over opponents' joint action space $\mathcal{A}_{-r}$, incurring computational complexity of $O(|\mathcal{A}|^{R-1})$ for $R$ SPs. To overcome this combinatorial bottleneck, we introduce \texttt{PAC-MCoFL-p}, a variant of \texttt{PAC-MCoFL} with an extra, $\varphi_r$-parameterized generator ${\rm Gen}_{\varphi_r}(\cdot)$ for policy conjecture.

Specifically, \texttt{PAC-MCoFL-p} directly produces optimal joint actions of other agents through 
\begin{equation}
    \label{eq:PCG}
    \bestgenotherspi \triangleq {\rm softmax}\left({\rm Gen}_{\varphi_r}(a\subscripttr,o\subscripttr,h\subscriptotheragent)\right),
\end{equation}
where $h\subscriptotheragent$ is the hidden state derived from $o\subscriptotheragent$. The generator is optimized through a compound loss function in Eq. \eqref{eq:loss conj}, where $\chi$ is constant coefficient, $D_{\rm KL}$ denotes KL-divergency, $\pi^{\dagger,\rm{tar}}_{-r}$ represents moving average of other agents' empirical policy distributions. This formulation ensures the conjectured actions maintain temporal consistency with historically optimal behavior while maximizing expected return. We enable end-to-end training of the generator by incorporating the aforementioned loss into the policy gradient in Eq. \eqref{eq:policy_gradient}. The theoretical justification for this approach is provided by Theorem \ref{thm:PCG}.
\begin{theorem}
\label{thm:PCG}
    Consider the scenario where the true optimal joint action distribution, denoted as $\bestotherspi$ (from Eq. \eqref{eq:bestotherspi}), is approximated by a parameterized ${\rm Gen}_{\varphi_r}$ yielding distribution $\bestgenotherspi$. Under Lipschitz continuity of $Q$-function given in Assumption 4, the expected $Q$-value error for the parameterized policy satisfies
     \begin{equation}
         \label{eq:Q error}
         \lvert \mathbb{E}_{\bestgenotherspi}\left[ Q\subscriprbestpi \right] - \max_{a\subscriptotheragent}Q\subscriprbestpi \rvert \leq C \cdot \sqrt{D_{\rm KL}\left( \bestgenotherspi || \bestotherspi\right)},
     \end{equation}
      where $C$ denotes the Lipschitz constant of the $Q$-function. 
\end{theorem}
We leave Assumption 4 and the proof in Appendix A-A. This theorem shows that \texttt{PAC-MCOFL-p} is not merely a heuristic but a principled approximation with a provably bounded error. This justifies its application to large-agent scenarios where an exhaustive search is intractable, a finding further corroborated by our experiments.}

\subsection{Proof of Convergence}
\label{sec:proof_convergence}
\begin{figure*}[tbp]
    \begin{equation}
    \label{update}Q\subscripttr[t+1]\left(o\subscripttr,\jointaction\right)=
    Q\subscripttr\left(o\subscripttr,\jointaction\right)+ \alpha\left({\rm rwd}\subscripttr+\gamma \mathbb{E}_{o\subscripttr[t+1], a\subscripttr[t+1]}\left[\max _{\otheraction} Q\subscripttr\left(o\subscripttr[t+1], a\subscripttr[t+1], \otheraction\right)\right]
    -Q\subscripttr\left(o\subscripttr,\jointaction\right) \right)
    \end{equation}
    \begin{equation}
        \label{Pareto-alo}
            \mathscr{H}^p Q\subscripttr \left(o\subscripttr,\jointaction\right)=\mathbb{E}_{o\subscripttr[t+1] \sim \Lambda\left(\cdot \mid o\subscripttr,\jointaction\right)}\left[ {\rm rwd}\subscripttr +\gamma\max_{\otheraction} Q\subscripttr\left(o\subscripttr[t+1], \selfaction, \otheraction\right)\right]
    \end{equation}
    \begin{equation}
    \label{eq:loss conj}
    \Phi(\varphi_{r}) = 
    \mathbb{E}_{\tilde{a}\subscriptotheragent\sim \bestgenotherspi} \left[ -Q_r^{\bestjointpi}(o\subscripttr,a\subscripttr,\tilde{a}\subscriptotheragent)\right] + \chi \mathbb{E}\left[ D_{\rm KL}\left( \bestgenotherspi || \pi^{\dagger,\rm{tar}}_{-r}\right)\right]  
    \end{equation}
\begin{equation}
\label{eq:expectile pareto operator}
    \mathscr{H}^{p,\tau} Q\subscripttr \left(o\subscripttr,\jointaction\right)=\mathbb{E}_{o\subscripttr[t+1] \sim \Lambda}\left[ {\rm rwd}\subscripttr +\gamma\tau\max_{\otheraction} Q^+\subscripttr\left(o\subscripttr[t+1], \selfaction, \otheraction\right)+\gamma(1-\tau)\max_{\otheraction} Q\subscripttr^-\left(o\subscripttr[t+1], \selfaction, \otheraction\right)\right]
\end{equation}
    \hrulefill
\end{figure*}
We first consider the case without expectile regression, where the critic loss is given by Eq. \eqref{eq:critic opt_obj original}. Afterward, we will extend the result to the variant with expectile regression. In the former case, the update rule for the weight adjustment in Eq. \eqref{eq:critic} is derived from an iterative update\cite{mean-field2}, denoted as Eq. \eqref{update}, on which the action value function $Q\subscripttr$ is iterated\footnote{In the rest of the paper, the superscript $\jointpi$ for $Q$ and $V$ will be omitted for simplicity of representation.}.
The update of the $Q$-function through PAC, specifically the inclusion of additional, conjectured action inputs from other agents based on the max operation, results in a significantly expanded exploration space and compromises existing conclusions. Consequently, the convergence dynamics that are missing in the classical PAC framework \cite{paretoac} are worthy of further investigation. 
To prove that the joint $Q$-function $\boldsymbol{Q}_{t}=\left[ Q\subscriptproofrt[1],\cdots,Q\subscriptproofrt[R] \right] $ converges gradually to Nash $Q$-value $\boldsymbol{Q}^{\ast}=\left[ Q\subscriptrstar[1],\cdots,Q\subscriptrstar[R] \right] $, we first define a Pareto operator as Eq. \eqref{Pareto-alo} on a complete metric space, as shown in Lemma \ref{lemma1}. 
\begin{lemma}
\label{lemma1}
	Considering any two action value functions $Q_1,Q_2 \in \mathbb{R}^{|\mathcal{O}||\mathcal{A}|}$ of a single agent, for Pareto operator in Eq. \eqref{Pareto-alo}, the following contraction mapping holds
	\begin{equation}
	\| \mathscr{H}^PQ_1-\mathscr{H}^PQ_2 \|_\infty \leq \gamma \| Q_1-Q_2\|_\infty,
	\end{equation}
    where $\| \cdot \|_\infty$ represents the supremum norm, $\gamma \in [0,1)$. Besides, Pareto operator $\mathscr{H}^P$, which constitutes a contraction mapping, has the fixed point at $\boldsymbol{Q}^\ast$. 
\end{lemma}
We leave the proof in Appendix A-B.

Hu \textit{et al.}\cite{nashq} define an iterative process for obtaining Nash strategies through the Lemke-Howson algorithm and introduce a Nash operator $\mathscr{H}^N$. 
\begin{definition}
	(\textbf{Nash Operator}) The expression for value function iteration is given by a Nash operator as
\begin{equation}
        \begin{aligned}
            \mathscr{H}^N \boldsymbol{Q}\left(o,\boldsymbol{a}\right)
            =&\mathbb{E}_{o_{t+1} \sim \Lambda\left(\cdot \mid o,\boldsymbol{a}\right)}\\
            &\left[{\rm rwd}\left(o,\boldsymbol{a}\right)+\gamma \boldsymbol{V}^{\text {Nash }}\left(o_{t+1}\right)\right].
        \end{aligned}
\end{equation}
\end{definition}
\begin{definition}
	(\textbf{Nash Equilibrium}) In a stochastic game, a joint policy $\jointpi^\ast \triangleq [\pi\subscriptrstar[1],\cdots,\pi\subscriptrstar[R]]$ constitutes an NE if no agent can achieve a higher return by unilaterally changing its policy. Formally, it satisfies 
	\begin{equation}
		\forall o_r \in \mathcal{O}_r, \  V_r\left(o_r ; \jointpi^\ast\right)=V_r\left(o_r ; \pi\subscriptrstar, \pi\subscriptrstar[\!-\!r]\right) \geq V_r\left(o_r ; \selfpi, \pi\subscriptrstar[\!-\!r]\right).
	\end{equation}
\end{definition}
\noindent Therefore, by the proof of Lemma \ref{lemma1}, the Nash operator constructs a contraction mapping on a complete metric space, ensuring that the $Q$-function ultimately converges to the value corresponding to the NE of the game, meaning there exists a fixed point $\mathscr{H} ^N \boldsymbol{Q}^{\ast}=\boldsymbol{Q}^{\ast}$\cite{nashq}.

Then, we find the update rule for the $Q$-function in Eq. \eqref{update} follows a stochastic approximation structure. By Theorem 1 in \cite{Jaakkola1994OnTC} and Corollary 5 in \cite{szepesvari1999unified}, 
the stochastic approximation has the following convergence properties.
\begin{lemma}
\label{lemma2}
	The random process $\left\{\Delta_t\right\}$ defined in $\mathbb{R}$ as
	$$
	\Delta_{t+1}(x)=\left(1-\alpha_t(x)\right) \Delta_t(x)+\alpha_t(x) F_t(x),
	$$
	converges to zero with probability 1 (w.p.1) when\newline
	\textbf{1)} $0 \leq \alpha_t(x) \leq 1, \sum_t \alpha_t(x)=\infty, \sum_t \alpha_t^2(x)<\infty$;\newline
	\textbf{2)} $x \in \mathscr{X}$, the set of possible states, and $|\mathscr{X}|<\infty$;\newline
	\textbf{3)} $\left\|\mathbb{E}\left[F_t(x) \mid \mathscr{F}_t\right]\right\|_W \leq \gamma\left\|\Delta_t\right\|_W+c_t$, where $\gamma \in[0,1)$ and $c_t$ converges to zero w.p.1;\newline
	\textbf{4)} $\operatorname{var}\left[F_t(x) \mid \mathscr{F}_t\right] \leq U\left(1+\left\|\Delta_t\right\|_W^2\right)$ with constant $U>0$.\newline
	Here $\mathscr{F}_t$ denotes the filtration of an increasing sequence of $\sigma$-fields including the history of processes; $\alpha_t, \Delta_t, F_t \in \mathscr{F}_t$ and $\|\cdot\|_W$ is a weighted maximum norm.
\end{lemma}
\noindent To mimic the structure of Lemma \ref{lemma2}, we can subtract $Q\subscriptrstar$ from both sides of Eq. \eqref{update} and define
\begin{equation}
	\Delta_t\left(x\right)=Q\subscripttr\left(o\subscripttr, \jointaction\right)-Q\subscriptrstar\left(o\subscripttr, \jointaction\right) ,
\end{equation}
\begin{equation}
	F_t\left(x\right)\!=\!{\rm rwd}\subscripttr\!+\!\gamma \max _{\otheraction} Q\subscripttr\left(o\subscripttr[t+1],\! a\subscripttr[t+1],\! \otheraction\right)\!-\!Q\subscriptrstar\left(o\subscripttr,\! \jointaction\right) ,
\end{equation}
where $(o,\boldsymbol{a})=x$ represents the visited observation-action pairs, and $\alpha_t(x)$ is interpreted as the learning rate. The $Q$-function is updated by each agent using only the pairs $(o\subscripttr,\jointaction)$ at the time $t$ when they are accessed. In other words, $\alpha_t(o^{\prime},\boldsymbol{a}^{\prime})=0$ for any $(o^{\prime},\boldsymbol{a}^{\prime}) \neq (o\subscripttr,\jointaction)$. 

Besides, we introduce several key assumptions.
\begin{assumption}
\label{assump:finite_state}
	The observation space $\mathcal{O}$ and action space $\mathcal{A}$ of the MDP are finite; any observation-action pair $(o,\boldsymbol{a})$ can be accessed an infinite number of times.
\end{assumption}
\begin{assumption}
\label{assump:bounded_reward}
	The reward function is independent of the next state, and the reward function is bounded.
\end{assumption}
\begin{assumption}
    \label{assump:same_ne}
	For the joint action value function $\boldsymbol{{Q}}_t$ at any stage of the game, the NE policy $\jointpi^\ast$ can only be 1) a global optimum or 2) a saddle point. Moreover, both cases share the same Nash value ($\Omega$ is the policy space), namely, 
 
    \noindent 1. 
    \begin{equation}
        \mathbb{E}_{\jointpi^\ast} \left[Q\subscripttr(o\subscripttr,\jointaction)\right] \geq \mathbb{E}_{\bm \pi}\left[Q\subscripttr(o\subscripttr,\jointaction)\right], \forall \jointpi \in \bigcup_r\Omega_r ; 
    \end{equation}
    \noindent 2. 
    \begin{equation} 
        \mathbb{E}_{\jointpi^\ast} \left[Q\subscripttr(o\subscripttr,\jointaction)\right] \geq \mathbb{E}_{\selfpi} \mathbb{E}_{\otherspi^\ast}\left[Q\subscripttr(o\subscripttr,\jointaction)\right], \forall \selfpi \in \Omega_r 
    \end{equation}
    and
    \begin{equation}
        \mathbb{E}_{\jointpi} \left[Q\subscripttr(o\subscripttr,\jointaction)\right] \leq \mathbb{E}_{\selfpi^\ast} \mathbb{E}_{\otherspi}\left[Q\subscripttr(o\subscripttr,\jointaction)\right], \forall \otherspi \in \bigcup_{j\neq r}\Omega_j.
    \end{equation}
\end{assumption}
Assumption \ref{assump:finite_state} (finite actions) is inherently satisfied by our TCAD mechanism (Sec. \ref{sec:TCAD}), which explicitly discretizes the multi-dimensional action space. Assumption \ref{assump:bounded_reward} (bounded rewards) holds because our reward function, i.e. Eq. \eqref{reward}, is a composition of physically bounded metrics, while Assumption \ref{assump:same_ne} is a standard requirement in multi‑agent equilibrium analyses \cite{nashq,mean-field2,mean-filed1}.
Finally, the convergence theorem without involving expectile regression can be summarized as follows. 
\begin{theorem}
\label{thm:convergence}
	 In a multi-agent stochastic game setting, under Assumptions \ref{assump:finite_state}, \ref{assump:bounded_reward}, and \ref{assump:same_ne}, as well as the first condition of Lemma \ref{lemma2}, the values updated iteratively through Eq. \eqref{update} will ultimately converge to an NE point, specifically the Nash Q-value $\boldsymbol{Q}^{\ast}=\left[ Q_{1}^{\ast},Q_{2}^{\ast},\cdots ,Q_{R}^{\ast} \right] $.
\end{theorem}
We leave the proof in Appendix A-C. Next, considering a Pareto operator with expectile regression defined as Eq. \eqref{eq:expectile pareto operator}, the same convergence result can be transferred accordingly.

\begin{corollary}
\label{thm:expectile}
Let the assumptions of Theorem \ref{thm:convergence} hold. Suppose each agent $r\in\mathcal{R}$ employs an expectile coefficient $\tau_r \in (0,1)$, the $Q$-function sequence generated by a modified Pareto operator maintains the almost sure convergence properties established in Theorem \ref{thm:convergence}.
\end{corollary}
We give the proof in Appendix A-D. This theoretical guarantee of convergence is empirically validated in our experiments (e.g., Fig. \ref{fig:training reward}), where the gradually stabilized reward curves confirm the practical reliability.

\section{Simulation Settings and Results}
\label{section4}
\subsection{Default Simulation Settings}
\begin{table}[tbp]
	\caption{Simulation parameters}
	\label{tab:simulation paras}
	\centering
		\begin{tabularx}{\columnwidth}{cXc}
			\toprule
			Parameters &Descriptions&Values \\
			\midrule
                $N$ &Number of clients& $5$\\
                $R$ &Number of tasks&  $3$\\
                $\rho$ &Degree of non-IID data &  $1$\\
                $\iota$ &FL local update steps&  $3$\\
                $\tau$ &Expectile factor&  $0.5$\\
                $T$ &FL global training rounds &  $35$\\
			$g\subscriptit$ &Channel gain & $\left[-63,-73\right]\ $dB\\
			$N_0$&Noise power spectral density&$[-124, -174]$ dBm/Hz\\
			$p\subscriptit$& Clients’ transmission power& $\left[10,33\right]$ dBm\\
			$\mu_i$& Effective switching capacitance constant&${10}^{-27}$ \\
                $\cyc_{i,1}, \cyc_{i,2}$&CPU cycles consumed per sample for $r_1,r_2$ &$\left[6.07,7.41\right]\times{10}^5$ \\
			$\cyc_{i,3}$&CPU cycles consumed per sample for $r_3$&$\left[1.10,1.34\right]\times{10}^8$ \\
                $\sigma_1$  & Weighting factor 1 for $r_1,r_2,r_3$& 100,\ 100,\ 100 \\
                $\sigma_2$  & Weighting factor 2 for $r_1,r_2,r_3$& 4.8,\ 31.25,\ 12.5 \\
                $\sigma_3,\sigma_4$  & Weighting factor 3 and 4 for $r_1,r_2,r_3$& 0.8,\ 25,\ 16.6 \\
                $\lr,\alpha$ &  Learning rate  &$0.001$, $0.001$\\
                $\varsigma_n,\varsigma_q,\varsigma_f,\varsigma_B,$ & TCAD Granularity &1,4,0.5,2 \\
                $\Sigma_q,\Sigma_f$ &  Jitter factor for quantization and CPU frequency &$0.25,0.5$\\
			$f^{\min},f^{\max}$ & CPU frequency range& $[0.5, 3.5]$ GHz\\
                $q^{\min},q^{\max}$ & Quantization level range& $[2, 32]$ Bits\\
			$B^{\min},B^{\max}$ &  SP's bandwidth range & $[0, 30]$ MHz\\
                
			\bottomrule
		\end{tabularx}
\end{table}
We consider an FL environment involving five clients (i.e., $N=5$), coordinated by three SPs (i.e., $R=3$). These SPs schedule the clients to train three distinct tasks with CIFAR-10 [Task $r_{1}$, high complexity/$50,000$ RGB images/$10$ classes], FashionMNIST [Task $r_{2}$, medium complexity/$70,000$ grayscale images/$10$ classes], and MNIST [Task $r_{3}$, low complexity/$60,000$ grayscale images/$10$ classes] datasets, respectively. This benchmark selection, across varying complexities, evaluates our algorithm's efficacy in a multi-SP FL setting, consistent with common/mainstream edge intelligence applications \cite{intro6,liu2024fedeco}. Experiments are performed on an Ubuntu 20.04 workstation equipped with an NVIDIA GeForce RTX 3090 GPU and AMD Ryzen 9 7950X 16-Core CPU, using Python, CUDA 12.2, and PyTorch 2.5. Accordingly, three image classification models are leveraged. For Task $r_{1}$, we apply a convolutional neural network (CNN)  with four convolutional layers, followed by three fully connected layers. The model for Task $r_{2}$ is a shallow CNN, with two convolutional layers followed by two fully connected layers. The model for Task $r_{3}$ only contains two fully connected layers. The number of model parameters for three tasks are $9,074,474$, $21,840$ and $101,770$, respectively.
The datasets are distributed among the clients in an independent and identically distributed (IID) manner (i.e., a non-IID degree of $\rho =1$), indicating each client receives data covering $100\%$ of the label categories. FL training utilizes the Adam optimizer with learning rate ${\eta_k}=0.001$, $k \in \{1,\cdots, \iota\}$, local iterations $\iota=3$, and a mini-batch size of $64$. Each simulation episode spans up to $T=35$ FL rounds.\footnote{In each round $t$, the agent observes the state $o_{r, t}$, conjectures the optimal opponent action $\pi_{-r}^{\dagger}$ that maximizes its $Q$-value, and then selects its own action $a_{r, t}$ via its policy network. Upon execution of the joint action, the environment returns a reward ${\rm rwd}_{r,t}$ and the next state $o_{r, t+1}$. This complete transition is then stored in a replay buffer for RL training.}
We use the reward function in Eq. \eqref{reward}, along with its associated metrics, to evaluate the performance of a particular SP. 
Within our algorithm, the critic network employs two fully connected layers of $64$ and $128$ neurons to estimate state–action values, while the actor network uses three fully connected layers of $64$, $128$, and $64$ neurons to map observations to actions. 
We consider the following comparison baselines: 
\begin{itemize}
    \item \emph{FedAvg}: Vanilla FL \cite{fedavg} for each SP without quantization/resource optimization.
    \item \emph{FedProx-u}: Uniform $8$-bit quantization variants of \cite{FedProx} with equal resource allocation.
    \item \emph{FedDQ-h \& AdaQuantFL-h}: Adaptive quantization policy during FL training \cite{FedDQ,Jhunjhunwala2021AdaptiveQO}, with heuristic computation and communication resource allocation rules tied to quantization levels.
    \item \emph{MAPPO}\footnote{The implementation details of MARL solutions are given in Appendix B-C.}: Resource scheduling and FL operation management based on the MAPPO \cite{MAPPO}, where no conjectured joint policy is introduced. MAPPO is specifically included as the most relevant RL competitor given its proven effectiveness in FL resource allocation \cite{10487895,FedMarl}.
    \item  \emph{RSM-MASAC}: Resource scheduling and FL operation management based on the RSM-MASAC \cite{rsmmasac}, an enhanced variant of multi-agent soft actor–critic (MASAC).
\end{itemize}
These baseline algorithms collectively cover essential optimization dimensions in multi-SP FL scenarios—including conventional and adaptive quantization, uniform and heuristic resource allocation, and MARL-based global scheduling—thereby enabling a comprehensive evaluation of the performance impact. 
Finally, main simulation parameters are summarized in Table \ref{tab:simulation paras}.

To comprehensively compare the performance in the non-cooperative game setting, besides metrics like individual and global rewards, we adopt the hypervolume indicator (HVI) \cite{hvi1,hvi2}—a standard metric in multi-objective optimization—to assess the quality of the Pareto front for per-SP rewards. Specifically, let 
\begin{equation}
    \label{eq:hvi solution}
   \mathcal{P}_{(\rm alg)} =\Big\{ \mathbf{v}^{(m)} = \bigl({\rm rwd}^{(m)}_{1},\,{\rm rwd}^{(m)}_{2},\,\ldots,\,{\rm rwd}^{(m)}_{R}\bigr) \Big\}_{m=1}^{\rm total\ runs}  
\end{equation}
denote the set of reward vectors obtained from different independent training runs by an algorithm, where ${\rm rwd}^{(m)}_{r}$ is the average reward of SP $r$ in the $m$-th run. Given a reference point $\mathbf{v}^{\rm ref}\in \mathbb{R}^R$ dominated by all solutions, the HVI is computed as
\begin{equation}
    \label{eq:hvi}
    {\rm HVI} = \varpi \Big(\bigcup_{\mathbf{v}\in \mathcal{P}_{\rm alg}}[{\rm rwd}_1,{\rm rwd}_1^{\rm ref}]\times\cdots [{\rm rwd}_R,{\rm rwd}_R^{\rm ref}]\Big),
\end{equation}
where $\varpi(\cdot)$ denotes the Lebesgue measure \cite{hvi1}, a larger HVI inidcates both closer alignment with ideal per-SP rewards and a broader spectrum of trade-offs. Additional HVI computation details are given in Appendix B-A.

\subsection{Simulation Results}

\begin{table*}[tb]
  \setlength{\tabcolsep}{3.5pt} 
  \scriptsize
  \centering
  \caption{Performance evaluation of multiple FL SPs frameworks.}
  \label{tab:perform_compared}
  \begin{threeparttable}
    \begin{tabularx}{\textwidth}{@{}l|*{8}{>{\centering\arraybackslash}X}@{}}
    \toprule
    \multirow{2}{*}{\textbf{Metric}} & \textbf{Fixed} & \textbf{Uniform} & \multicolumn{2}{c}{\textbf{Heuristic}} & \multicolumn{4}{c}{\textbf{MARL}} \\
    \cmidrule(lr){2-2} \cmidrule(lr){3-3} \cmidrule(lr){4-5} \cmidrule(lr){6-9}
    & \textbf{FedAvg} & \textbf{FedProx-u} & \textbf{FedDQ-h} & \textbf{AdaQuantFL-h} & \textbf{MAPPO} &  \textbf{RSM-MASAC} & \textbf{PAC-MCoFL} & \textbf{PAC-MCoFL-p} \\
    \midrule
    $\overline{\rm vol}_{1,t}$  & $65.33$\textcolor{gray}{$(\pm 3.86)$} & $27.22$\textcolor{gray}{$(\pm 4.32)$} & $36.3$\textcolor{gray}{$(\pm 5.87)$} & $\bm{8.97}$\textcolor{gray}{$(\pm 2.73)$} & $23.01$\textcolor{gray}{$(\pm 4.86)$} & $20.65$\textcolor{gray}{$(\pm 3.86)$} & $18.74$\textcolor{gray}{$(\pm 3.23)$} & $22.02$\textcolor{gray}{$(\pm 2.08)$} \\
    $\overline{T}_{1,t}$    & $46.81$\textcolor{gray}{$(\pm 6.06)$} & $34.44$\textcolor{gray}{$(\pm 7.22)$} & $15.82$\textcolor{gray}{$(\pm 4.21)$} & $9.23$\textcolor{gray}{$(\pm 2.36)$} & $11.62$\textcolor{gray}{$(\pm 3.06)$} & $4.94$\textcolor{gray}{$(\pm 2.34)$} & $6.62$\textcolor{gray}{$(\pm 1.26)$} & $\bm{4.26}$\textcolor{gray}{$(\pm 2.34)$} \\
    $\overline{E}_{1,t}$    & $12.62$\textcolor{gray}{$(\pm 3.41)$} & $8.01$\textcolor{gray}{$(\pm 1.21)$} & $19.17$\textcolor{gray}{$(\pm 3.22)$} & $1.84$\textcolor{gray}{$(\pm 0.14)$} & $11.28$\textcolor{gray}{$(\pm 2.02)$} & $\bm{4.12}$\textcolor{gray}{$(\pm 1.12)$} & $9.96$\textcolor{gray}{$(\pm 2.12)$} & $9.61$\textcolor{gray}{$(\pm 1.87)$} \\
    $\overline{\rm rwd}_{1,t}$  & $48.92$\textcolor{gray}{$(\pm 9.67)$} & $65.41$\textcolor{gray}{$(\pm 5.98)$} & $56.51$\textcolor{gray}{$(\pm 10.2)$} & $54.74$\textcolor{gray}{$(\pm 14.5)$} & $76.03$\textcolor{gray}{$(\pm 8.79)$} & $76.19$\textcolor{gray}{$(\pm 7.42)$} & $\bm{83.85}$\textcolor{gray}{$(\pm 9.14)$} & $79.14$\textcolor{gray}{$(\pm 11.3)$} \\
    \midrule
    $\overline{\rm vol}_{2,t}$  & $0.26$\textcolor{gray}{$(\pm 0.08)$} & $\bm{0.08}$\textcolor{gray}{$(\pm 0.03)$}  & $0.44$\textcolor{gray}{$(\pm 0.11)$} & $0.22$\textcolor{gray}{$(\pm 0.09)$} & $0.23$\textcolor{gray}{$(\pm 0.04)$} & $0.08$\textcolor{gray}{$(\pm 0.04)$} & $0.15$\textcolor{gray}{$(\pm 0.04)$} & $0.11$\textcolor{gray}{$(\pm 0.03)$} \\
    $\overline{T}_{2,t}$    & $0.27$\textcolor{gray}{$(\pm 0.11)$} & $0.15$\textcolor{gray}{$(\pm 0.09)$} & $0.57$\textcolor{gray}{$(\pm 0.13)$} & $0.17$\textcolor{gray}{$(\pm 0.05)$} & $0.07$\textcolor{gray}{$(\pm 0.03)$} & $\bm{0.05}$\textcolor{gray}{$(\pm 0.02)$} & $0.13$\textcolor{gray}{$(\pm 0.02)$} & $0.19$\textcolor{gray}{$(\pm 0.04)$} \\
    $\overline{E}_{2,t}$    & $0.48$\textcolor{gray}{$(\pm 0.16)$} & $0.26$\textcolor{gray}{$(\pm 0.11)$} & $0.51$\textcolor{gray}{$(\pm 0.14)$} & $\bm{0.09}$\textcolor{gray}{$(\pm 0.05)$} & $0.72$\textcolor{gray}{$(\pm 0.09)$} & $0.37$\textcolor{gray}{$(\pm 0.09)$} & $0.16$\textcolor{gray}{$(\pm 0.03)$} & $0.12$\textcolor{gray}{$(\pm 0.05)$} \\
    $\overline{\rm rwd}_{2,t}$  & $62.65$\textcolor{gray}{$(\pm 8.63)$} & $68.76$\textcolor{gray}{$(\pm 4.97)$} & $46.73$\textcolor{gray}{$(\pm 16.4)$} & $75.14$\textcolor{gray}{$(\pm 10.1)$} & $76.08$\textcolor{gray}{$(\pm 8.51)$} & $76.58$\textcolor{gray}{$(\pm 6.37)$} & $\bm{77.86}$\textcolor{gray}{$(\pm 8.83)$} & $75.93$\textcolor{gray}{$(\pm 7.88)$} \\
    \midrule
    $\overline{\rm vol}_{3,t}$  & $1.25$\textcolor{gray}{$(\pm 0.12)$} & $0.32$\textcolor{gray}{$(\pm 0.07)$} & $0.64$\textcolor{gray}{$(\pm 0.09)$} & $1.05$\textcolor{gray}{$(\pm 0.13)$} & $0.52$\textcolor{gray}{$(\pm 0.03)$} & $0.25$\textcolor{gray}{$(\pm 0.06)$} & $0.34$\textcolor{gray}{$(\pm 0.07)$} & $\bm{0.18}$\textcolor{gray}{$(\pm 0.05)$} \\
    $\overline{T}_{3,t}$    & $0.34$\textcolor{gray}{$(\pm 0.10)$} & $0.16$\textcolor{gray}{$(\pm 0.09)$} & $\bm{0.11}$\textcolor{gray}{$(\pm 0.03)$} & $0.21$\textcolor{gray}{$(\pm 0.11)$} & $0.21$\textcolor{gray}{$(\pm 0.06)$} & $0.25$\textcolor{gray}{$(\pm 0.06)$} & $0.21$\textcolor{gray}{$(\pm 0.05)$} & $0.28$\textcolor{gray}{$(\pm 0.14)$} \\
    $\overline{E}_{3,t}$    & $0.16$\textcolor{gray}{$(\pm 0.05)$} & $\bm{0.07}$\textcolor{gray}{$(\pm 0.03)$}  & $0.59$\textcolor{gray}{$(\pm 0.12)$} & $0.14$\textcolor{gray}{$(\pm 0.05)$} & $0.32$\textcolor{gray}{$(\pm 0.12)$} & $0.11$\textcolor{gray}{$(\pm 0.05)$} & $0.25$\textcolor{gray}{$(\pm 0.05)$} & $0.34$\textcolor{gray}{$(\pm 0.16)$} \\
    $\overline{\rm rwd}_{3,t}$  & $71.85$\textcolor{gray}{$(\pm 7.62)$} & $77.95$\textcolor{gray}{$(\pm 3.08)$} & $67.28$\textcolor{gray}{$(\pm 10.9)$} & $73.33$\textcolor{gray}{$(\pm 5.81)$} & $77.67$\textcolor{gray}{$(\pm 6.25)$} & $79.63$\textcolor{gray}{$(\pm 4.43)$} & $81.56$\textcolor{gray}{$(\pm 6.51)$} & $\bm{82.75}$\textcolor{gray}{$(\pm 8.22)$} \\
    \midrule
    \midrule
    ${\rm vol}^{\rm total}$  & $66.84$\textcolor{gray}{$(\pm 4.06)$} & $27.62$\textcolor{gray}{$(\pm 4.42)$}  & $37.38$\textcolor{gray}{$(\pm 6.07)$} & $\bm{8.08}$\textcolor{gray}{$(\pm 4.06)$} & $23.76$\textcolor{gray}{$(\pm 4.93)$} & $10.96$\textcolor{gray}{$(\pm 3.34)$} & $19.22$\textcolor{gray}{$(\pm 3.34)$} & $22.31$\textcolor{gray}{$(\pm 2.16)$} \\
    ${T}^{\rm total}$    & $47.24$\textcolor{gray}{$(\pm 6.27)$} & $34.75$\textcolor{gray}{$(\pm 7.40)$}  & $34.75$\textcolor{gray}{$(\pm 7.40)$} & $9.35$\textcolor{gray}{$(\pm 2.89)$} & $11.9$\textcolor{gray}{$(\pm 3.15)$} & $\bm{3.95}$\textcolor{gray}{$(\pm 2.42)$} & $6.89$\textcolor{gray}{$(\pm 1.33)$} & $4.70$\textcolor{gray}{$(\pm 2.46)$} \\
    ${E}^{\rm total}$    & $13.26$\textcolor{gray}{$(\pm 3.62)$} & $1.35$\textcolor{gray}{$(\pm 1.35)$}  & $20.27$\textcolor{gray}{$(\pm 3.48)$} & $\bm{1.07}$\textcolor{gray}{$(\pm 0.24)$} & $12.32$\textcolor{gray}{$(\pm 2.32)$} & $4.60$\textcolor{gray}{$(\pm 1.36)$} & $10.23$\textcolor{gray}{$(\pm 2.28)$} & $9.91$\textcolor{gray}{$(\pm 2.01)$} \\
    ${\rm rwd}^{\rm total}$  & $183.42$\textcolor{gray}{$(\pm 25.92)$} & $212.12$\textcolor{gray}{$(\pm 14.03)$} & $170.52$\textcolor{gray}{$(\pm 37.5)$} & $203.3$\textcolor{gray}{$(\pm 30.41)$} & $229.78$\textcolor{gray}{$(\pm 23.54)$} & $232.4$\textcolor{gray}{$(\pm 18.22)$} & $\bm{245.27}$\textcolor{gray}{$(\pm 24.48)$} & $237.82$\textcolor{gray}{$(\pm 27.4)$} \\
    \midrule
    {\rm HVI} & $0.0926$ & $0.5154$ & $0.1505$ & $0.3284$ & $0.7045$ & $0.6866$ & $\bm{0.7258}$ & $0.6576$ \\
    \bottomrule
    \end{tabularx}%
    \label{tab:perform_compared_updated}%
      \begin{tablenotes}
        \footnotesize
        \item[1] $\overline{\rm vol}$, $\overline{T}$, $\overline{E}$, $\overline{\rm rwd}$ denote communication overheads (Mbits), latency (s), energy cost (J) and average reward (no units), respectively. The superscript ${\rm total}$ represents the sum over all tasks. HVI denotes the hypervolume indicator for multi-objective performance (higher is better).
        \item[2] We conduct five independent repeated experiments with different random seeds and train for $35$ communication rounds with three SPs.
      \end{tablenotes}
  \end{threeparttable}
\end{table*}

\begin{figure}[tb]
	\centerline{\includegraphics[width=0.9\linewidth]{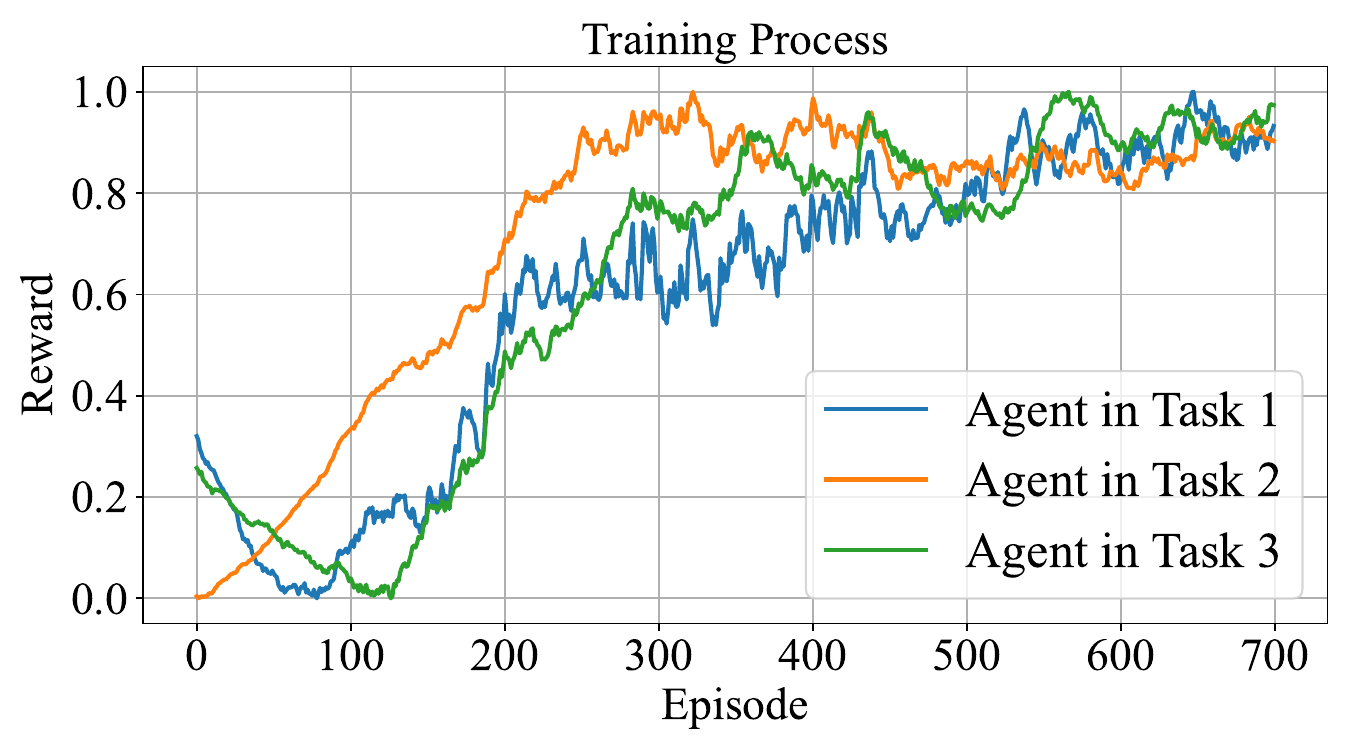}}
	\caption{Agents' training process, where rewards are normalized for display purposes.}
	\label{fig:training reward}
\end{figure}
\begin{figure}[tbp]
	\centerline{\includegraphics[width = .99\linewidth]{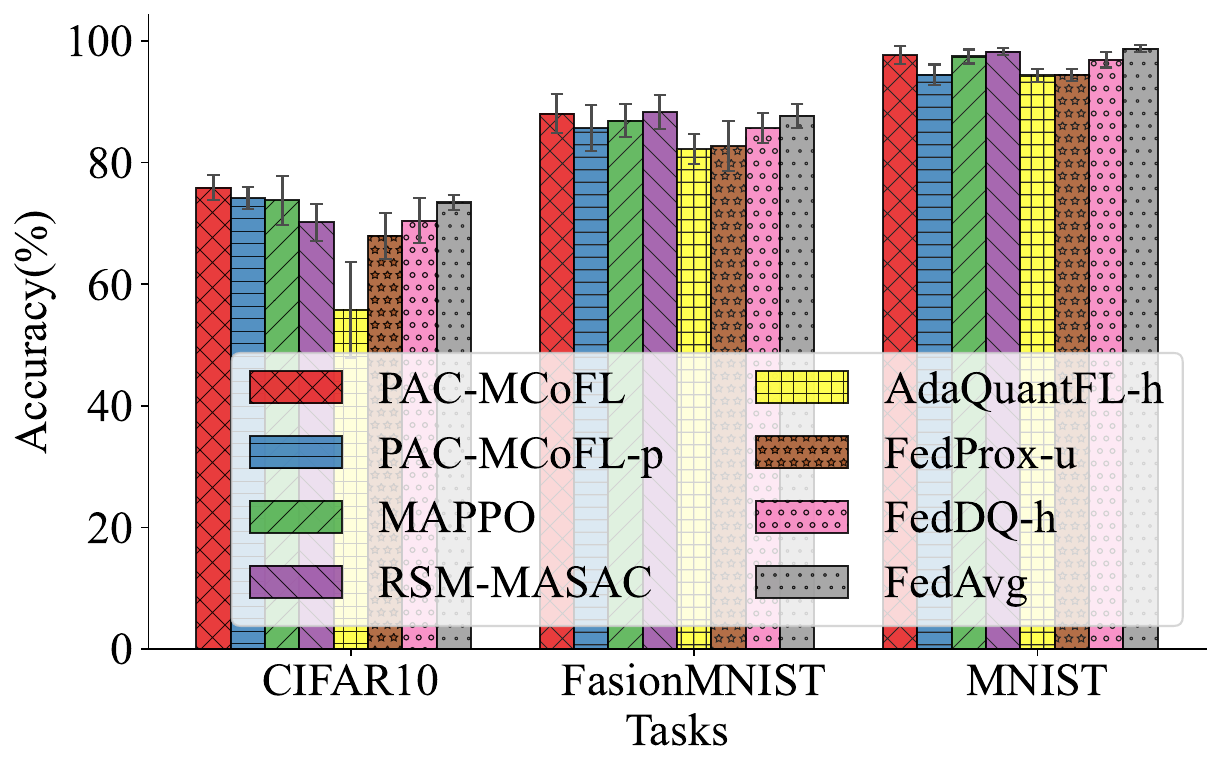}}
	\caption{Average test accuracy for the three services after a fixed number of FL global training rounds.}
	\label{fig:acc_compare}
\end{figure}

\begin{figure*}[tbp]
	\centerline{\includegraphics[width = \linewidth]{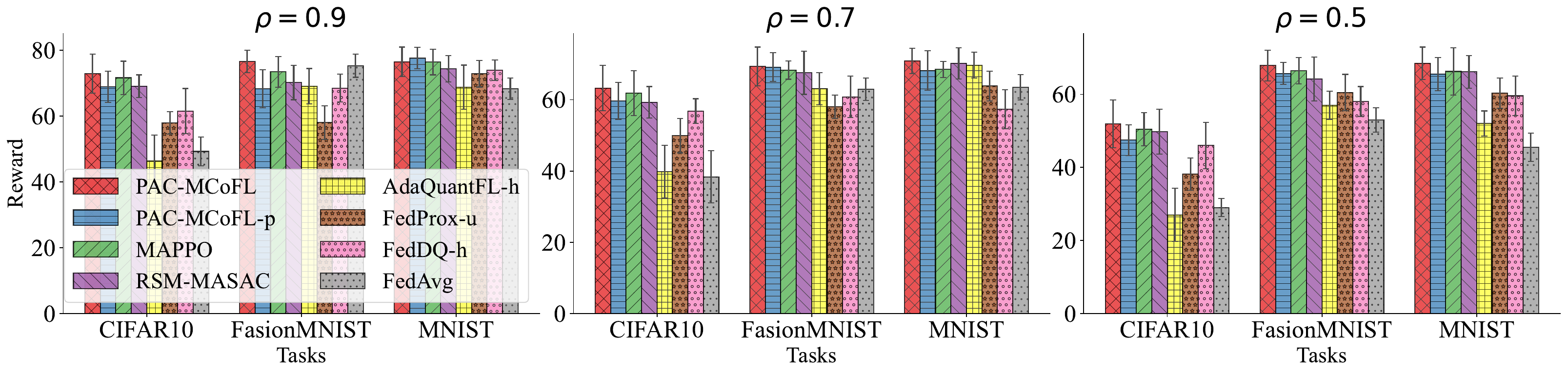}}
	\caption{Comparison of test accuracy across algorithms under varying non-IID degrees. We conduct five independent repeated experiments with different random seeds and train for $35$ communication rounds with three SPs.}
	\label{fig:noniid_comparision}
\end{figure*}

\subsubsection{Comparison with Baselines}
Fig. \ref{fig:training reward} depicts the training process of the PAC-based agent across three FL tasks. The reward trajectories converge to stability within several episodes, empirically validating the convergence guarantee established in Theorem \ref{thm:convergence} and ensuring the trained agents are effectively utilized for subsequent testing. Table \ref{tab:perform_compared} and Fig. \ref{fig:acc_compare} underscore the superior performance of the proposed \texttt{PAC-MCoFL} in balancing model accuracy, communication efficiency, and energy consumption under multi-SP scenarios. As evidenced in Table \ref{tab:perform_compared}, \texttt{PAC-MCoFL} significantly outperforms non-MARL baselines, with a $10.4\%$ increase in total reward and a $121\%$ improvement in HVI score over FedDQ-h and AdaQuantFL-h, respectively, which are limited to solving a part of the optimization task. This superiority thereby underscores the pronounced benefits of holistic joint optimization over localized or heuristic strategies. Within the more competitive landscape of MARL frameworks, it achieves the highest total reward ($245.27 \pm 24.48$) and demonstrates a $6.7\%$ lead over MAPPO and a $5.5\%$ lead over RSM-MASAC. While total reward serves as a valuable macro-indicator, the HVI offers a more nuanced assessment of Pareto efficiency in this multi-objective, competitive setting. \texttt{PAC-MCoFL} again leads with the highest HVI of $0.7258$, surpassing MAPPO by $3.0\%$ and RSM-MASAC by $5.7\%$. This quantitative superiority in both metrics indicates that \texttt{PAC-MCoFL} not only maximizes collective output but also discovers a more efficient and balanced equilibrium.
The variant \texttt{PAC-MCoFL-p} affirms its role as a computationally scalable yet potent alternative, as it secures the second-highest total reward ($237.82\pm27.4$) and a robust HVI of $0.6576$. Meanwhile, the modest, quantifiable performance gap between \texttt{PAC-MCoFL} and \texttt{PAC-MCoFL-p} also practically manifests the ``provably bounded error" established in Theorem \ref{thm:PCG}. It explicitly demonstrates that our parameterized generator provides a computationally scalable solution with only a minimal and acceptable deviation from the exhaustive search policy, thereby justifying its use.
Furthermore, Fig. \ref{fig:acc_compare} shows the average task accuracy achieved by various algorithms. \texttt{PAC-MCoFL} leads with the highest accuracy over all tasks, while \texttt{PAC-MCoFL-p}'s advantage is most pronounced in the more complex CIFAR-10 task where its accuracy is only marginally inferior to \texttt{PAC-MCoFL}. In summary, incorporating policy conjecture of other SPs effectively benefits multi-SP FL communication-computation co-optimization.

Fig. \ref{fig:noniid_comparision} illustrates the comparison of test accuracy across different algorithms under varying non-IID degrees. As non-IID severity increases ($\rho$ decreases), both \texttt{PAC-MCoFL} and \texttt{PAC-MCoFL-p} exhibit a graceful performance degradation, yet \texttt{PAC-MCoFL} outperforms the other algorithms in most cases. Even under extreme heterogeneity, i.e. $\rho = 0.5$, 
\texttt{PAC-MCoFL} consistently outperforms MAPPO and RSM-MASAC baseline, with reward improvements ranging from $1.8\text{--}3.1\%$ and $2.1\text{--}5.6\%$, respectively, across different tasks.
This robustness highlights \texttt{PAC-MCoFL}'s adaptive optimization capabilities in non-IID multi-SP environments.

\begin{figure}[tbp]
	\centerline{\includegraphics[width = \linewidth]{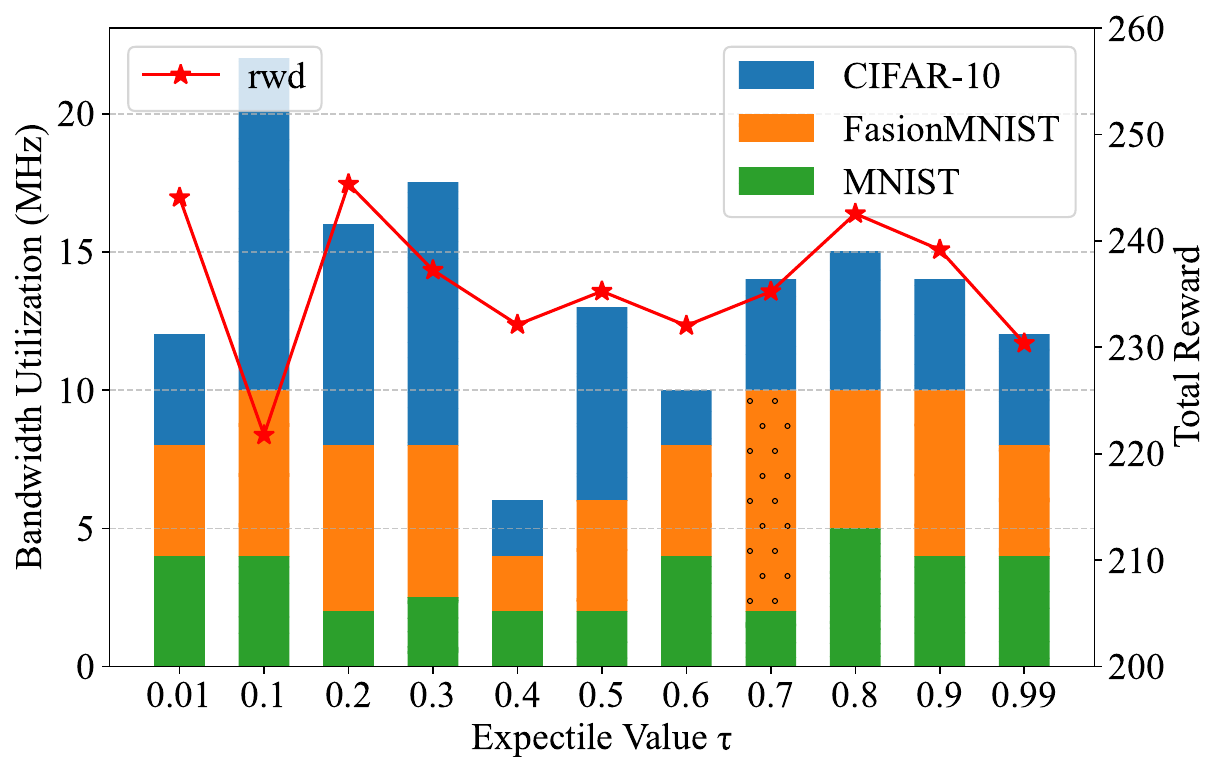}}
	\caption{Per-task bandwidth resource assignment and total reward under different expectile values. Total Reward means the sum of the average reward of all SPs.}
	\label{fig:expectile}
\end{figure}
\begin{figure}[tb]
    \centering
    \includegraphics[width=\linewidth]{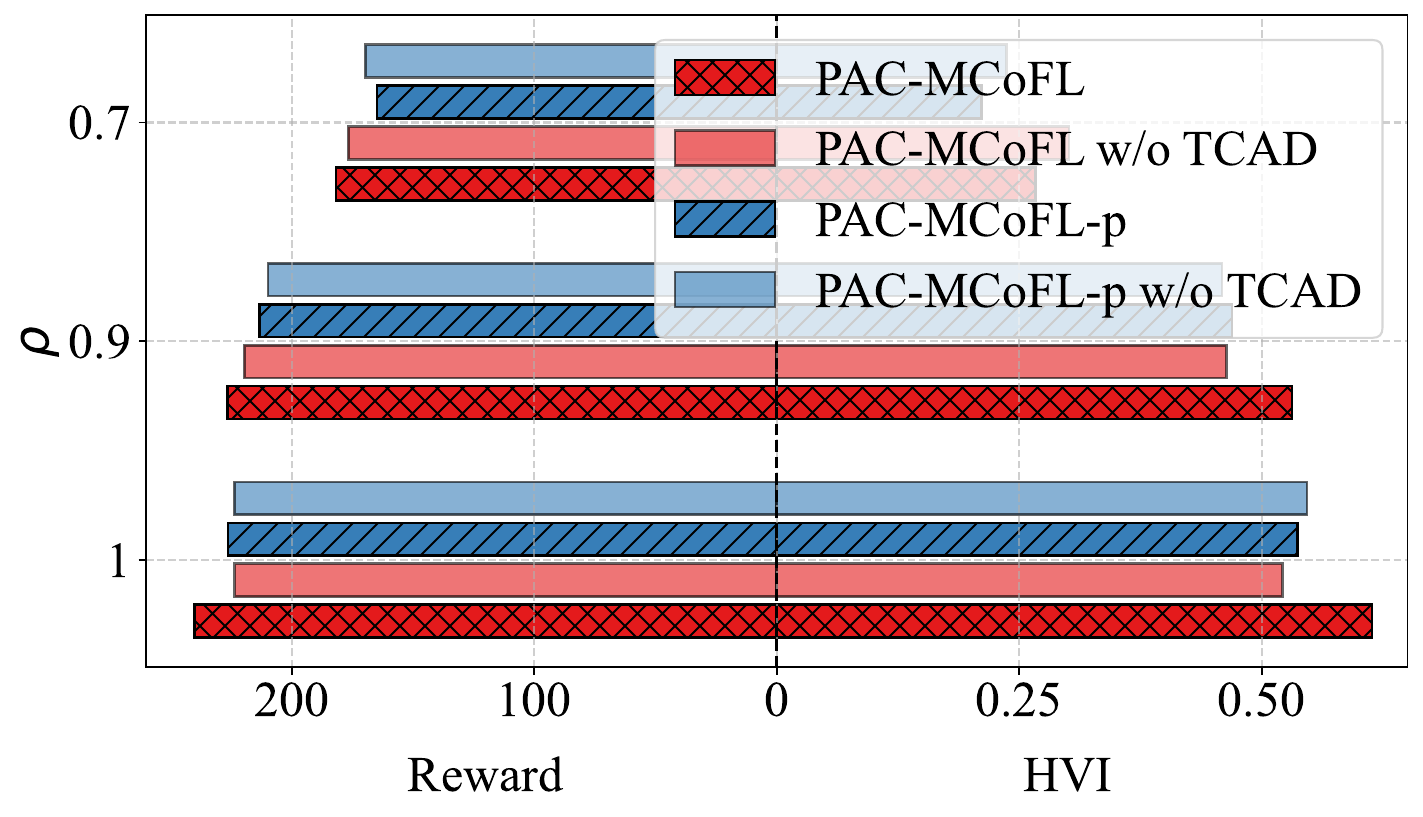}
    \caption{Impact of the TCAD mechanism.}
    \label{fig:tcad ablation}
\end{figure}

\subsubsection{Impact of Expectile Regression}
Fig. \ref{fig:expectile} shows the impact of the expectile value $\tau$ on SPs’ bandwidth allocation and the total system reward. Through an asymmetric penalty mechanism, $\tau$ implicitly guides SP policies, resulting in a U-shaped relationship between $\tau$ and bandwidth utilization. When $\tau< 0.5$, the critic network imposes a heavier penalty on negative TD errors, promoting aggressive policies to compensate for underestimated $Q$-values. This leads to bandwidth over-allocation and a higher proportion of resources for the complex CIFAR-10 task. Conversely, $\tau > 0.5$ yields more conservative and balanced bandwidth allocation. While the overall system reward fluctuates within the range of $230$–$240$, both extremes (i.e., $\tau=0.1$ and $\tau=0.99$) can cause up to $10.4\%$ performance degradation. Therefore, we set $\tau = 0.5$ hereafter for simplicity.

\subsubsection{Impact of TCAD Mechanism}
To validate the efficacy of the proposed TCAD, we conducted an ablation study by replacing it with a naive discretization variant\footnote{See Appendix C-B for related configurations and further discussion.}. As illustrated in Fig. \ref{fig:tcad ablation}, removing TCAD consistently results in a degradation in both total reward and the HVI score for the \texttt{PAC-MCoFL} and \texttt{PAC-MCoFL-p} frameworks alike. Across all scenarios, TCAD delivers an average increase of approximately $7.6\%$ in total reward and $14.2\%$ in HVI compared to the w/o TCAD variant. This performance gap underscores the limitations of this naive variant, which restricts the agent to coarse, disjointed action selections. In contrast, TCAD's capacity for fine-grained, incremental control is crucial for discovering sophisticated policies and navigating complex action spaces effectively.
\begin{table}[tbp]
  \centering
  \caption{GPU memory consumption (in MB) for different methods and values of $R$.}
  \label{tab:gpu}
  \begin{threeparttable} 
  \begin{tabular}{>{\raggedright}p{60pt} >{\raggedright}p{25pt} >{\raggedright}p{25pt} >{\raggedright}p{25pt} >{\raggedright\arraybackslash}p{25pt}}
    \bottomrule
    \bottomrule
    \textbf{Method} & \textbf{R=2} & \textbf{R=3} & \textbf{R=4} & \textbf{R=5} \\
    \midrule 
    MAPPO         & $512$M & $498$M  & $524$M & $524$M \\
    PAC-MCoFL   & $536$M & $2132$M & OOM\tnote{1}   & OOM   \\
    PAC-MCoFL-p   & $496$M & $490$M  & $508$M  & $508$M  \\
    \bottomrule
    \bottomrule
  \end{tabular}
  \begin{tablenotes}   
    \footnotesize            
    \item[1]  Out of Memory (OOM) denotes that the memory requirements exceeded the capacity of the current experimental hardware (NVIDIA GeForce RTX 3090).
  \end{tablenotes}
  \end{threeparttable}     
\end{table}
\begin{figure}[tbp]
	\centerline{\includegraphics[width = \linewidth]{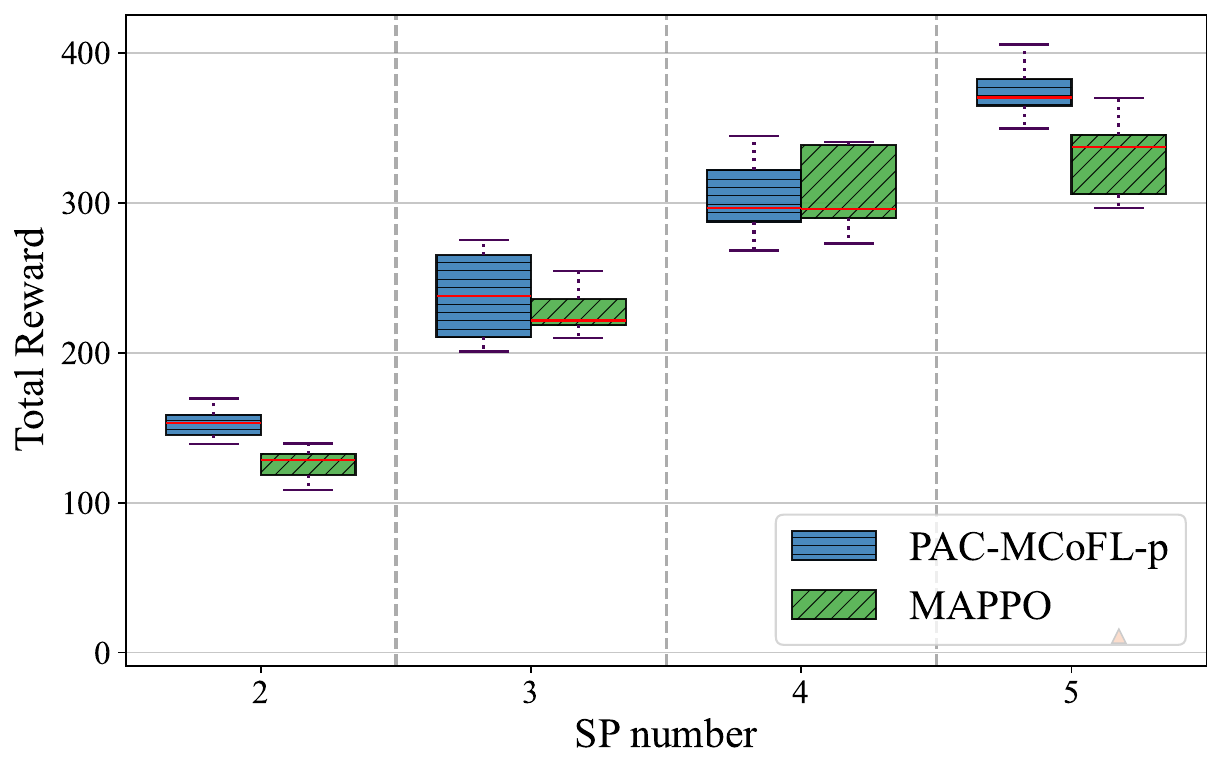}}
	\caption{Experiment with broader multi-agent complexities. The task datasets and corresponding models used by different SPs under different $R$ will be given in detail in the Appendix B-B. Total Reward is the sum of the average rewards across all SPs.}
	\label{fig:magents}
\end{figure}
\subsubsection{Impact of SP Numbers}
The exhaustive policy conjecture in Eq. \eqref{eq:bestotherspi} incurs a computational complexity that grows exponentially with the number of SPs, rendering \texttt{PAC-MCoFL} intractable in larger scenarios. \texttt{PAC-MCOFL-p} is specifically designed to overcome this scalability bottleneck. Table \ref{tab:gpu} shows that the standard \texttt{PAC-MCoFL} encounters out-of-memory (OOM) errors for $R\geq4$ on our hardware, whereas \texttt{PAC-MCOFL-p} remains computationally feasible. Fig. \ref{fig:magents} further reveals that this efficiency does not compromise performance. \texttt{PAC-MCOFL-p} consistently outperforms the scalable baseline MAPPO\footnote{Notably, MAPPO does not involve policy conjecture, thereby avoiding the costly enumeration of all possible joint actions of neighboring agents.}, achieving an average reward improvement of approximately $21.3\%$. Crucially, this scalability analysis also accounts for increased task heterogeneity, as each new SP introduces a task with distinct characteristics as detailed in Appendix B-B.
\begin{table*}[!t]
    \centering
    \caption{Results of different objective weight factors $\boldsymbol{\sigma}$.}
    \label{tab:ablation_exp}
    \fontsize{8}{12.5}\selectfont
    \begin{threeparttable}
        \begin{tabular}{l|ccccc|ccccc|ccccc}
            \toprule
            \emph{Service} & \multicolumn{5}{c|}{\emph{CIFAR-10}} & \multicolumn{5}{c|}{\emph{Fashion-MNIST}} & \multicolumn{5}{c}{\emph{MNIST}} \\
            \emph{Metric} & \multicolumn{1}{c}{$\Acc^{\rm max}_{1,t}$} & \multicolumn{1}{c}{$\overline{\vol}_{1,t}$} & \multicolumn{1}{c}{$\overline{T}_{1,t}$} & \multicolumn{1}{c}{$\overline{E}_{1,t}$} & \multicolumn{1}{c|}{$\overline{\text{rwd}}$} & \multicolumn{1}{c}{$\Acc^{\rm max}_{2,t}$} & \multicolumn{1}{c}{$\overline{\vol}$} & \multicolumn{1}{c}{$\overline{T}_{2,t}$} & \multicolumn{1}{c}{$\overline{E}_{2,t}$} & \multicolumn{1}{c|}{$\overline{\text{rwd}}$} & \multicolumn{1}{c}{$\Acc^{\rm max}_{3,t}$} & \multicolumn{1}{c}{$\overline{\vol}$} & \multicolumn{1}{c}{$\overline{T}_{3,t}$} & \multicolumn{1}{c}{$\overline{E}_{3,t}$} & \multicolumn{1}{c}{$\overline{\text{rwd}}$} \\
            \midrule
            $\sigma_{1\!\sim\!4}\neq0$ & \textbf{78.55}\% & 18.74 & \textbf{6.62} & \textbf{9.96} & \textbf{83.85} & \textbf{91.01}\% & 0.146 & 0.132 & \textbf{0.161} & \textbf{77.86} & \textbf{98.23}\% & 0.315 & \textbf{0.162} & 0.21 & 81.56 \\
            $\sigma_1\!=\!0$ & 40.52\% & \sout{13.67} & \sout{5.41} & \sout{6.58} & \sout{48.92} & 88.64\% & 0.123 & \textbf{0.08} & 0.21 & 64.76 & 97.84\% & 0.35 & 0.19 & 0.61 & 77.69 \\
            $\sigma_2\!=\!0$ & 75.28\% & \textbf{16.7} & 6.98 & 13.25 & 76.03 & 86.02\% & \textbf{0.092} & 0.152 & 0.178 & 62.14 & 96.07\% & \textbf{0.27} & 0.24 & 0.353 & \textbf{82.82} \\
            $\sigma_3\!=\!0$ & 68.13\% & 18.14 & 7.42 & 11.53 & 54.39 & 59.35\% & \sout{0.031} & \sout{0.121} & \sout{0.082} & \sout{49.93} & 97.68\% & 0.46 & 0.232 & \textbf{0.271} & 71.85 \\
            $\sigma_4\!=\!0$ & 46.06\% & \sout{22.13} & \sout{23.17} & \sout{7.44} & \sout{56.51} & 88.32\% & 0.224 & 0.242 & 0.85 & 72.82 & 97.78\% & 0.68 & 0.315 & 0.518 & 67.28 \\
            \bottomrule
        \end{tabular}
        \begin{tablenotes}
			\footnotesize
			\item[1] $\overline{{\rm vol}}\subscripttr,\overline{T}\subscripttr,\overline{E}\subscripttr$ which are in Mbits, s, and J, respectively, represent the average communication overheads, latency, and energy consumption consumed during FL training rounds. $\overline{\text{rwd}}$ is the average reward and has no units.
			\item[2] Data with a strikethrough indicates that the business model does not eventually converge (accuracy still fluctuates significantly), and bolded data indicates an advantage in comparison with the same metric.
		\end{tablenotes}
      \end{threeparttable}
\end{table*}
\begin{table*}[!tb]
  \centering
  \caption{Performance Comparison Under Different Numbers of Clients ($N$) and Degrees of Non-IID Level ($\rho$).}
  \label{tab:diff ue and noniid}
  \renewcommand{\arraystretch}{1.25} 
  \setlength{\tabcolsep}{4.5pt} 
  \begin{threeparttable}
    \begin{tabular}{@{}c|c|ccc|ccc@{}}
    \toprule
    \multicolumn{1}{c}{\multirow{2}{*}{\textbf{Clients ($N$)}}} & \multicolumn{1}{c}{\multirow{2}{*}{\textbf{Non-IID ($\rho$)}}} & \multicolumn{3}{c|}{\textbf{Resource Overheads}} & \multicolumn{3}{c}{\textbf{Final Test Accuracy (\%)}} \\
    \cmidrule(lr){3-5} \cmidrule(lr){6-8}
    \multicolumn{1}{c}{} & \multicolumn{1}{c}{} & \textbf{$E^{\rm total}$} (J) & \textbf{$T^{\rm total}$} (s) & \textbf{${\rm vol}^{\rm total}$} (Mbits) & \textbf{CIFAR-10} & \textbf{Fashion-MNIST} & \textbf{MNIST} \\
    \midrule
    \multirow{4}{*}{\textbf{5}} 
    & 1.0 & $360.34$\textcolor{gray}{$(\pm 38.02)$} & $205.74$\textcolor{gray}{$(\pm 36.40)$} & $543.11$\textcolor{gray}{$(\pm 58.26)$} & $75.89$\textcolor{gray}{$(\pm 2.54)$} & $88.06$\textcolor{gray}{$(\pm 3.24)$} & $97.65$\textcolor{gray}{$(\pm 1.25)$} \\
    & 0.9 & $374.22$\textcolor{gray}{$(\pm 32.65)$} & $351.96$\textcolor{gray}{$(\pm 47.63)$} & $634.74$\textcolor{gray}{$(\pm 57.64)$} & $66.08$\textcolor{gray}{$(\pm 6.94)$} & $83.64$\textcolor{gray}{$(\pm 3.77)$} & $91.56$\textcolor{gray}{$(\pm 1.12)$} \\
    & 0.7 & $324.96$\textcolor{gray}{$(\pm 28.96)$} & $281.39$\textcolor{gray}{$(\pm 32.50)$} & $867.19$\textcolor{gray}{$(\pm 80.24)$} & $55.26$\textcolor{gray}{$(\pm 6.34)$} & $76.35$\textcolor{gray}{$(\pm 3.44)$} & $86.89$\textcolor{gray}{$(\pm 2.57)$} \\
    & 0.5 & $521.92$\textcolor{gray}{$(\pm 40.25)$} & $384.33$\textcolor{gray}{$(\pm 43.56)$} & $1156.61$\textcolor{gray}{$(\pm 108.68)$} & $45.89$\textcolor{gray}{$(\pm 7.64)$} & $69.86$\textcolor{gray}{$(\pm 4.21)$} & $73.50$\textcolor{gray}{$(\pm 2.42)$} \\
    \midrule
    \multirow{4}{*}{\textbf{10}} 
    & 1.0 & $488.79$\textcolor{gray}{$(\pm 48.99)$} & $286.56$\textcolor{gray}{$(\pm 21.26)$} & $991.21$\textcolor{gray}{$(\pm 90.87)$} & $74.76$\textcolor{gray}{$(\pm 3.12)$} & $86.84$\textcolor{gray}{$(\pm 3.21)$} & $97.67$\textcolor{gray}{$(\pm 1.21)$} \\
    & 0.9 & $633.69$\textcolor{gray}{$(\pm 64.68)$} & $369.98$\textcolor{gray}{$(\pm 40.38)$} & $1416.21$\textcolor{gray}{$(\pm 118.68)$} & $68.76$\textcolor{gray}{$(\pm 7.24)$} & $80.64$\textcolor{gray}{$(\pm 4.20)$} & $94.17$\textcolor{gray}{$(\pm 1.50)$} \\
    & 0.7 & $760.75$\textcolor{gray}{$(\pm 65.96)$} & $426.75$\textcolor{gray}{$(\pm 44.50)$} & $2010.98$\textcolor{gray}{$(\pm 155.48)$} & $62.98$\textcolor{gray}{$(\pm 6.69)$} & $72.38$\textcolor{gray}{$(\pm 4.23)$} & $91.43$\textcolor{gray}{$(\pm 3.48)$} \\
    & 0.5 & $788.81$\textcolor{gray}{$(\pm 88.48)$} & $325.03$\textcolor{gray}{$(\pm 51.11)$} & $1923.45$\textcolor{gray}{$(\pm 143.68)$} & $48.76$\textcolor{gray}{$(\pm 8.03)$} & $69.75$\textcolor{gray}{$(\pm 5.54)$} & $87.24$\textcolor{gray}{$(\pm 4.23)$} \\
    \midrule
    \multirow{4}{*}{\textbf{15}} 
    & 1.0 & $924.45$\textcolor{gray}{$(\pm 53.79)$} & $392.63$\textcolor{gray}{$(\pm 24.86)$} & $1379.64$\textcolor{gray}{$(\pm 88.37)$} & $74.04$\textcolor{gray}{$(\pm 2.32)$} & $83.17$\textcolor{gray}{$(\pm 3.17)$} & $94.16$\textcolor{gray}{$(\pm 1.46)$} \\
    & 0.9 & $876.56$\textcolor{gray}{$(\pm 58.63)$} & $394.98$\textcolor{gray}{$(\pm 39.30)$} & $1250.35$\textcolor{gray}{$(\pm 118.68)$} & $68.84$\textcolor{gray}{$(\pm 3.19)$} & $80.22$\textcolor{gray}{$(\pm 2.88)$} & $93.85$\textcolor{gray}{$(\pm 1.29)$} \\
    & 0.7 & $986.68$\textcolor{gray}{$(\pm 42.36)$} & $467.65$\textcolor{gray}{$(\pm 46.96)$} & $2656.65$\textcolor{gray}{$(\pm 160.69)$} & $67.26$\textcolor{gray}{$(\pm 5.26)$} & $78.79$\textcolor{gray}{$(\pm 2.16)$} & $89.77$\textcolor{gray}{$(\pm 4.26)$} \\
    & 0.5 & $906.68$\textcolor{gray}{$(\pm 46.63)$} & $405.05$\textcolor{gray}{$(\pm 36.12)$} & $2170.15$\textcolor{gray}{$(\pm 152.44)$} & $61.32$\textcolor{gray}{$(\pm 5.21)$} & $75.67$\textcolor{gray}{$(\pm 4.55)$} & $86.58$\textcolor{gray}{$(\pm 2.16)$} \\
    \bottomrule
    \end{tabular}
    \begin{tablenotes}
        \item[*] ${\rm vol}^{\rm total}$, $T^{\rm total}$, $E^{\rm total}$ denote the sum of communication overheads, latency and energy cost over all tasks, respectively. We conduct five independent repeated experiments with different random seeds and train for $35$ communication rounds with three SPs.
    \end{tablenotes}
  \end{threeparttable}
\end{table*}


\subsubsection{Impact of Different Weight Factors $\boldsymbol{\sigma}$}
To evaluate the impact of the different terms introduced in the objective function, specifically the reward function in Eq. \eqref{reward}, on the performance of \texttt{PAC-MCoFL}, we sequentially set $\sigma_1$ to $\sigma_4$ to zero in turn. We then train the agents accordingly, and test the performance across various metrics within a fixed FL global training round. The results are summarized as shown in Table \ref{tab:ablation_exp}, where accuracy $\Acc\subscripttr^{\rm max}$ refers to the obtained maximum value, while the other metrics are averaged over the FL rounds. It can be observed that the absence of the accuracy factor in the reward function ($\sigma_1 = 0$) leads to the most significant performance degradation, with \texttt{PAC-MCoFL} reward dropping by $41.6\%$, $14.8\%$, and $4.7\%$ across the three tasks, underscoring the critical importance of accuracy. Excluding the adversarial factor ($\sigma_2 = 0$) has minimal impact on the overall reward, though it negatively affects individual task metrics such as $T\subscripttr$, $E\subscripttr$, and $\vol\subscripttr$. Omitting latency ($\sigma_3 = 0$) reduces CIFAR-10 and Fashion-MNIST reward by $35.13\%$ and $34.3\%$, respectively, while excluding energy consumption ($\sigma_4 = 0$) results in a $32.6\%$ drop in the CIFAR-10 reward. These results suggest that without latency and energy constraints, the adversarial factor becomes overly dominant, leading to weaker performance in certain tasks. The weight factors study reveals that accuracy is the most critical factor for \texttt{PAC-MCoFL}, with latency and energy constraints also playing key roles in maintaining task balance. While removing the adversarial factor has a minimal overall impact, it is likely to degrade individual task performance.
\begin{figure*}[tbp]
	\centerline{\includegraphics[width=\linewidth]{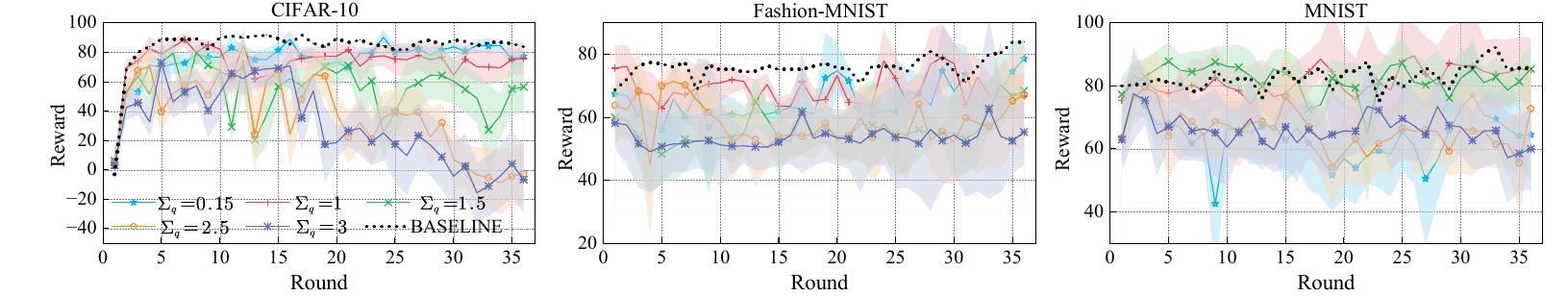}}
	\caption{Impact of different coefficient $\Sigma_q$ on system performance. The ``BASELINE" configuration corresponds to the simulation parameters in Table \ref{tab:simulation paras} (i.e., $\Sigma_q = 0.5$ and $\Sigma_f = 0.25$).}
	\label{fig:sigmaq}
\end{figure*}
\begin{figure*}[tbp]
	\centerline{\includegraphics[width=\linewidth]{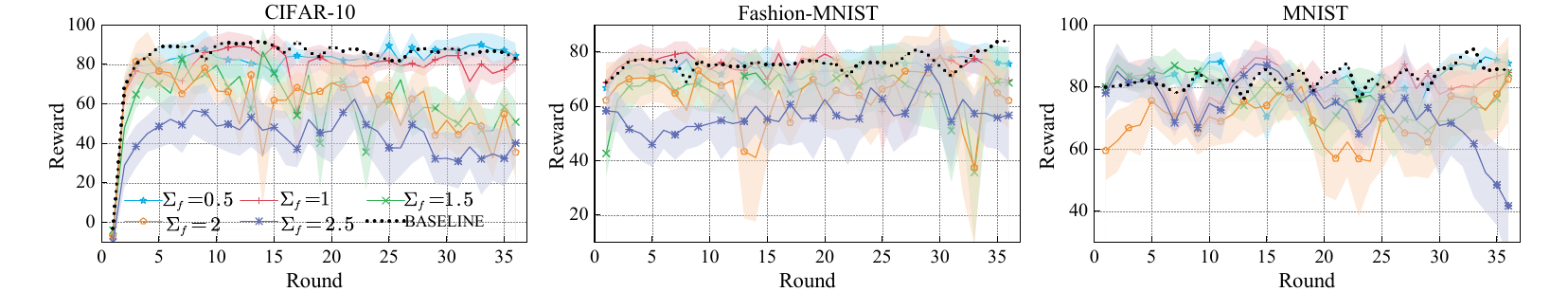}}
	\caption{Impact of different coefficient $\Sigma_f$ on system performance. The ``BASELINE" configuration corresponds to the simulation parameters in Table \ref{tab:simulation paras} (i.e., $\Sigma_q = 0.5$ and $\Sigma_f = 0.25$).}
	\label{fig:sigmaf}
\end{figure*}
\subsubsection{Different Number of Clients at Different non-IID Degrees}
We investigate the impact of different degrees of non-IID settings and varying numbers of clients on the performance of \texttt{PAC-MCoFL}. The results in Table \ref{tab:diff ue and noniid} reveal two primary trends. First, scaling the number of clients from $5$ to $15$ introduces a clear tension between resource overhead and model performance. As expected, total energy $E^{\rm total}$, delay $T^{\rm total}$, and communication overheads ${\vol}^{\rm total}$ exhibit a general upward trend with an increase of $N$, reflecting the costs of coordinating a larger client pool. Concurrently, a consistent degradation in final test accuracy is observed across all tasks, highlighting the inherent challenges of scaling. Second, for a fixed client number, increasing data heterogeneity (i.e., decreasing $\rho$) not only markedly reduces final model accuracy but also simultaneously inflates the resource overhead required for training. This demonstrates that the framework adaptively expends more resources to counteract the negative effects of statistical heterogeneity.

\subsubsection{Effect of Different Jitter Variance Factors}
In \texttt{PAC-MCoFL}, decision-making occurs at the SP level, i.e., on the server side. As mentioned in Section \ref{sec:Markov Decision Process}, the behavior of clients, such as quantization level $q\subscriptitr$ and CPU frequency $f\subscriptitr$, are sampled based on the SP's decisions, with an additional variance $\Sigma_q$ and $\Sigma_f$. To further assess performance, we evaluate different values of $\Sigma$, conducting five independent experiments for each configuration. Fig. \ref{fig:sigmaq} and Fig. \ref{fig:sigmaf} depict the reward achieved by the three FL tasks during a fixed number of global training rounds under varying $\Sigma_q$ and $\Sigma_f$ settings. As shown in Fig. \ref{fig:sigmaq}, increasing $\Sigma_q$ results in lower reward and greater variability. In particular, when $\Sigma_q = 3$, the reward significantly deteriorates across all three tasks. In the CIFAR-10 task, $\Sigma_q = 2.5$ and $\Sigma_q = 3$ even exhibit a downward trend, indicating that larger $\Sigma_q$ values may adversely affect the system's convergence performance. This suggests that when quantization strategies among FL clients vary too widely, the resulting divergence in the uploaded local models can hinder convergence.
Similarly, Fig. \ref{fig:sigmaf} shows that $\Sigma_f = 2$ and $\Sigma_f = 2.5$ lead to the poorest performance across all tasks, with $\Sigma_f = 2.5$ even exhibiting a lack of convergence in reward for the MNIST task. In contrast, the results for $\Sigma_f = 0.5$ and $\Sigma_f = 1$ exhibit only minor fluctuations compared to the BASELINE, indicating that small $\Sigma_f$ values have a negligible impact on system performance.

\section{Conclusion and Future Work}
\label{section_conclusion}
In this paper, we have developed the \texttt{PAC-MCoFL}—a novel game-theoretic MARL framework for jointly optimizing communication and computation in non-cooperative multi-SP FL. By integrating PAC principles with expectile regression, \texttt{PAC-MCoFL} steers various SP agents toward Pareto-optimal equilibria while capturing their heterogeneous risk preferences. Furthermore, \texttt{PAC-MCoFL} leverages a TCAD mechanism to enable fine-grained, multi-dimensional action control. To ensure both practicality and scalability, we have also designed a lightweight variant,  \texttt{PAC-MCoFL-p}, which provably overcomes the exponential complexity of policy conjecture under bounded-error guarantees. In addition to the theoretical convergence results,  extensive simulations demonstrate superior performance. Prominently, \texttt{PAC-MCoFL} achieves improvements of approximately $5.8\%$ in total reward and $4.2\%$ in HVI over strong MARL-based baselines. Subsequent research will investigate adaptation mechanisms, such as dynamic $\tau$-adaptation for time-varying risk profiles and automatic tuning of the reward weights $\sigma$ to balance multi-objective trade-offs, as well as the extension to mixed cooperative-competitive settings.
\bibliographystyle{IEEEtran}
\bibliography{paper}
\begin{IEEEbiography}[{\includegraphics[width=1in,height=1.25in,clip,keepaspectratio]{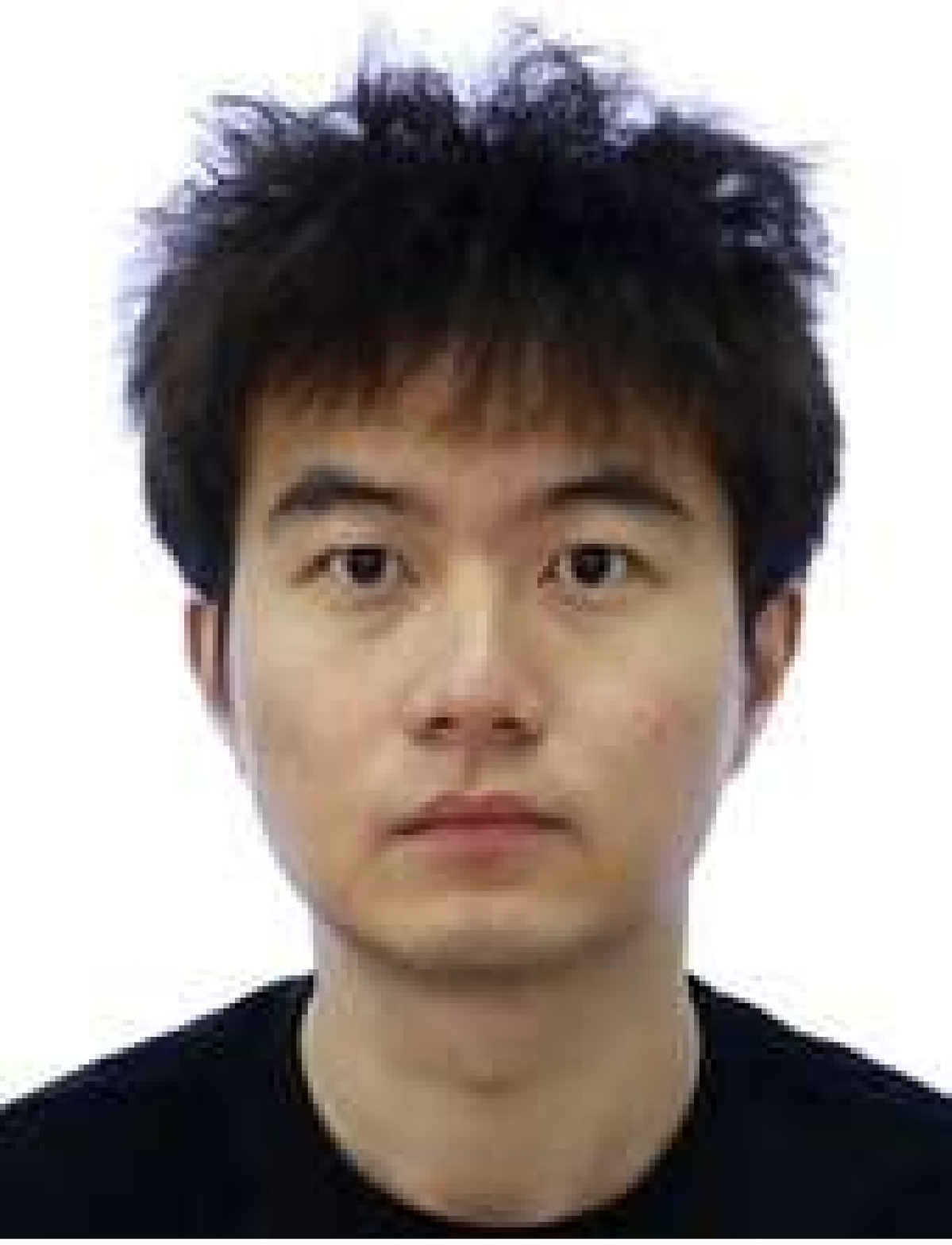}}]{Renxuan Tan}(Student Member, IEEE) received the
 B.E. degree in communication engineering from Sichuan University, Chengdu, China, in 2024. He is currently pursuing the Ph.D. degree with the College of Information Science and Electronic Engineering, Zhejiang University, Hangzhou, China. His research interests include reinforcement learning, wireless communications.
\end{IEEEbiography}
\begin{IEEEbiography}[{\includegraphics[width=1in,height=1.25in,clip,keepaspectratio]{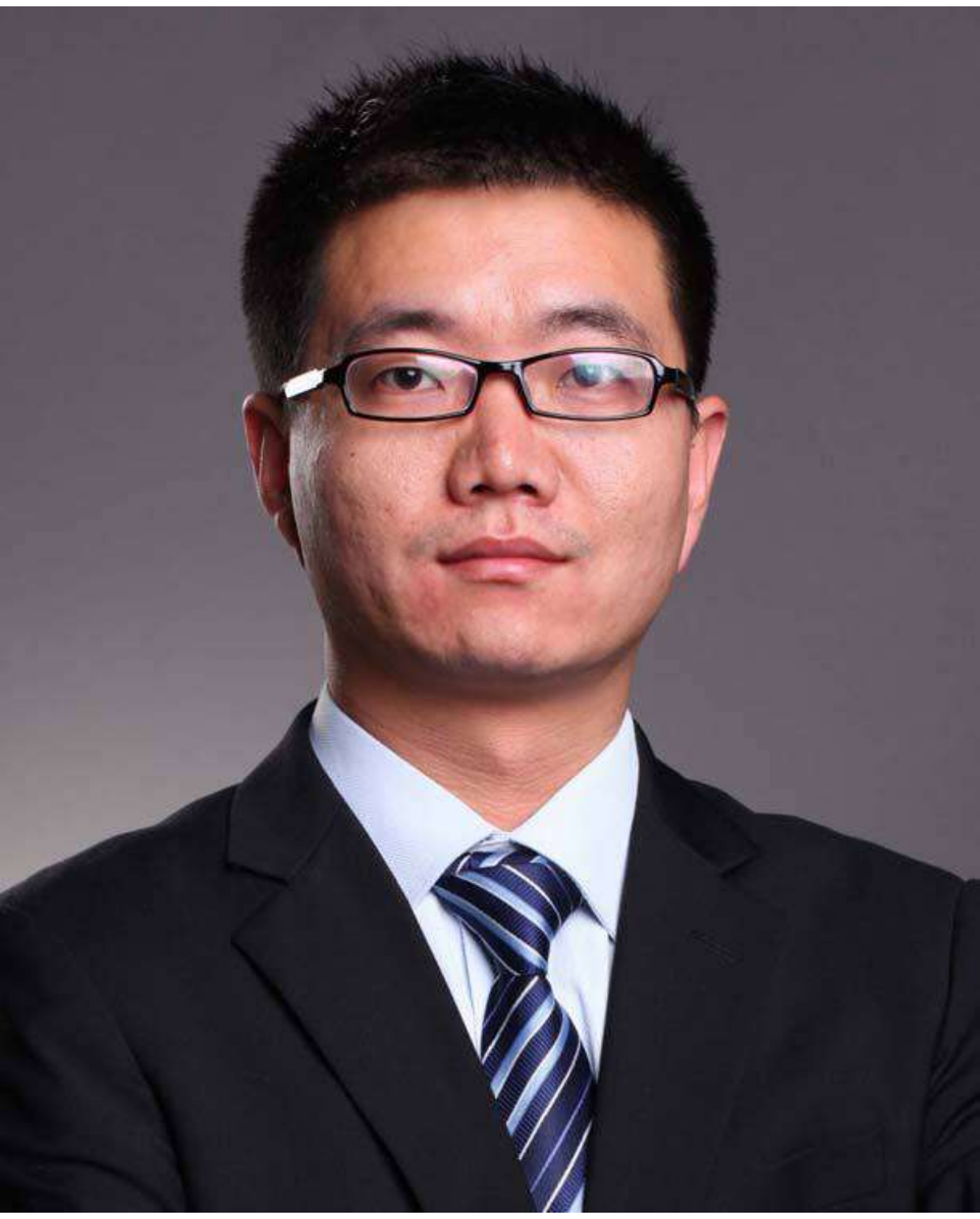}}]{Rongpeng Li}(Senior Member, IEEE) is currently an Associate Professor with the College of Information Science and Electronic Engineering, Zhejiang University. He received the B.E. degree from Xidian University, Xi’an, China, in June 2010, and the Ph.D. degree from Zhejiang University, Hangzhou, China, in June 2015. From August 2015 to September 2016, he was a Research Engineer with the Wireless Communication Laboratory, Huawei Technologies Company Ltd., Shanghai, China. He was a Visiting Scholar with the Department of Computer Science and Technology, University of Cambridge, Cambridge, U.K., from February 2020 to August 2020. His current research interests focus on networked intelligence for comprehensive efficiency (NICE).
\end{IEEEbiography}
\vspace{-1cm}
\begin{IEEEbiography}[{\includegraphics[width=1in,height=1.25in,clip,keepaspectratio]{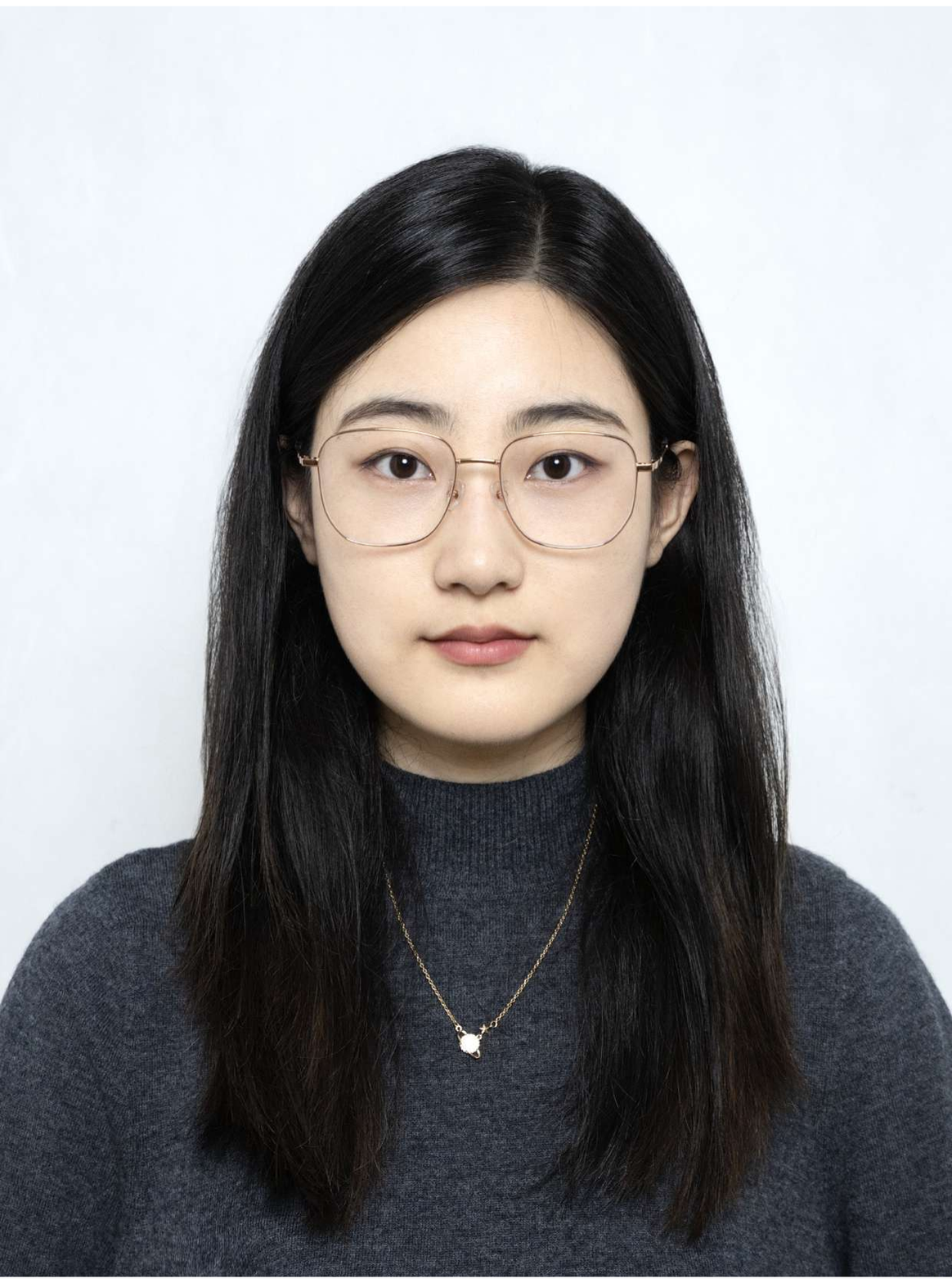}}]{Xiaoxue Yu}(Student Member, IEEE) received the
 B.E. degree in communication engineering from Xidian University, Xi’an, China. She is currently pursuing the Ph.D. degree with the College of Information Science and Electronic Engineering, Zhejiang University, Hangzhou, China. Her research interests include communications in distributed learning and multiagent reinforcement learning.
\end{IEEEbiography}
\vspace{-1cm}
\begin{IEEEbiography}[{\includegraphics[width=1in,height=1.25in,clip,keepaspectratio]{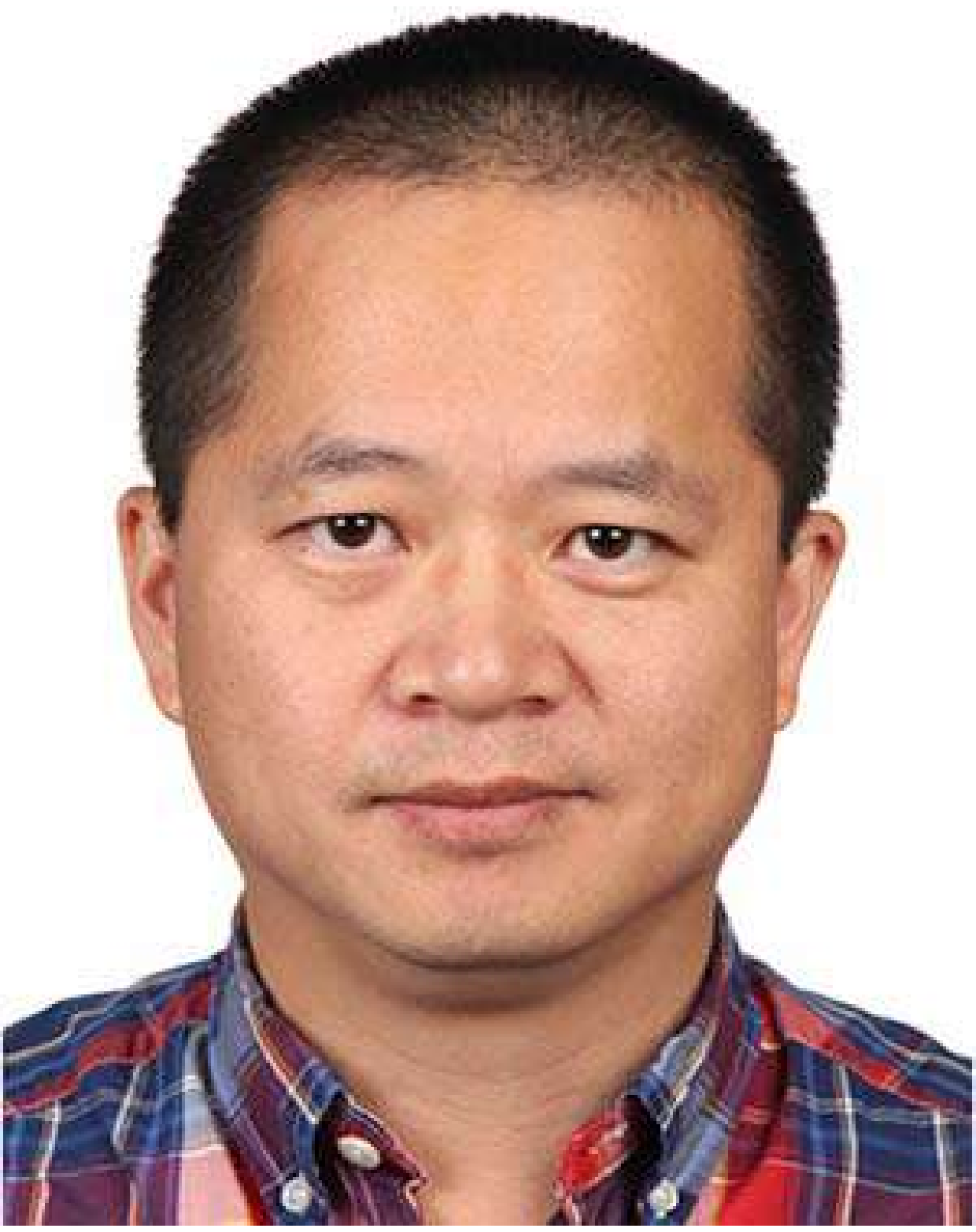}}]{Xianfu Chen}(Member, IEEE) received his Ph.D. degree (with Hons.) from the Zhejiang University, Hangzhou, China, in 2012. In 2012, he joined the VTT Technical Research Centre of Finland, Oulu, Finland, as a Research Scientist and as a Senior Scientist from 2013 to 2023. He is currently a Chief Research Engineer with the Shenzhen CyberAray Network Technology Co., Ltd, Shenzhen, China. His research interests include various aspects of wireless communications and networking, with emphasis on human-level and artificial intelligence for resource awareness in next-generation communication networks. He was the recipient of the 2021 IEEE Communications Society Outstanding Paper Award, and the 2021 IEEE Internet of Things Journal Best Paper Award. He is an Editor of IEEE Open Journal of the Communications Society, an Academic Editor of Wireless Communications and Mobile Computing, and an Associate Editor of China Communications.
\end{IEEEbiography}
\vspace{-1cm}
\begin{IEEEbiography}[{\includegraphics[width=1in,height=1.25in,clip,keepaspectratio]{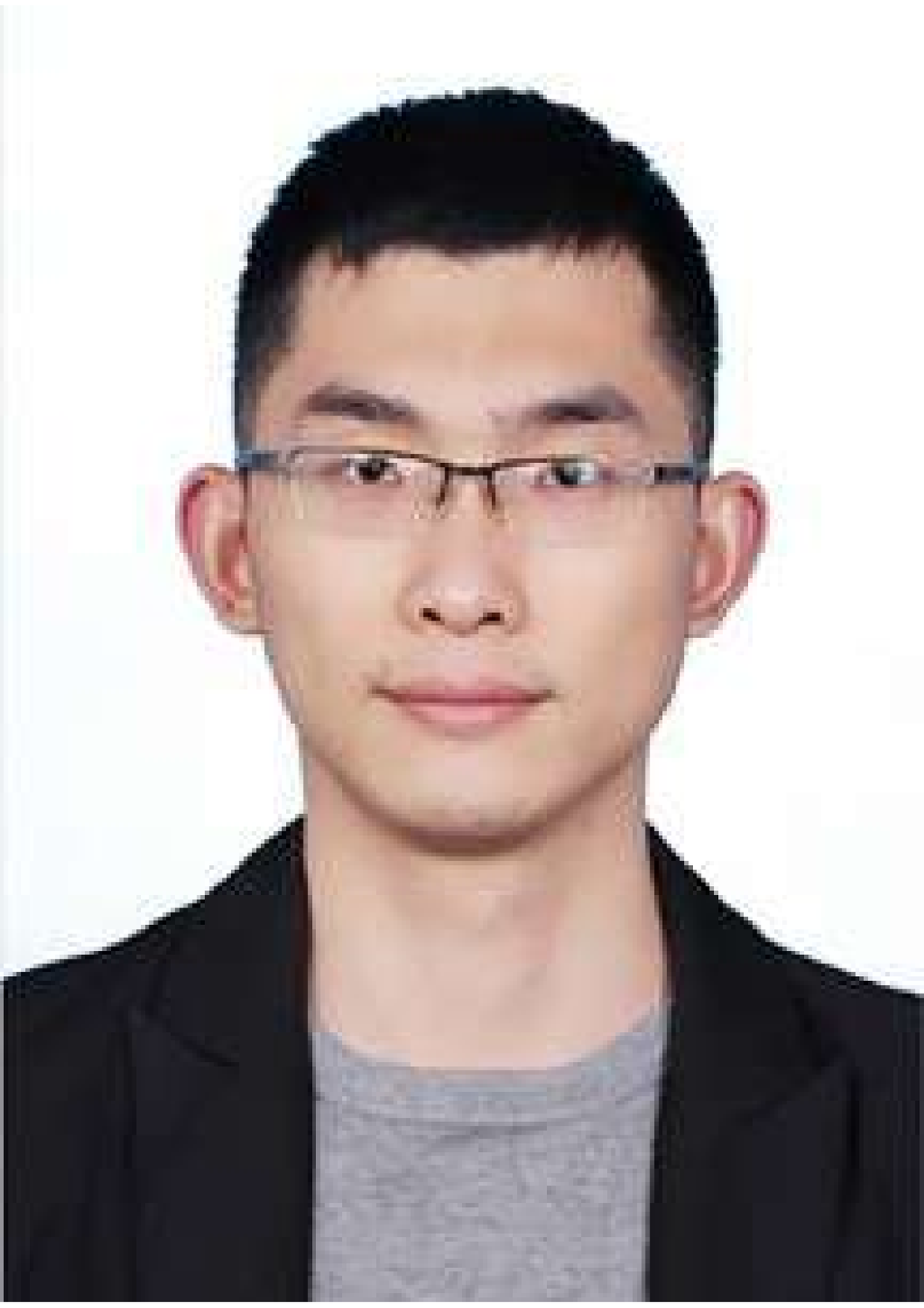}}]{Xing Xu}is now a data engineer of Information and Communication Branch of State Grid Hebei Electric Power Co., Ltd., Shijiazhuang, China. He received the Ph.D. degree in the College of Information Science and Electronic Engineering, Zhejiang University, Hangzhou, China. His research interests include collective intelligence, reinforcement learning, multi-agent reinforcement learning, federated learning, and data mining.
\end{IEEEbiography}
\vspace{-1cm}
\begin{IEEEbiography}[{\includegraphics[width=1in,height=1.25in,clip,keepaspectratio]{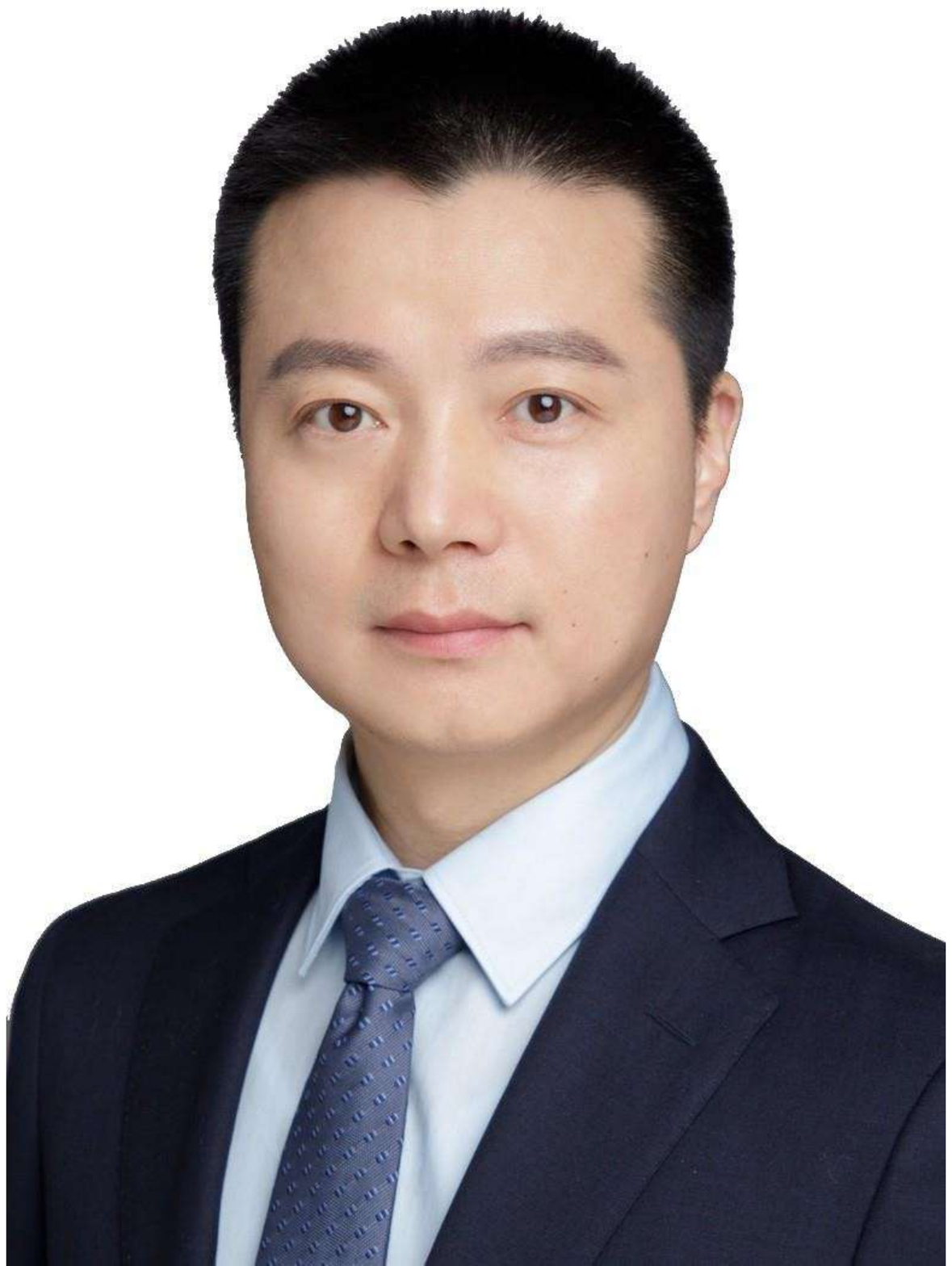}}]{Zhifeng Zhao}(Member, IEEE) received the B.E. degree in computer science, the M.E. degree in communication and information systems, and the Ph.D. degree in communication and information systems from the PLA University of Science and Technology, Nanjing, China, in 1996, 1999, and 2002, respectively. From 2002 to 2004, he acted as a Post-Doctoral Researcher with Zhejiang University, Hangzhou, China, where his researches were focused on multimedia next-generation networks (NGNs) and softswitch technology for energy efficiency. Currently, he is with the Zhejiang Lab, Hangzhou as the Chief Engineering Officer. His research areas include software defined networks (SDNs), wireless network in 6G, computing networks, and collective intelligence. He is the Symposium Co-Chair of ChinaCom 2009 and 2010. He is the Technical Program Committee (TPC) Co-Chair of the 10th IEEE International Symposium on Communication and Information Technology (ISCIT 2010).
\end{IEEEbiography}
\clearpage

\begin{appendices}
\section{Theoretical Proofs}
\subsection{Proof of Theorem \ref{thm:PCG} (Bounded Error of Parameterized Conjecture)}
\label{app:proofPCG}
Let $Q\subscriprbestpi(o\subscripttr,a\subscripttr,a\subscriptotheragent)$ denote the true optimal joint action-value function, and $Q\subscriprbestpi(o\subscripttr,a\subscripttr,\tilde{a}\subscriptotheragent)$ the one induced by a parameterized generator, where $\tilde{a}\subscriptotheragent \sim \bestgenotherspi$. We define the policy distribution error via the Kullback–Leibler (KL) divergence $D_{\rm KL}(\bestotherspi||\bestgenotherspi)$. Under Assumption \ref{assumpsion:Lipschitz}, the $Q$-value approximation error as Eq. \eqref{eq:Q-estimate-error}. Notably, (a) the approximation error is reformulated as the expected $Q$-value difference under two policies; (b) the error is decomposed into the function approximation error and the policy estimation error; (c) the boundedness of the function approximation error and the Lipschitz continuity of the $Q$-function in Assumption \ref{assumpsion:Lipschitz} are utilized; (d) the equality $d_A(a,a^\prime)=\mathbb{I}_{a\neq a^\prime}$ holds due to the discrete and finite action space, where $d(\cdot)$ is distance metric and $\mathbb{I}$ denotes indicator function; (e) Lemma \ref{lemma:Pinsker} is applied; and (f) by setting $C = L\sqrt{1/2}$ and assuming sufficiently small $\varkappa_{\rm approx}$, the KL divergence is shown to dominate the overall error.

\begin{lemma}
\label{lemma:Pinsker}
(Pinsker’s Inequality) Let $\mu$ and $\nu$ be two probability distributions defined over a universe $\mathcal{U}$. Then, the total variation distance between $\mu$ and $\nu$ is bounded by the KL divergence as follows
\begin{equation}
    \label{eq:pinsker}
    \| \mu - \nu \|_{\rm TV} \leq \frac{1}{2} D_{\rm KL}(\mu||\nu),
\end{equation}
where $|| \cdot||_{\rm TV}$ is total variation distance.
\end{lemma}
\begin{assumption}
\label{assumpsion:Lipschitz}
 (Lipschitz Continuity of $Q$-function).
Assuming that the $Q$-function satisfies Lipschitz continuous in the action space. That is, there exists a constant $L > 0$ such that for all $a\subscriptotheragent,\tilde{a}\subscriptotheragent\in \mathcal{A}_{-r}$, we have
\begin{equation}
|Q(o\subscripttr,a\subscripttr,a\subscriptotheragent) - Q(o\subscripttr,a\subscripttr,\tilde{a}\subscriptotheragent)| \leq L \cdot d_{A}(a\subscripttr,\tilde{a}\subscriptotheragent),
\end{equation}
where $d_{A}(\cdot,\cdot)$ is the distance metric in action space and $L$ is the Lipschitz constant.
\end{assumption}

This assumption is standard for $Q$-functions approximated by neural networks, as such networks are inherently Lipschitz continuous when built with common activations and practically controlled via regularization \cite{Gogianu2021SpectralNF}.
\subsection{Proof of Lemma \ref{lemma1} (Contraction Mapping Property)}
\label{app:proof_lemma1}
Beforehand, without loss of generality, a metric space can be represented as an ordered pair $(X, d)$, where $X$ is a set and $d$ is a metric defined on $X$. A contraction mapping $\mathscr{T}$ on a metric space $(X, d)$ satisfies $ d(\mathscr{T}(x), \mathscr{T}(y)) \leq \theta d(x, y) $ for any $ x, y \in X $, where $ 0 \leq \theta < 1 $.

Recalling the definition in Section \ref{sec:proof_convergence}, for each $o \in \mathcal{O}$, the operator $\mathscr{H}^P$ maps a generic function $Q\in \mathbb{R}$ to $\mathscr{H}^PQ\in \mathbb{R}$. Therefore, if we choose the metric space $(X, \| \cdot \|_\infty)$, where $X\in \mathbb{R}^{|\mathcal{O}||\mathcal{A}|}$, we have Eq. \eqref{eq:constract}. Notably, the inequality $(a)$ of Eq. \eqref{eq:constract} comes from the extraction and maximization over the remaining joint actions to obtain an upper bound. The equality $(b)$ utilizes the property that ${\Lambda \left( o_{t+1}|o_t,\boldsymbol{a} \right)}$ is a probability distribution and thus sums to 1. 
\begin{figure*}[b]
\hrulefill
    \begin{equation}	
     		\label{eq:Q-estimate-error}
     		\begin{aligned}
     			\varkappa &= \mathbb{E}_{\bestgenotherspi}\left[ \tilde{Q}\subscriprbestpi(o\subscripttr,a\subscripttr,\tilde{a}\subscriptotheragent) \right] - \max_{a\subscriptotheragent}Q\subscriprbestpi(o\subscripttr,a\subscripttr,a\subscriptotheragent)\\
     			&\stackrel{(a)}{=}\underbrace{\mathbb{E}_{\bestgenotherspi}\left[ \tilde{Q}\subscriprbestpi(o\subscripttr,a\subscripttr,\tilde{a}\subscriptotheragent) \right]}_{\text{estimated value}} - \underbrace{\mathbb{E}_{\bestotherspi}\left[Q\subscriprbestpi(o\subscripttr,a\subscripttr,a\subscriptotheragent)\right]}_{\text{true optimal value}}\\
     			&\stackrel{(b)}{\leq}\underbrace{\lvert \mathbb{E}_{\bestgenotherspi}\left[\tilde{Q}\subscriprbestpi(o\subscripttr,a\subscripttr,\tilde{a}\subscriptotheragent)-Q\subscriprbestpi(o\subscripttr,a\subscripttr,\tilde{a}\subscriptotheragent) \right] \rvert}_{\text{functional approximation error}} +\underbrace{\lvert \mathbb{E}_{\bestgenotherspi}\left[Q\subscriprbestpi(o\subscripttr,a\subscripttr,\tilde{a}\subscriptotheragent)\right]-\mathbb{E}_{\bestotherspi}\left[Q\subscriprbestpi(o\subscripttr,a\subscripttr,a\subscriptotheragent) \right] \rvert}_{\text{policy approximation error}}\\
     			&\stackrel{(c)}{\leq}\varkappa_{\rm approx} + L\cdot \mathbb{E}_{a\subscripttr \sim \bestotherspi,\tilde{a}\subscripttr \sim \bestgenotherspi}\left[d_A(a\subscripttr,\tilde{a}\subscripttr) \right] 
     			\stackrel{(d)}{\leq} \varkappa_{\rm approx} +L\cdot \|\bestotherspi-\bestgenotherspi \|_{\rm TV}\\
     			&\stackrel{(e)}{\leq} \varkappa_{\rm approx} +L\cdot \sqrt{\frac{1}{2} D_{\rm KL}(\bestotherspi||\bestgenotherspi)}
     			\stackrel{(f)}{=} C \cdot \sqrt{D_{\rm KL}\left( \bestgenotherspi || \bestotherspi\right)}
     		\end{aligned}
     	\end{equation}
    \begin{equation}
    \label{eq:constract}
    \begin{aligned}
        \parallel\mathscr{H} ^PQ_1-\mathscr{H} ^PQ_2\parallel_\infty 
        &=\mathop {\max}_{o,a_r}\,\Big\vert{\rm rwd}+\gamma \,\sum\nolimits_{o^{\prime}\in \mathcal{O}}{\Lambda (o^{\prime}|o,\boldsymbol{a})\mathop {\max}_{a_{-r}}\,Q_1(o^{\prime},a_r,a_{-r})}\\
        &\qquad\qquad\qquad -{\rm rwd}-\gamma \sum\nolimits_{o^{\prime}\in \mathcal{O}}{\Lambda (o^{\prime}|o,\boldsymbol{a})\mathop {\max}_{a_{-r}}\,Q_2(o^{\prime},a_r,a_{-r})}\Big\vert\\
        &=\mathop {\max}\nolimits_{o,a_r}\,\,\gamma \Big\vert\sum_{o^{\prime}\in \mathcal{O}}\left\{\Lambda \left( o^{\prime}|o,\boldsymbol{a} \right) \left[ \mathop {\max}_{a_{-r}}Q_1(o^{\prime},a_r,a_{-r}) -\mathop {\max}_{a_{-r}}Q_2(o^{\prime},a_r,a_{-r}) \right]\right\}\Big\vert\\
        &\stackrel{(a)}{\le} \mathop {\max}_{o,a_r,a_{-r}}\,\,\gamma \big\vert\sum\nolimits_{o^{\prime}\in \mathcal{O}}{\Lambda \left( o^{\prime}|o,\boldsymbol{a} \right) \left[ Q_1(o^{\prime},\tilde{\boldsymbol{a}})-Q_2(o^{\prime},\tilde{\boldsymbol{a}}) \right]}\big\vert\\
        &\le \mathop {\max}_{o,\boldsymbol{a}}\,\,\gamma \sum\nolimits_{o^{\prime}\in \mathcal{O}}{\Lambda \left( o^{\prime}|o,\boldsymbol{a} \right)}|Q_1(o^{\prime},\tilde{\boldsymbol{a}})-Q_2(o^{\prime},\tilde{\boldsymbol{a}})|\\
        &\le \mathop {\max}_{o,\boldsymbol{a}}\,\,\gamma \sum\nolimits_{o^{\prime}\in \mathcal{O}}{\Lambda \left( o^{\prime}|o,\boldsymbol{a} \right)}\mathop {\max}_{\tilde{o},\tilde{\boldsymbol{a}}}|Q_1(\tilde{o},\tilde{\boldsymbol{a}})-Q_2(\tilde{o},\tilde{\boldsymbol{a}})|\\
        &\stackrel{(b)}{=}\gamma \mathop {\max}_{\tilde{o},\tilde{\boldsymbol{a}}}|Q_1(\tilde{o},\tilde{\boldsymbol{a}})-Q_2(\tilde{o},\tilde{\boldsymbol{a}})|\\
        &=\gamma \,\,\parallel Q_1-Q_2\parallel_\infty\\
    \end{aligned}
\end{equation}
\end{figure*}

According to the contraction mapping theorem \cite{banahe}, we can conclude that Eq. \eqref{Pareto-alo} has a unique fixed point, which means the optimal action-value function $\mathscr{H} ^PQ^{\ast}=Q^{\ast}$.

\subsection{Proof of Theorem \ref{thm:convergence} (Convergence of the $Q$-function)}
\label{app:convergenceproof}
Assumption \ref{assump:finite_state} confirms finite observation space, aligning with Lemma \ref{lemma2}'s Assumption 2. We will now address proving Assumptions 3 and 4 from Lemma \ref{lemma2}.

Let $\mathscr{F}_t$ denote the $\sigma$-fields generated by all random variables in the stochastic game history up to the $t$-th round. Therefore, for any $\left( o\subscripttr,\jointaction\right)$,
\begin{equation}
	\begin{aligned}
		&\mathbb{E} \left[ F_t\left( o\subscripttr,\jointaction \right) |\mathscr{F} _t \right] \\
		&=\!\mathbb{E} \left[ {\rm rwd}\subscripttr\!+\!\gamma \mathop {\max}_{\otheraction}Q\!\subscripttr\left( o\subscripttr[t+1],a\subscripttr[t+1],\otheraction \right) \!-\!Q\subscriptrstar\!\left( o\subscripttr,\jointaction \right) \right]\\
		&=\!\mathbb{E} \left[ {\rm rwd}\subscripttr\!+\!\gamma \mathop {\max}_{\otheraction}Q\subscripttr\!\left( o\subscripttr[t+1],a\subscripttr[t+1],\otheraction \right) \right] \!-\!Q\subscriptrstar\!\left( o\subscripttr,\jointaction \right)\\
		&=\!\mathscr{H} ^PQ\subscripttr\left( o\subscripttr,\jointaction \right) -Q\subscriptrstar\left( o\subscripttr,\jointaction \right)\\
		&=\!\mathscr{H} ^PQ\subscripttr\left( o\subscripttr,\jointaction \right) -\mathscr{H} ^PQ\subscriptrstar\left( o\subscripttr,\jointaction \right).\\
	\end{aligned}
\end{equation}
Furthermore, from Lemma 1, we know that the operator $\mathscr{H}^P$ acting on $Q$ is a $\gamma$-contraction mapping. Therefore,
\begin{equation}
\label{eq:exp bound}
	\begin{aligned}
		&\parallel \mathbb{E} \left[ F_t\left( o\subscripttr,\jointaction \right) |\mathscr{F} _t \right] \parallel _{\infty}\\
		&=\mathop {\max}_{o\subscripttr,\jointaction}|\mathscr{H} ^PQ\subscripttr\left( o\subscripttr,\jointaction \right) -\mathscr{H} ^PQ\subscriptrstar\left( o\subscripttr,\jointaction \right) |\\
		&=\left\| \mathscr{H} ^PQ\subscripttr-\mathscr{H} ^PQ\subscriptrstar \right\| _{\infty}\\
		&\le \gamma \left\| Q\subscripttr-Q\subscriptrstar \right\| _{\infty}\\
		&\le \gamma \left\| \varDelta _t \right\| _{\infty}.\\
	\end{aligned}
\end{equation}
Therefore, the third assumption of Lemma \ref{lemma2} is satisfied. 

We now proceed to prove the final assumption regarding the bounded variance. From Assumption \ref{assump:bounded_reward}, we know that the reward function is bounded, $\left\| Q\subscriptrstar \right\| _{\infty}$ is bounded and strictly positive. Specifically, we have Eq. \eqref{eq:var}.
\begin{figure*}
    \begin{equation}
    \label{eq:var}
	\begin{aligned}
		&\operatorname{var}\left[ F_t\left( o\subscripttr,\jointaction \right) |\mathscr{F} _t \right]=\mathbb{E} \left[ \left( {\rm rwd}\subscripttr+\gamma \mathop {\max}_{\otheraction}\,\,Q\subscripttr\left( o\subscripttr[t+1],a\subscripttr[t+1],\otheraction \right) -\mathbb{E} \left[ F_t\left( o\subscripttr,\jointaction \right) |\mathscr{F} _t \right] \right) ^2 \right]\\
		&=\mathbb{E} \left[ \left( {\rm rwd}\subscripttr+\gamma \mathop {\max}_{\otheraction}\,\,Q\subscripttr\left( o\subscripttr[t+1],a\subscripttr[t+1],\otheraction \right)-Q\subscriptrstar\left( o\subscripttr,\jointaction \right) -\left( \mathscr{H} ^PQ\subscripttr\left( o\subscripttr,\jointaction \right) -Q\subscriptrstar\left( o\subscripttr,\jointaction \right) \right) \right) ^2 \right]\\
		&=\mathbb{E} \left[ \left( {\rm rwd}\subscripttr+\gamma \mathop {\max}_{\otheraction}\,\,Q\subscripttr\left( o\subscripttr[t+1],a\subscripttr[t+1],\otheraction \right)-\left( {\rm rwd}\subscripttr+\gamma \mathbb{E} \left[ \max_{\otheraction} Q\subscripttr\left( o_{t+1},\selfaction,\otheraction \right) \right] \right) \right) ^2 \right]\\
		&=\gamma ^2\mathbb{E} \left[ \left( \mathop {\max}_{\otheraction}\,\,Q\subscripttr\left( o\subscripttr[t+1],a\subscripttr[t+1],\otheraction \right) -\mathbb{E}\left[ \max_{\otheraction} Q\subscripttr\left( o_{t+1},\selfaction,\otheraction \right) \right] \right) ^2 \right]\\
		&=\gamma ^2\operatorname{var}\left[ \mathop {\max}_{\otheraction}\,\,Q\subscripttr\left( o\subscripttr[t+1],a\subscripttr[t+1],\otheraction \right) \right]\\
		&=\gamma ^2\mathbb{E} \left[ \left( \mathop {\max}_{\otheraction}\,\,Q\subscripttr\left( o\subscripttr[t+1],a\subscripttr[t+1],\otheraction \right) \right) ^2 \right]-\gamma ^2\left( \mathbb{E} \left[ \max_{\otheraction} Q\subscripttr\left( o_{t+1},\selfaction,\otheraction \right) \right] \right) ^2\\
		&\le \gamma ^2\mathbb{E} \left[ \left( \mathop {\max}_{\otheraction}\,\,Q\subscripttr\left( o\subscripttr[t+1],a\subscripttr[t+1],\otheraction \right) \right) ^2 \right]\le \gamma ^2\mathop {\max}_{o\subscripttr,\selfaction}\mathop {\max}_{\otheraction}\left( \,\,Q\subscripttr\left( o\subscripttr,\boldsymbol{a} \right) \right) ^2\le \gamma ^2\left\| \varDelta _t+Q\subscriptrstar \right\| _{\infty}^{2}\\
		&=\gamma ^2\left\| \varDelta _t \right\| _{\infty}^{2}+2\gamma ^2\left\| \varDelta _t \right\| _{\infty}\left\| Q\subscriptrstar \right\| _{\infty}+\gamma ^2\left\| Q\subscriptrstar \right\| _{\infty}^{2}\\
	\end{aligned}
\end{equation}

    \begin{equation}
    \label{eq:revised varience F}
    \begin{aligned}
    \operatorname{Var}\left(F_t \mid \mathscr{F}_t\right) & \leq \mathbb{E}\left[\left(\delta^{(\tau)}\right)^2 \mid \mathscr{F}_t\right] \\
    & =\tau^2 \mathbb{E}\left[\delta_{+}^2 \mid \mathscr{F}_t\right]+2 \tau(1-\tau) \mathbb{E}\left[\delta_{+} \delta_{-} \mid \mathscr{F}_t\right]+(1-\tau)^2 \mathbb{E}\left[\delta_{-}^2 \mid \mathscr{F}_t\right]\\
    & \stackrel{(a)}{\leq} \tau^2 \mathbb{E}\left[\delta_{+}^2 \mid \mathscr{F}_t\right]+\tau(1-\tau)\left(\mathbb{E}\left[\delta_{+}^2 \mid \mathscr{F}_t\right]+\mathbb{E}\left[\delta_{-}^2 \mid \mathscr{F}_t\right]\right)+(1-\tau)^2 \mathbb{E}\left[\delta_{-}^2 \mid \mathscr{F}_t\right] \\
    & =\left(\tau^2+\tau(1-\tau)\right) \mathbb{E}\left[\delta_{+}^2 \mid \mathscr{F}_t\right]+\left((1-\tau)^2+\tau(1-\tau)\right) \mathbb{E}\left[\delta_{-}^2 \mid \mathscr{F}_t\right] \\
    & =\tau \mathbb{E}\left[\delta_{+}^2 \mid \mathscr{F}_t\right]+(1-\tau) \mathbb{E}\left[\delta_{-}^2 \mid \mathscr{F}_t\right]\\
    &\stackrel{(b)}{\leq}\tau \cdot \left[U\left(1+\left\|Q_t-Q^*\right\|_{\infty}\right)\right]^2+(1-\tau) \cdot \left[U\left(1+\left\|Q_t-Q^*\right\|_{\infty}\right)\right]^2 \\
    &\leq \tau \cdot U^2\left(1+\left\|Q_t-Q^*\right\|_{\infty}^2\right)+(1-\tau) \cdot U^2\left(1+\left\|Q_t-Q^*\right\|_{\infty}^2\right) \\
    & = U\left(1+\left\|Q_t-Q^*\right\|_{\infty}^2\right) \\
    \end{aligned}
    \end{equation}  
 \hrulefill
\end{figure*}

We now analyze two different scenarios for the value of$\| \varDelta_t\|_\infty$:
\begin{itemize}
	\item $\left\| \varDelta _t \right\| _{\infty}\le 1$, then
	$$
	\begin{aligned}
		\gamma ^2\left\| \varDelta _t \right\| _{\infty}^{2}&+2\gamma ^2\left\| \varDelta _t \right\| _{\infty}\left\| Q\subscriptrstar \right\| _{\infty}+\gamma ^2\left\| Q\subscriptrstar \right\| _{\infty}^{2}\\
		&\le \gamma ^2\left\| \varDelta _t \right\| _{\infty}^{2}+2\gamma ^2\left\| Q\subscriptrstar \right\| _{\infty}+\gamma ^2\left\| Q\subscriptrstar \right\| _{\infty}^{2}
	\end{aligned}
	$$
	
	\item $\left\| \varDelta _t \right\| _{\infty}>1$, so $\left\| \varDelta _t \right\| _{\infty}\le \left\| \varDelta _t \right\| _{\infty}^{2}$, then
	$$
	\begin{aligned}
			\gamma ^2\left\| \varDelta _t \right\| _{\infty}^{2}&+2\gamma ^2\left\| \varDelta _t \right\| _{\infty}\left\| Q\subscriptrstar \right\| _{\infty}+\gamma ^2\left\| Q\subscriptrstar \right\| _{\infty}^{2}\\
			&\le \gamma ^2\left( 1+2\gamma ^2\left\| Q\subscriptrstar \right\| _{\infty} \right) \left\| \varDelta _t \right\| _{\infty}^{2}+\gamma ^2\left\| Q\subscriptrstar \right\| _{\infty}^{2}
	\end{aligned}
$$
\end{itemize}
Consequently, we can select a constant $U$, which satisfies 
$$
U=\max \left\{ \gamma ^2+2\gamma ^2\left\| Q\subscriptrstar \right\| _{\infty},\gamma ^2\left( \left\| Q\subscriptrstar \right\| _{\infty}^{2}+2\left\| Q\subscriptrstar \right\| _{\infty} \right) \right\},$$ such that
$$
\begin{aligned}
	&\operatorname{var}\left[ F_t\left( o\subscripttr,\jointaction \right) |\mathscr{F} _t \right] \\
	=&\gamma ^2\left\| \varDelta _t \right\| _{\infty}^{2}+2\gamma ^2\left\| \varDelta _t \right\| _{\infty}\left\| Q\subscriptrstar \right\| _{\infty}+\gamma ^2\left\| Q\subscriptrstar \right\| _{\infty}^{2}
	\\
	\le & U\left( 1+\left\| \varDelta _t \right\| _{\infty}^{2} \right) 
\end{aligned}
$$
Thus, we have proven that all assumptions of Lemma \ref{lemma2} are satisfied. Therefore, $\varDelta_t$ converges to $0$ with probability $1$, which implies that $\boldsymbol{Q}_t=\left[ Q\subscriptproofrt[1],\cdots ,Q\subscriptproofrt[R] \right] $ converges to $\boldsymbol{Q}^\ast$ with probability $1$.

Considering Assumption \ref{assump:same_ne} that in each stage of the game, all optimal points/saddle points have the same Nash value and the agents are homogeneous, the action-value function will gradually converge to the Nash value, i.e., $\mathscr{H} ^P\boldsymbol{Q}^{\ast}=\boldsymbol{Q}^{\ast}=\mathscr{H} ^N\boldsymbol{Q}^{\ast}$.
\subsection{Proof of Corollary \ref{thm:expectile} (Convergence with Expectile Regression)}
\label{app:expectile}
To establish the convergence of the $Q$-function under heterogeneous expectile regression, we define the modified Pareto operator with $\tau$ as Eq. \eqref{eq:expectile pareto operator}.
The update of the $Q$ function with the introduction of expectile coefficient can be rewritten as
\begin{equation}
    \label{eq: expectile q update}
Q\subscripttr[t+1]\left(o\subscripttr,\jointaction\right)=Q\subscripttr\left(o\subscripttr,\jointaction\right)+ \alpha \mathbb{E}\left[ \tau_r \delta_{+} + (1-\tau_r) \delta_{-}\right]
\end{equation}
We continue our analysis based on Lemma \ref{lemma2}, first examining the conditional expectation of the TD error $\delta^{(\tau)}$:
\begin{equation}
    \begin{aligned}
    \mathbb{E}[F_t \mid \mathscr{F}_t] & =\mathbb{E}\left[\delta^{(\tau)} \mid \mathscr{F}_t\right] \\
    & =\tau \mathbb{E}\left[\delta_{+} \mid \mathscr{F}_t\right]+(1-\tau) \mathbb{E}\left[\delta_{-} \mid \mathscr{F}_t\right] \\
    & \leq \tau E\left[\left|\delta_{+}\right| \mid \mathscr{F}_t\right]+(1-\tau) E\left[\left|\delta_{-}\right| \mid \mathscr{F}_t\right] \quad  \\
    & =E\left[\tau\left|\delta_{+}\right|+(1-\tau) | \delta_{-}| \mid \mathscr{F}_t\right] \\
    & \leq E\left[(\tau+(1-\tau)) \cdot \max \left(\left|\delta_{+}\right|,\left|\delta_{-}\right|\right) \mid \mathscr{F}_t\right]   \\
    & = \mathbb{E}\left[\delta \mid \mathscr{F}_t\right] \stackrel{(a)}{\leq}\gamma\|\varDelta_t\|_\infty,
    \end{aligned}
\end{equation}
where $(a)$ is derived from the proof of Theorem \ref{thm:convergence}, as shown by Eq. \eqref{eq:exp bound}. We now fulfill the expectation condition in Lemma \ref{lemma2} and discuss the variance condition next. The variance of $F_t$ under expectile regression is bounded by Eq. \eqref{eq:revised varience F}, where $(a)$ follows from Yang's inequality, while $(b)$ results from the boundedness of the reward assumed in Assumption \ref{assump:bounded_reward}. A detailed analysis of $(b)$ is provided below.
\begin{equation}
    \begin{aligned}
        |\delta_+|,|\delta_-| &\leq {\rm rwd_{max}}+\|Q_t\|_\infty+\gamma\|Q_t\|_\infty\ \\
        &\leq{\rm rwd_{max}}+(\gamma+1)(\|Q_t-Q^\ast\|_\infty + \|Q^\ast\|_\infty)\\
        &=\underbrace{{\rm rwd_{max}}+(\gamma+1)\left\|Q^*\right\|_{\infty}}_{\text{constant term}\ U_1}+\underbrace{(\gamma+1)}_{U_2}\left\|Q_t-Q^*\right\|_{\infty} \\
        &\leq U\left(1+\left\|Q_t-Q^*\right\|_{\infty}\right),\ U=\max\{U_1,U_2\}
    \end{aligned}
\end{equation}
We have thus verified all the conditions required by Lemma \ref{lemma2} from stochastic approximation theory, thereby obtaining the same convergence result as stablished in Theorem \ref{thm:convergence}.
\section{Experimental Setup and Implementation Details}
\label{app:setup}
\subsection{HVI Calculation Details}
\label{app:hvi detail}
We now summarize the steps behind Eq. \eqref{eq:hvi}, where employ the \texttt{pygmo} library’s \texttt{hypervolume} routine to quantify each algorithm’s Pareto‐front quality over $R$ SP rewards. Given an algorithm's solution set as Eq. \eqref{eq:hvi solution}, we first normalize each SP reward dimension to $[0,1]$ by
            \begin{equation}
    		{\rm v}_{r, \mathrm{norm}}^{(m)}=\frac{{\rm rwd}_r^{(m)}-{\rm rwd}_r^{\min }}{{\rm rwd}_r^{\max }-{\rm rwd}_r^{\min }},
    	\end{equation}
    	where ${\rm rwd}_r^{\min}$, ${\rm rwd}_r^{\max}$ are taken over all algorithms and runs. To leverage the minimization-based \texttt{pygmo.hypervolume}, we invert against a fixed utopian point $\mathbf{v}^{\rm {ref}}=(1.1, \ldots, 1.1)$ to obtain 
    	$\mathbf{u}^{(m)}=\mathbf{v}^{\rm {ref }}-\mathbf{v}_{\rm {norm }}^{(m)}$ and compute the HVI as the $R$-dimensional Lebesgue volume dominated by the inverted solutions ($\texttt{pg.hypervolume}(\{\mathbf{u}^{(m)}\}). \texttt{compute}(\mathbf{v}^{\rm{ref}})$). A larger HVI thus confirms both closer proximity to the utopian ideal and a wider spread of trade-offs among SPs.
\subsection{Setups for Scalability and Generalization Experiments}
\label{app:sp_number}
We additionally introduce two datasets, QMNIST and SVHN, along with their corresponding models to evaluate the impact of different SP numbers.
For QMNIST, we apply a CNN consisting of two convolutional layers, followed by three fully connected layers. The input images are grayscale with a size of $28 \times 28$, and max-pooling is applied after each convolution to reduce spatial dimensions. For SVHN, we adopt a deeper CNN architecture with two convolutional layers incorporating Local Response Normalization (LRN) and max-pooling, followed by three fully connected layers. The input images are RGB with a size of $32 \times 32$, and the network uses ReLU activation throughout. The number of model parameters for the two tasks are approximately $439,550$ and $1,757,962$, respectively. Specifically, when $R = 2$, the SPs are assigned tasks on MNIST and FashionMNIST datasets. With $R = 3$, the task set expands to include MNIST, FashionMNIST, and CIFAR10. For $R = 4$, the tasks consist of MNIST, FashionMNIST, QMNIST, and SVHN. Finally, when $R = 5$, all the aforementioned datasets are involved.

We further present key performance metrics across different task models as Table \ref{tab:cross-model}. On one hand, task complexity constrains performance: the more complex CIFAR-10 yields significantly lower peak accuracy than other tasks. On the other, model complexity directly impacts computational and communication costs — for example, due to parameter differences, QMNIST incurs $2.5\times$–$4.2\times$ higher overhead than the similarly complex MNIST. These results also demonstrate that our algorithm can efficiently schedule tasks with varying complexity while balancing both individual and overall network performance.
\begin{table}[tbp]
  \centering
  \footnotesize
  \caption{A summary of key performance metrics across different models.}
    \renewcommand{\arraystretch}{1} 
    \setlength{\tabcolsep}{1.2pt} 
    \begin{threeparttable}
    \begin{tabular}{cccccc}
    \toprule
    Tasks & \makecell{Model \\ Parameters} & \makecell{Comm. \\ Overheads (MB)} & \makecell{Energy \\Cost (J)} & \makecell{Time \\Latency (s)} & \makecell{Max. \\Accuracy} \\
    \midrule
    CIFAR-10 & $9,074,474$ & $26.54$ & $18.18$ & $21.07$ & $79\%$ \\
    FashionMNIST & $21,840$ & $0.31$ & $0.7$  & $0.12$ & $91\%$ \\
    MNIST & $101,770$ & $1.48$ & $1.19$ & $0.34$ & $98\%$ \\
    QMNIST & $439,550$ & $6.25$ & $3.02$ & $1.21$ & $99\%$ \\
    SVHN & $1,757,962$ & $3.64$ & $2.89$ & $6.33$ & $89\%$ \\
    \bottomrule
    \end{tabular}%
    \label{tab:cross-model}
      \begin{tablenotes}   
    \footnotesize            
    \item[1] With $R = 5$ and \texttt{PAC-MCoFL-p}, we compare different task models by model size, average communication overheads, latency, energy cost, and peak accuracy under fixed FL rounds.
  \end{tablenotes}
  \end{threeparttable}   
\end{table}

\subsection{Implementation of MARL Baselines}
\label{app:marl}
The MAPPO baseline is implemented using the standard centralized training with decentralized execution (CTDE) paradigm. During decentralized execution, each agent's actor makes decisions based solely on its own local observation $o_{r,t}$. The centralized critic, active only during training, accesses the joint observations (the collection of all agents' local observations) to learn a global value function. Crucially, the agents act without conjecture of others' policies, relying on the critic to implicitly guide coordination.

The RSM-MASAC baseline assumes each SP can communicate directly with all other SPs and acts in a a fully decentralized manner. During collaboration rounds, SPs exchange segments of their policy network's parameters with each other \cite{rsmmasac}. Each SP uses a regulated segment mixture mechanism, on the basis of a theory-guided metric, to selectively combine the received parameter segments from other SPs to improve its own policy without a central coordinator \cite{rsmmasac}.

\section{Supplementary Results}
\label{app:supplementary_results}

\subsection{Computational and Deployment Overheads}
\label{app:results_overhead}
To contextualize the practical merits of our methods, we report key overhead metrics under the $R=3$ setting. As shown in Table \ref{tab:deployment}, all the methods exhibit similar deployment efficiency: decision latency is consistently low ($1.3$ ms), and GPU memory usage varies only slightly ($498$ MB for MAPPO vs. $490$ MB for \texttt{PAC-MCOFL-p}). In terms of training cost, \texttt{PAC-MCOFL-p} incurs a modest increase ($37.45$ ms/episode) compared to MAPPO ($29.83$ ms), reflecting the overhead of conjecture generation. But it saves significantly than \texttt{PAC-MCOFL}  ($48.25$ ms). In summary, \texttt{PAC-MCOFL-p} delivers superior performance with minimal overhead, achieving real-time efficiency comparable to MAPPO while offering greater learning capability.
\begin{table}[bt]
	\centering
	\caption{Computational and Deployment Overheads ($R=3$).}
	\label{tab:deployment}
	\footnotesize
	\begin{tabular}{@{\hspace{2pt}}cccc@{\hspace{2pt}}}
		\toprule
		\textbf{Metric} & \textbf{MAPPO} & \textbf{PAC-MCOFL} & \textbf{PAC-MCOFL-p} \\ 
		\midrule
		\makecell{Training Time\\per Episode} & $\approx 29.83$ ms & $\approx 48.25$ ms & $\approx 37.45$ ms \\
		Decision Latency & $\approx 1.3$ ms & $\approx 1.4$ ms & $\approx 1.3$ ms \\
		\makecell{GPU Memory} & $\approx 498$ MB & $\approx 2132$ MB & $\approx 490$ MB \\
		\bottomrule
	\end{tabular}
\end{table}
\subsection{Efficacy of the TCAD Mechanism}
\label{app:tcad ablation}
The core advantage of TCAD is that it decouples the decision-making complexity from the action granularity. It maintains a constant, low-complexity decision space ($3^4=81$ options) while enabling incremental, fine-grained adjustments to the continuous action variables.
A naive discretization approach, however, faces a direct trade-off between granularity and complexity. As shown in Table \ref{tab:tcad_complexity}, the complexity of the naive approach is a direct function of its granularity ${\rm level}_m$. Only when we set ${\rm level}_m=3$ for all four dimensions, its complexity (i.e., $O((3^4)^{R-1}) = O(81^{R-1})$) becomes identical to TCAD's. However, this equivalence comes at the cost of forcing the agent to make decisions from a very coarse action set, which can degrade performance. TCAD avoids this compromise by using a fixed set of incremental operators to navigate the action space, inherently allowing for much finer control without increasing the complexity of the conjecture operation. In the performance comparison shown in Fig. \ref{fig:tcad ablation}, to ensure fair comparison, each of the four action dimensions (client selection $n$, CPU frequency $f$, bandwidth $B$, and quantization level $q$) is divided into three levels (${\rm level}_n={\rm level}_f={\rm level}_B={\rm level}_q=3$).

\begin{table*}[tb]
    \centering
    \caption{Comparison of Action Space Structure and Complexity.}
    \label{tab:tcad_complexity}
    \begin{tabular}{l|c|c|c}
    \toprule
    \textbf{Method} & \textbf{Granularity (Levels per Dim.)} & \textbf{Action Space Size} & \textbf{Conjecture Complexity} \\
    \midrule
    Naive Discretization & ${\rm level}_m$ for each dimension $m$ & $|\mathcal{A}_{\text{naive}}| = \prod_{m=1}^{4} L_m$ & $O\left(|\mathcal{A}_{\text{naive}}|^{R-1}\right)$ \\
    \midrule
    \textbf{TCAD (Ours)} & Incremental (effectively fine-grained) & $|\mathcal{A}_{\text{TCAD}}| = 3^4 = 81$ & $O\left(81^{R-1}\right)$ \\
    \bottomrule
    \end{tabular}
\end{table*}

\subsection{Hyperparameter Sensitivity Analysis}
\label{app:results_hyperparams}
\subsubsection{Effects of Different Discount Factors and Learning Rates}
\label{app:effect of discount factor}
\begin{table*}[tb]
    \centering
    \caption{Effect of different learning rates and discount factors on system performance.}
    \renewcommand{\arraystretch}{0.7}
    \setlength{\tabcolsep}{3.5pt}
    \label{tab:hyperparams_extended}
    \begin{threeparttable}
    \begin{tabular}{@{} lll *{12}{c} @{}}
			\toprule
			\multicolumn{2}{c}{\multirow{2}{*}{\textbf{Hyperparameters}}} & \multirow{2}{*}{\textbf{Value}} & \multicolumn{4}{c}{\textbf{CIFAR-10}} & \multicolumn{4}{c}{\textbf{Fashion-MNIST}} & \multicolumn{4}{c}{\textbf{MNIST}} \\
			\cmidrule(lr){4-7} \cmidrule(lr){8-11} \cmidrule(lr){12-15}
			& & & \multicolumn{1}{c}{$\Acc^{\max}_{1,t}$} & \multicolumn{1}{c}{$\overline{ \rm vol}_{1,t}$} & \multicolumn{1}{c}{$\bar{T}_{1,t}$} & \multicolumn{1}{c}{$\bar{E}_{1,t}$} & \multicolumn{1}{c}{$\Acc^{\max}_{2,t}$} & \multicolumn{1}{c}{$\overline{ \rm vol}_{2,t}$} & \multicolumn{1}{c}{$\bar{T}_{2,t}$} & \multicolumn{1}{c}{$\bar{E}_{2,t}$} & \multicolumn{1}{c}{$\Acc^{\max}_{3,t}$} & \multicolumn{1}{c}{$\overline{ \rm vol}_{3,t}$} & \multicolumn{1}{c}{$\bar{T}_{3,t}$} & \multicolumn{1}{c}{$\bar{E}_{3,t}$} \\
			\midrule
			
			\multirow{10}{*}{Learning rate $\zeta$}
			& \texttt{PAC-MCOFL} & \multirow{2}{*}{$ 1 e^{-2}$} & 32.43\% & \sout{17.70} & \sout{11.25} & \sout{5.41} & 88.18\% & 0.13 & \textbf{0.11} & 0.15 & 96.66\% & \textbf{0.15} & 0.58 & 0.16 \\
			& \texttt{PAC-MCOFL-p} & & 38.11\% & \sout{26.25} & \sout{11.89} & \sout{5.73} & 88.55\% & 0.14 & 0.12 & 0.12 & 95.82\% & 0.16 & 0.62 & 0.18 \\
			\addlinespace
			& \texttt{PAC-MCOFL} & \multirow{2}{*}{$5e^{-3}$} & 75.97\% & 37.68 & 15.45 & 8.42 & 89.82\% & 0.19 & 0.81 & \textbf{0.06} & 96.51\% & 0.54 & 0.50 & \textbf{0.14} \\
			& \texttt{PAC-MCOFL-p} & & 72.52\% & 20.15 & 13.91 & 6.88 & 90.91\% & 0.21 & 0.85 & 0.07 & 95.96\% & 0.57 & 0.53 & 0.15 \\
			\addlinespace
			& \texttt{PAC-MCOFL} & \multirow{2}{*}{$1e^{-3}$} & \textbf{78.55\%} & 18.14 & \textbf{7.24} & 6.58 & \textbf{91.01}\% & 0.12 & 0.11 & 0.09 & \textbf{98.23\%} & 0.46 & \textbf{0.23} & 0.27 \\
			& \texttt{PAC-MCOFL-p} & & 77.13\% & \textbf{17.55} & 9.51 & 7.22 & 90.26\% & \textbf{0.10} & 0.12 & 0.10 & 97.88\% & 0.48 & 0.25 & 0.29 \\
			\addlinespace
			& \texttt{PAC-MCOFL} & \multirow{2}{*}{$5 e^{-4}$} & 68.75\% & 24.92 & 10.75 & \textbf{5.79} & 83.45\% & 0.98 & 0.12 & 0.15 & 97.62\% & 0.79 & 0.29 & 0.18 \\
			& \texttt{PAC-MCOFL-p} & & 67.22\% & 25.88 & 11.13 & 6.05 & 82.17\% & 1.05 & 0.13 & 0.16 & 97.15\% & 0.82 & 0.31 & 0.20 \\
			\addlinespace
			& \texttt{PAC-MCOFL} & \multirow{2}{*}{$1 e^{-5}$} & 46.45\% & \sout{20.03} & \sout{11.53} & \sout{6.78} & 50.46\% & \sout{0.17} & \sout{0.45} & \sout{0.24} & 97.45\% & 0.27 & 0.35 & \textbf{0.14} \\
			& \texttt{PAC-MCOFL-p} & & 45.21\% & \sout{20.95} & \sout{11.96} & \sout{7.03} & 49.15\% & \sout{0.18} & \sout{0.49} & \sout{0.26} & 96.93\% & 0.29 & 0.37 & 0.15 \\
			\midrule
			
			\multirow{8}{*}{Discount factor $\gamma$}
			& \texttt{PAC-MCOFL} & \multirow{2}{*}{$0.99$} & \textbf{78.21}\% & 22.96 & 6.62 & 9.96 & \textbf{91.42\%} & 0.16 & 0.13 & 0.15 & 95.46\% & \textbf{0.34} & 0.21 & \textbf{0.25} \\
			& \texttt{PAC-MCOFL-p} & & 77.92\% & \textbf{19.52} & 6.89 & 10.25 & 90.88\% & 0.17 & 0.14 & 0.16 & 94.95\% & 0.36 & 0.23 & 0.27 \\
			\addlinespace
			& \texttt{PAC-MCOFL} & \multirow{2}{*}{$0.90$} & 73.46\% & 24.63 & \textbf{6.11} & 10.34 & 87.56\% & 0.08 & \textbf{0.11} & 0.21 & \textbf{97.66}\% & 0.52 & 0.13 & 0.38 \\
			& \texttt{PAC-MCOFL-p} & & 72.15\% & 25.41 & 6.35 & 10.81 & 86.95\% & 0.09 & 0.12 & 0.23 & 97.02\% & 0.55 & 0.14 & 0.41 \\
			\addlinespace
			& \texttt{PAC-MCOFL} & \multirow{2}{*}{$0.85$} & 68.56\% & 48.83 & 20.51 & 6.14 & 75.66\% & \textbf{0.07} & 0.22 & \textbf{0.11} & 97.62\% & 0.71 & 0.14 & 0.26 \\
			& \texttt{PAC-MCOFL-p} & & 67.92\% & 49.95 & 21.03 & \textbf{5.52} & 76.02\% & 0.08 & 0.24 & 0.12 & 97.11\% & 0.75 & 0.17 & 0.29 \\
			\addlinespace
			& \texttt{PAC-MCOFL} & \multirow{2}{*}{$0.70$} & 42.31\% & \sout{19.97} & \sout{6.43} & \sout{15.68} & 81.87\% & 0.13 & 0.13 & 0.14 & 96.21\% & 0.35 & \textbf{0.09} & 0.61 \\
			& \texttt{PAC-MCOFL-p} & & 51.55\% & \sout{12.84} & \sout{10.21} & \sout{16.22} & 83.56\% & 0.14 & 0.14 & 0.15 & 95.68\% & 0.38 & 0.13 & 0.46 \\
			
			\bottomrule
		\end{tabular}
    \begin{tablenotes}
        \footnotesize
        \item[1] $\overline{{\rm vol}}\subscripttr,\overline{T}\subscripttr,\overline{E}\subscripttr$ which are in Mbits, s, and J, respectively, represent the average communication overheads, latency, and energy consumption consumed during FL training rounds.
        \item[2] Data with a strikethrough indicates that the task model does not eventually converge (accuracy still fluctuates significantly), and bolded data indicates an advantage in comparisons with the same metric.
    \end{tablenotes}
\end{threeparttable}
\end{table*}
We evaluate the sensitivity of two pivotal hyperparameters—learning rate $\zeta$ and discount factor $\gamma$—with detailed results for both \texttt{PAC-MCoFL} and \texttt{PAC-MCoFL-p} summarized in Table \ref{tab:hyperparams_extended}. The analysis reveals that the learning rate $\lr$ has a significant impact, with a clear optimal range emerging. Specifically, $\lr= 10^{-3}$ consistently achieves the best performance, yielding the highest model accuracies across all three tasks. In contrast, both excessively high $\lr = 10^{-2}$ and excessively low $\lr = 10^{-5}$ values can lead to oscillations or suboptimal solutions, potentially hindering convergence altogether, as seen in the CIFAR-10 results in Table \ref{tab:hyperparams_extended}. A similar trend holds for the discount factor $\gamma$, where higher values that emphasize long-term rewards, such as $0.99$ and $0.9$, result in superior outcomes, particularly for the more complex CIFAR-10 task. This underscores the value of foresight in our co-optimization scheme. Overall, we conclude that these hyperparameters have an acceptable range that is not overly narrow and can be effectively tuned through standard hyperparameter optimization.

\subsubsection{Effects of Different TCAD Granularity}
\label{app:tcad_granularity}
Table \ref{tab:tcad_granularity} reports the impact of varying each TCAD granularity hyperparameter on system performance. Altering $\varsigma_n\in\{1,2,3\}$, $\varsigma_f\in\{0.2,0.5,1.0\}\,\mathrm{GHz}$, $\varsigma_B\in\{0.5,2,4.0\}\,\mathrm{MHz}$ or $\varsigma_q\in\{2,4,8\}$ levels induces moderate performance oscillations, yet these remain within acceptable bounds. For instance, accuracy on CIFAR‑10, FashionMNIST and MNIST fluctuates by at most $7.9\%$, $4.9\%$ and $1.2\%$, respectively, while yielding comparable latency $\bar T$ and energy $\bar E$. Overall, with TCAD granularity set within a suitable range, \texttt{PAC‑MCoFL} delivers robust and stable performance.
\begin{table*}[tb]
    \centering
    \caption{Effect of Different TCAD Granularity ($\varsigma_m$) on System Performance.}
    \label{tab:tcad_granularity}
    \footnotesize 
    \setlength{\tabcolsep}{3pt} 
    \renewcommand{\arraystretch}{0.9} 
    \begin{tabular}{@{}lcccc|cccc|cccc|cccc@{}}
        \toprule
        \multirow{2}{*}{\textbf{Hyperparameters}} 
        & \multicolumn{4}{c|}{\textbf{TCAD Granularity}} 
        & \multicolumn{4}{c|}{\textbf{CIFAR-10} (Task $r_1$)} 
        & \multicolumn{4}{c|}{\textbf{Fashion-MNIST} (Task $r_2$)} 
        & \multicolumn{4}{c}{\textbf{MNIST} (Task $r_3$)} \\
        \cmidrule(lr){2-5} \cmidrule(lr){6-9} \cmidrule(lr){10-13} \cmidrule(lr){14-17}
        & $\varsigma_n$ & $\varsigma_f$ (GHz) & $\varsigma_B$ (MHz) & $\varsigma_q$ (level)
        & $ \Gamma^{\max}_{1,t}$ & $\overline{\rm vol}_{1,t}$ & $\overline{T}_{1,t}$ & $\overline{E}_{1,t}$
        & $ \Gamma^{\max}_{2,t}$ & $\overline{\rm vol}_{2,t}$ & $\overline{T}_{2,t}$ & $\overline{E}_{2,t}$
        & $ \Gamma^{\max}_{3,t}$ & $\overline{\rm vol}_{3,t}$ & $\overline{T}_{3,t}$ & $\overline{E}_{3,t}$ \\
        \midrule
        \textbf{BASELINE}
        & 1   & 0.5 & 2.0 & 4
        & 78.60\% & 18.74 & 6.62 & 9.96
        & 91.07\% & 0.15  & 0.13 & 0.16
        & 98.20\% & 0.34  & 0.21 & 0.25 \\
        \midrule
        \multirow{2}{*}{Granularity of $\varsigma_n$}
        & 3   & 0.5 & 2.0 & 4
        & 77.22\% & 24.15 & 7.43 & 10.88
        & 90.87\% & 0.21  & 0.25 & 0.16
        & 97.41\% & 0.25  & 0.28 & 0.14 \\
        & 2   & 0.5 & 2.0 & 4
        & 78.14\% & 20.41 & 7.02 & 10.21
        & 89.35\% & 0.15  & 0.14 & 0.08
        & 97.93\% & 0.31  & 0.19 & 0.32 \\
        \midrule
        \multirow{2}{*}{Granularity of $\varsigma_f$}
        & 1   & 1.0 & 2.0 & 4
        & 77.16\% & 19.53 & 7.95 & 14.33
        & 91.25\% & 0.14  & 0.11 & 0.32
        & 97.81\% & 0.33  & 0.09 & 0.21 \\
        & 1   & 0.2 & 2.0 & 4
        & 76.83\% & 18.43 & 9.26 & 11.62
        & 91.01\% & 0.15  & 0.13 & 0.15
        & 97.33\% & 0.27  & 0.23 & 0.12 \\
        \midrule
        \multirow{2}{*}{Granularity of $\varsigma_B$}
        & 1   & 0.5 & 4.0 & 4
        & 75.51\% & 22.48 & 8.51 & 10.43
        & 91.83\% & 0.18  & 0.09 & 0.11
        & 97.24\% & 0.39  & 0.31 & 0.30 \\
        & 1   & 0.5 & 0.5 & 4
        & 76.20\% & 19.03 & 7.05 & 10.27
        & 89.79\% & 0.15  & 0.14 & 0.16
        & 98.11\% & 0.33  & 0.26 & 0.28 \\
        \midrule
        \multirow{2}{*}{Granularity of $\varsigma_q$}
        & 1   & 0.5 & 2.0 & 8
        & 72.32\% & 27.43 & 12.81 & 11.05
        & 86.54\% & 0.19  & 0.11 & 0.09
        & 96.97\% & 0.21  & 0.28 & 0.15 \\
        & 1   & 0.5 & 2.0 & 2
        & 74.13\% & 16.11 & 9.65 & 7.85
        & 90.51\% & 0.15  & 0.28 & 0.27
        & 98.42\% & 0.54  & 0.42 & 0.17 \\
        \bottomrule
    \end{tabular}%
\end{table*}

\subsection{Generality in a Mixed-Task Scenario}
    To evaluate the cross-domain effectiveness of our framework, a heterogeneous service environment is configured, wherein the original CIFAR-10 task is replaced by a next-token prediction (NTP) task on the Penn Treebank (PTB) dataset, implemented with a Recurrent Neural Network (RNN)\footnote{We employ a simply RNN model with a vocabulary of approximately $11,000$ tokens. The model is composed of an embedding layer to process the input vocabulary, a single-layer RNN backbone to capture sequential dependencies, and a final linear layer to decode the output into a probability distribution over the vocabulary, totaling approximately $2.3$ million parameters. Critically, to align the reward signal with this task's objective, the framework's performance metric is accordingly adapted from accuracy $\Gamma$ to perplexity.}. All other experimental conditions remained consistent with those previously detailed. The empirical results, summarized in Table \ref{tab:nlp}, reveal a distinct performance advantage for \texttt{PAC-MCOFL} over the MAPPO baseline, as evidenced by its superiority in both total reward ($150.3$ versus $147.21$) and the HVI ($0.3855$ versus $0.3358$). Both \texttt{PAC-MCOFL} and its scalable variant \texttt{PAC-MCOFL-P} exhibit competitive perplexity on the NTP task while maintaining the performance of the CNN-based services. This case study effectively demonstrates the intrinsic adaptability of our proposed framework to task heterogeneity.
\begin{table}[tb]
  \centering
    \setlength{\tabcolsep}{3pt} 
    \renewcommand{\arraystretch}{1.2} 
  \caption{Performance Comparison in the Mixed-Task Scenario}
    \begin{tabular}{ccccc}
    \toprule
    Tasks & Metric & MAPPO & PAC-MCOFL & PAC-MCOFL-p \\
    \midrule
    \multirow{2}[2]{*}{PTB} & $\overline{\rm rwd}$  & -12.98 & -11.68 & -17.68 \\
         & Ppl  & 302.77 & 294.32 & 321.56 \\
    
    \multirow{2}[2]{*}{FashionMNIST} & $\overline{\rm rwd}$  & 75.46 & 73.46 & 70.39 \\
         & Acc  & 89.12 & 89.87 & 87.64 \\
    
    \multirow{2}[2]{*}{MNIST} & $\overline{\rm rwd}$  & 84.73 & 88.52 & 86.16 \\
         & Acc  & 97.81 & 97.77 & 96.69 \\
    
    \midrule
    \multirow{2}[2]{*}{Total} & ${\rm rwd}^{\rm total}$ & 147.21 & 150.3 & 138.87 \\
         & HVI  & 0.3358 & 0.3855 & 0.2698 \\
    \bottomrule
    \multicolumn{5}{l}{\footnotesize{Acc (\%): Accuracy (higher is better); Ppl: Perplexity (lower is better).}}
    \end{tabular}%
  \label{tab:nlp}%
\end{table}%
\section{Supplementary Discussions}
\label{app:discussion}
\subsection{On Information Provisioning in the Non-Cooperative Game}
\label{app:discuss_info}
\subsubsection{Implementation Considerations of Action Sharing}
\label{appx:act share}
Each SP’s action vector (client selection, quantization level, bandwidth allocation, and CPU frequency) requires less than $20$ bytes per round using float32 encoding, which is negligible compared to model updates (e.g., $66.84$ MB for \texttt{FedAvg} method without quantization in Table IV). Quantitatively, sharing of historic actions constitutes less than $0.01\%$ of the total traffic (model parameters \& actions), as exchanges of FL model gradients dominate bandwidth usage in Eq. (7).
\subsubsection{Game-Theoretic Rationale for Information Provisioning}
To ground our framework, we posit a practical mechanism mediated by a cellular network operator, collecting key external operational metrics from all SPs and then broadcasting this post-anonymization data as public information. Typically, the public information can constitute the observable opponent state $h_{-r,t}$ derived from $o_{-r,t}$ and the aggregated metrics $\sum_{j\in \mathcal{R}/\{r\}} n_{j,t} q_{j,t} $ within its reward function, thus facilitating SPs to more effectively adjust strategies toward equilibrium. As a common approach in resource allocation games \cite{5738225}, this realistically establishes a game that is strictly non-cooperative in its objectives but operates with partial observability.
\color{black}
\subsection{On Convergence Dynamics in Scaled Scenarios}
\label{app:discuss_convergence}
Theoretical guarantees and empirical results collectively support the scalability of our proposed frameworks. However, as the number of SPs increases, it is important to assess how convergence behavior and solution quality evolve.
\subsubsection{Impact on Convergence Rate (\texttt{PAC-MCOFL})} 
While Theorem 2 guarantees convergence to NE, the convergence rate is expected to decrease as $R$ increases. The proof of Theorem 2 relies on stochastic approximation, where the update variance influences convergence speed \cite{Jaakkola1994OnTC}. The term $\max_{a_{-r,t}}Q_t$ in the update rule unavoidably introduces high variance as $R$ grows, which slows the stochastic approximation process. Therefore, while convergence is guaranteed, its speed is expected to decrease with large $R$.
\subsubsection{Impact on Equilibrium Quality (\texttt{PAC-MCOFL-p})} From Theorem 1, the Q-value approximation error is bounded by the KL divergence between the generated and optimal joint policies. Assume the generator adopts a factorized approximation $\tilde{\pi}_{-r}^{\dagger}\left(a_{-r}\right)=\prod_{j \neq r} \tilde{\pi}_j\left(a_j \mid \cdot\right)$, and each KL term satisfies $D_{\mathrm{KL}}\left(\tilde{\pi}_j \| \pi_j\right) \leq \epsilon$. Then,
\begin{equation}
    |\mathbb{E}_{\tilde{\pi}_{-r}^{\dagger}}[Q]-\max _{a_{-r}} Q| \leq C \cdot \sqrt{(R-1) \cdot \epsilon}=O(\sqrt{R}).
\end{equation}
This implies that as $R$ grows, the quality degradation remains sublinear and bounded.

More importantly, our framework is explicitly designed to mitigate this potential error growth. In practice, our generator is trained not for perfect imitation of true opponents' policy but for utility maximization via the loss in Eq. (36). This objective steers the generator toward conjectures that maximize our agent’s return while remaining consistent with historical behavior. As $R$ increases and the true joint policy becomes combinatorially complex, the network’s capacity and KL regularization bias it toward a simpler, robust abstraction of opponents’ policies. Although this abstraction incurs a nonzero KL error (bounded by Theorem 1), it prevents overfitting to chaotic, high-dimensional opponent actions, yielding stable learning dynamics and a high-quality, practical equilibrium.

\subsection{Future Research Directions}
\label{app:discuss_future}
\subsubsection{Considerations for Dynamic Expectile Factor $\tau$ Adaptation}
This appendix elaborates on potential mechanisms for dynamically adapting the expectile factor, $\tau$, for each SP. A natural idea is to link $\tau$ with the agent's real-time performance by designing simple and practical update rules for state-driven heuristic adaptation. For example, consider the following update rule 
\begin{equation}
\tau_{r, t+1}= \begin{cases}\min \left(\tau_{\max }, \tau_{r, t}+\Delta \tau_{\rm inc}\right) & \text { if } \Gamma_{r, t}<{\rm Thr_{acc}} \\ \max \left(\tau_{\min }, \tau_{r, t}-\Delta \tau_{\rm dec}\right) & \text { if } \Gamma_{r, t} \geq {\rm Thr_{acc}}\end{cases}
\end{equation}
where ${\rm Thr_{acc}}$ denotes target accuracy threshold that triggers the adaptation, $\tau_{\rm inc}$ and $\tau_{\rm dec}$ are small, constant step sizes, $\tau_{\max}$ and $\tau_{\min}$ denote a reasonable range (e.g., 0.1 and 0.9). This rule-based mechanism allows an agent to automatically become more conservative (by increasing $\tau$) when its performance is unsatisfactory, and more aggressive (by decreasing $\tau$) when its performance meets the goal, directly linking strategic risk-taking to empirical outcomes. Furthermore, more sophisticated update rules can also be explored—for instance, the update could be governed by a function $f(\cdot)$, taking as input a summary of the agent’s performance history (e.g., average reward or accuracy over the last several rounds). These are preliminary ideas, and we plan to investigate such mechanisms in greater depth in future work.
\subsubsection{Considerations for Adaptive Granularity $\zeta_m$}
While we use pre-defined granularities $\zeta_m$ in TCAD, a promising future direction is to enable their dynamic adaptation. For instance, the granularity $\zeta_m$ could be adjusted based on the training stage or policy uncertainty. An agent might start with a larger step size for rapid exploration and reduce it as the policy converges to allow for fine-tuning. This could be formulated as $\varsigma_{m, t+1} = \alpha \cdot \varsigma_{m, t}$, where $\alpha<1$. Investigating such adaptive mechanisms remains an open topic for our future work.


\end{appendices}

\end{document}